%% file: main.tex
\documentclass{article}

\PassOptionsToPackage{numbers}{natbib}
\usepackage[preprint]{neurips_2023}

\usepackage{algorithm}
\usepackage{algorithmic}

\usepackage[utf8]{inputenc} 
\usepackage[T1]{fontenc}    
\usepackage{hyperref}       
\usepackage{url}            
\usepackage{booktabs}       
\usepackage{amsfonts}       
\usepackage{nicefrac}       
\usepackage{microtype}      
\usepackage{xcolor}         
\usepackage{graphicx}
\usepackage{subcaption}

\usepackage{amsmath}
\usepackage{amssymb}
\usepackage{mathtools}
\usepackage{amsthm}
\usepackage{bbm}

\usepackage[capitalize,noabbrev]{cleveref}

\usepackage{tikz}
\usepackage{listofitems} 
\usetikzlibrary{arrows.meta, shapes.misc, positioning, shapes.geometric,calc} 
\usepackage[outline]{contour} 
\usepackage{wrapfig}
\usepackage{soul}
\usepackage{float}

\theoremstyle{plain}
\newtheorem{theorem}{Theorem}[section]
\newtheorem{proposition}[theorem]{Proposition}

\theoremstyle{definition}

\theoremstyle{remark}

\usepackage[textsize=tiny]{todonotes}
\newcommand{\fbase}{f_{\mathrm{b}}}
\newcommand{\fnew}{f_{\mathrm{n}}}

\newcommand{\thetabase}{\theta_{\mathrm{base}}}
\newcommand{\thetanew}{\theta_{\mathrm{new}}}

\newcommand{\acc}{\mathrm{acc}}

\newcommand{\crel}{C_{\mathrm{rel}}}

\newcommand{\dkl}{D_{\mathrm{kl}}}
\newcommand{\dtrain}{\mathcal{D}_{\mathrm{train}}}
\newcommand{\dval}{\mathcal{D}_{\mathrm{val}}}
\newcommand{\dupdate}{\mathcal{D}_{\mathrm{update}}}
\newcommand{\data}{\mathcal{D}}
\newcommand{\ntrain}{n_{\mathrm{train}}}
\newcommand{\nval}{n_{\mathrm{val}}}

\newcommand{\amc}{AMC}
\newcommand{\conf}{Conf}
\newcommand{\avgconf}{AvgConf}

\DeclareMathOperator*{\argmax}{arg\,max}
\DeclareMathOperator*{\argmin}{arg\,min}

\tikzset{>=latex} 
\usepackage{xcolor}
\definecolor{myred}{HTML}{4CAF50}
\definecolor{myblue}{HTML}{2196F3}
\definecolor{mygreen}{HTML}{FF9800}
\colorlet{myorange}{orange!70!red!60!black}
\colorlet{mydarkred}{red!30!black}
\colorlet{mydarkblue}{blue!40!black}
\colorlet{mydarkgreen}{green!30!black}
\tikzstyle{node}=[thick,circle,draw=myblue,minimum size=18,inner sep=0.5,outer sep=0.6]
\tikzstyle{node in}=[node,green!20!black,draw=mygreen!30!black,fill=mygreen!25]
\tikzstyle{node hidden}=[node,blue!20!black,draw=myblue!30!black,fill=myblue!20]
\tikzstyle{node convol}=[node,orange!20!black,draw=myorange!30!black,fill=myorange!20]
\tikzstyle{node out}=[node,red!20!black,draw=myred!30!black,fill=myred!20]
\tikzstyle{connect}=[thick,mydarkblue] 
\tikzstyle{connect arrow}=[-{Latex[length=4,width=2]},thick,mydarkblue,shorten <=0.5,shorten >=1]
\tikzset{ 
  node 1/.style={node in},
  node 2/.style={node hidden},
  node 3/.style={node out},
}
\def\nstyle{int(\lay<\Nnodlen?min(2,\lay):3)} 

\title{	Maintaining Stability and Plasticity for Predictive Churn Reduction}

\author{%
  George Adam \\
  Department of Computer Science\\
  University of Toronto\\
  \texttt{alex.adam@mail.utoronto.ca} \\
  \And
  Benjamin Haibe-Kains \\
  Department of Medical Biophysics \\
  University of Toronto \\
  \texttt{benjamin.haibe.kains@utoronto.ca} \\
  \AND
  Anna Goldenberg \\
  Department of Computer Science \\
  University of Toronto \\
  \texttt{anna.goldenberg@utoronto.ca}
}

\begin{document}
\include{body}

\bibliography{neurips}
\bibliographystyle{plainnat}
\include{supp}

\end{document}

%% file: body.tex
\maketitle

\begin{abstract}
Deployed machine learning models should be updated to take advantage of a larger sample size to improve performance, as more data is gathered over time.
Unfortunately, even when model updates improve aggregate metrics such as accuracy, they can lead to errors on samples that were correctly predicted by the previous model causing per-sample regression in performance known as predictive churn. Such prediction flips erode user trust thereby reducing the effectiveness of the human-AI team as a whole.
We propose a solution called Accumulated Model Combination (\amc) based keeping the previous and current model version, and generating a meta-output using the prediction of the two models. \amc{} is a general technique and we propose several instances of it, each having their own advantages depending on the model and data properties. \amc{} requires minimal additional computation and changes to training procedures. We motivate the need for \amc{} by showing the difficulty of making a single model consistent with its own predictions throughout training thereby revealing an implicit stability-plasticity tradeoff when training a single model.
We demonstrate the effectiveness of \amc{} on a variety of modalities including computer vision, text, and tabular datasets comparing against state-of-the-art churn reduction methods, and showing superior churn reduction ability compared to all existing methods while being more efficient than ensembles.
\end{abstract}

\section{Introduction}
Model updates are necessary for many machine learning applications to improve performance over time \citep{Neyshabur2017-sp}. Performance evaluation using common aggregate metrics such as accuracy can hide nuanced differences between the original model and updated model \cite{hossin, Hardt2016-mu}. For example, models can make diverse errors at the sample level where switching from one model to another can cause perceived instability from the perspective of users even if overall the models perform the same, or the new model is better. This perceived instability is particularly relevant in decision support settings where the user and model can be viewed as a team, and trust is required to maximize the model's utility. Trust is the probability that the user will accept/incorporate the model's predictions as part of their decision making. It is conditional on the particular sample being predicted since the model might not make any mistakes on easy samples, thus enabling the user to have a high degree of trust on such samples. As the user works with the model for some time, they develop a mental model of when it is likely to be correct. The utility of the model to the workflow that it is supporting can be measured by the increase in speed or improvement in outcomes relative to a human-only version of the workflow. Utility therefore depends on trust since if a user does not trust the model's predictions, the model cannot support decisions. 
The flipping of predictions made by a new model relative to a base model is referred to as predictive churn \cite{Cormier_undated-st}. Not all churn is undesirable as prediction flips which result in correct classification are ideal, and flips between erroneous predictions are benign. Thus, we focus on \emph{negative flips} (NFs) or relevant churn: samples correctly predicted by the base model and incorrectly predicted by the new model. Negative flips cause distrust because the user's previous idea of what samples the model should predict correctly is invalidated, thus decreasing utility \cite{Yin2019-mp,Bansal_undated-yh}.

Reducing negative flips has received more attention recently as the adoption of ML grows, with the most common solutions being variance reduction and prediction matching using ensembles and distillation respectively. Ensembles aim to average out the stochastic aspects of neural network training resulting from random initialization, data augmentation, stochastic regularization techniques such as dropout, and the non-convex nature of the loss landscape \cite{Zhao2022-uo}. They do so at a large computational cost since in order to reduce churn, the base model must be an ensemble, as does the new model, and this increases both training and inference costs linearly with the number of models in the ensemble. The assumption that the base model is an ensemble is impractical since any model that has already been deployed is not amenable to this technique. Distillation instead tries to balance learning on a combination of true labels and the base model's predictions thus introducing a stability-plasticity tradeoff. Focusing too much on matching base model predictions inhibits the ability to learn from new data, and ignoring the base model's predictions does not limit churn \cite{Jiang2021-ti}. 
To address the limitations of existing approaches, we make the following contributions:
\begin{enumerate}
    \item Show that completely eliminating negative flips between epochs when training a single model is infeasible even on small datasets. This instability compounds the stability-plasticity tradeoff that prediction-matching methods such as distillation have, motivating the need for a method which does not suffer from these limitations (Section \ref{sec:understanding_churn}).
    \item Introduce \amc{}: a general framework used to combine base and new model outputs for churn reduction. Provide several versions of \amc{} giving practitioners the flexibility to adjust for efficiency depending on their use case (Section \ref{sec:amc}).
    \item Show superior churn reduction performance compared to distillation and ensembles. In particular, when combining average prediction confidence throughout training with final model prediction confidence to choose between models, churn is decreased substantially compared to using either score individually. We also investigate what role model calibration plays in reducing churn when (Section \ref{sec:churn_results}).
\end{enumerate}

\section{Related Works}
\noindent Predictive churn has been mentioned in the literature under other terms such as performance regression and model backward compatibility. We focus on the model adaptation notion of churn where the goal is to learn a new model with the same architecture as the base model while leveraging additional data. 

\paragraph{Distillation Methods}
The most efficient class of methods for reducing churn are based on distillation \cite{Hinton2015-wb} where the optimization objective is modified to include a term that biases the predictions of the new model towards those of the base model. \citeauthor{Cormier_undated-st} introduced the general concept of a stabilization operator meant to allow learning on new data while keeping predictions consistent with the base model. The stabilization operator used is referred to as anchor loss by other works, and is a distillation-based objective which trains on a mixture distribution of the ground truth one-hot targets and predictions made by the base model, similar to label smoothing \cite{Muller2019-ux}. \citeauthor{Anil2018-dn} proposed co-distillation for reducing the variance notion of churn such that predictions are reproducible even when it is not feasible to control the random initialization. Co-distillation works by training 2 models in parallel using each other's predictions for distillation, and only one of the models is kept upon convergence. This procedure is then repeated, again keeping just one of the models upon convergence such that the churn between the two runs is significantly less than when training two models independently from different initializations. \citeauthor{Bhojanapalli2021-wl} built upon co-distillation by combining it with entropy regularization for further churn reduction. 
Both co-distillation and anchor loss modify the training procedure for the base model which is impractical for models that are already deployed. \citeauthor{Jiang2021-ti} address this limitation with an objective similar to anchor loss that achieves SOTA churn reduction in the model adaptation setting. \citeauthor{Yan2020-hg} introduced a new distillation objective called focal loss where samples correctly predicted by the base model undergo a stronger distillation loss than other samples thereby allowing the new model to learn more on samples that the base model predicted incorrectly.

\paragraph{Ensemble Methods}

\citeauthor{Yan2020-hg} and \citeauthor{Bahri2021-ef} showed that though impractical due to their high training and inference costs, ensembles operating in logit space are by far the most effective way of reducing churn. Instead of having a single base model and single new model, ensembles in the context of churn reduction use a collection of base models and a collection of new models. In an attempt to reduce ensemble inference cost, \citeauthor{Zhao2022-uo} distill the knowledge from the average logits of an ensemble to a single model. Their method ELODI was shown to be more efficient, but not quite as effective in terms of both accuracy and churn reduction as ensembles. ELODI also requires training an ensemble for both the base and new model prior to applying distillation, so it is still very expensive, and is not applicable to already deployed models. \citeauthor{Yan2020-hg} showed that much of the churn reduction benefit of ensembles is due to the increase in accuracy, though some can be attributed to the decreased variance. \citeauthor{Cai2022-ry} showed that storing models over time is effective for reducing churn in the structured prediction setting for both syntactic and semantic parsing tasks. In particular, they focus on the case of switching from one type of parser to another (model upgrade), rather than our setting of accumulating more data over time (model adaptation).  A re-ranking procedure is used where the new model generates a set of candidate predictions, and the base model chooses from these predictions which is fundamentally different from \amc{} and is not applicable to classification tasks.

\section{Problem Setup}
\label{sec:problem_setup}
The task of interest is supervised classification setting where $\dtrain = \{(x^{(i)}, y^{(i)}) \}_{i=1}^{\ntrain}$ is our initial training data comprised of samples $x \in \mathbb{R}^d$, $y \in [k]$. We would like to learn the parameters $\theta$ of a model $f(\cdot ; \theta)$ where $f: \mathbb{R}^d \rightarrow \mathbb{R}^k$. Let $\ell$ be the cross-entropy loss, and $\phi: \mathbb{R}^k \rightarrow \Delta^{k}$ be the Softmax function mapping from logits to the $k-1$ dimensional simplex. The parameters of the base model $\thetabase$ are obtained via empirical risk minimization
\begin{align*}
    \thetabase = \argmin_{\theta} \frac{1}{|\dtrain|} \sum_{(x, y) \in \dtrain} \ell (\phi(f(x; \theta)), y)
\end{align*}

\noindent where $\ell$ is the cross entropy loss function. Given additional data $\dupdate$ we would like to learn new parameters $\thetanew$ such that the churn between $\thetabase$ and $\thetanew$ is less than a tolerance $\epsilon$. Specifically, let $\mathcal{D} = \dtrain \cup \dupdate$, and $\sigma: \mathbb{R}^{k} \rightarrow [k]$ be the $\argmax$ function mapping from soft model scores to a hard prediction. We use $\fbase$ and $\fnew$ to denote $f(\cdot; \thetabase)$ and $f(\cdot; \thetanew)$ for cleaner notation. The churn-constrained optimization problem is formulated as follows \cite{Jiang2021-ti}
\begin{align*}
    \min_{\fnew} \frac{1}{|\mathcal{D}|} \sum_{(x, y) \in \mathcal{D}} \ell \left(\phi \left(\fnew(x) \right), y \right) \:  \: \mathrm{s.t.} \:  \:  C(\fbase, \fnew) < \epsilon \\
    C(\fbase, \fnew) := \frac{1}{|\mathcal{D}|} \sum_{(x, y) \in \mathcal{D}} \mathbbm{1}[\sigma(\fbase(x)) \neq \sigma(\fnew(x))]
\end{align*}

\noindent We use a hard definition of churn is used rather than the soft, divergence-based churn provided by \cite{Jiang2021-ti}. We follow the standard assumption in the churn reduction literature that $\dtrain$ and $\dupdate$ are sampled from the same distribution $p(x, y)$. This ensures that observed churn is attributable to updating, and is not confounded by changes in data distribution.

\citeauthor{Jiang2021-ti} showed that the above constrained optimization problem is equivalent to distillation by training with a mixture of the ground truth one-hot target and output probability from $\fbase$
\begin{equation}
\label{eq:vanilla_distillation}
    \min_{\fnew} \frac{1 - \alpha}{|\mathcal{D}|} \sum_{(x, y) \in \mathcal{D}} \ell (\phi(\fnew(x)), y) + \frac{\alpha}{|\mathcal{D}|} \sum_{(x, y) \in \mathcal{D}} \ell (\phi(\fnew(x)), \phi(\fbase(x)))
\end{equation}

\noindent where $\alpha$ is a hyperparameter. Distillation introduces an explicit stability-plasticity tradeoff between learning on new data and matching the predictions of the base model which can be controlled by varying $\alpha$. It allows for substantial churn reduction and is SOTA among non-ensemble based methods \cite{Jiang2021-ti}, but achieving further churn reduction requires limiting the performance of $\fnew$. 

In addition to negative flips, $C(\fbase, \fnew)$ also counts samples that both $\fbase$ and $\fnew$ predict incorrectly, what can be considered as benign flips. Negative flips are most disruptive to user-model workflows, so the \textit{relevant churn} quantity that focuses on negative flips instead is the focus of our work \cite{Yan2020-hg}
\begin{align*}
      \label{eqn:relevant_churn}
      \crel(\fbase, \fnew) = \frac{1}{|\mathcal{D}|} \sum_{(x, y) \in \mathcal{D}} \mathbbm{1} 
      \left[  \sigma(\fnew(x)) \neq \sigma(\fbase(x)) \land \sigma(\fbase(x)) = y \right]
\end{align*}

\section{Feasibility of Self-Consistent Learning}
\label{sec:understanding_churn}
While churn is defined above between to occur between two models trained on different data, it is important to note that it also occurs during the course of training a single model. \cite{Toneva2018-js} introduced the notion of a forgetting event where if $\hat{y}_i^{t} = \argmax_{k} f(x_{i}, \theta^{t})$ and $\acc_{i}^{t} = \mathbbm{1} [\hat{y}_{i}^{t} = y_{i}]$, then a forgetting event occurs at epoch $t+1$ if $\acc_{i}^{t} > \acc_{i}^{t+1}$. On CIFAR-10, it was discovered that only $\sim 15$k samples are unforgettable (i.e. no forgetting event occurs once correctly predicted the first time), meaning that the remaining $\sim 35$k samples have unstable predictions during training. \cite{Srivastava2020-gx} observe that samples which undergo the most prediction flips during training tend to be the samples which incur negative flips between model versions. Furthermore, if $\fnew$ forgets its own predictions during training, this can limit the effectiveness of churn reduction methods such as distillation which rely on matching predictions between models. This is because even if $\fnew$ matches the prediction of $\fbase$ on a sample $x$ at some timestep $t$, this might not longer be true at timestep $t+1$.  Hence, a churn reduction method that works by regularizing $\fnew$ should ideally also make model training more stable for maximum churn reduction. We hypothesize that some of the negative flips from one epoch to another can be attributed to samples being incompatible with a given update. We define gradient compatibility as the cosine similarity between the batch gradient and individual sample gradient 
\begin{gather*}
    g_{i} =  \nabla_{\theta} \ell(\phi(f(x^{(i)}; \theta)), y^{(i)}) \qquad g = \frac{1}{|\mathcal{B}|} \sum_{i \in \mathcal{B}} g_{i} \qquad
    \mathrm{comp}(g, g_{i}) := \frac{\left\langle g, g_{i} \right\rangle}{||g|| \: ||g_{i}||} 
\end{gather*}

The set of incompatible samples when performing a weight update on a given batch with  set of indices $\mathcal{B}$ is then $\mathcal{S}_{\mathrm{inc}} = \{ (x^{(i)}, y^{(i)}) \in \mathcal{D} \: | \: \mathrm{comp}(g, g_{i}) < 0\}$, i.e. samples that incur an increase in loss when taking a step in the direction of the average gradient. 
The blue region in figure \ref{fig:gradient_similarity_distribution} shows the distribution of cosine similarities between the average gradient and per-sample gradients at a given timestep when training a LeNet model on SVHN. Even as the model fits the training data increasingly well, there are still many incompatible samples.

\subsection{Reduction of Incompatible Gradients}
Reducing incompatible gradients can be formulated as a quadratic programming problem as follows
\begin{align*}
    \min_{\Tilde{g}} \quad \frac{1}{2} &||g - \Tilde{g}||^{2}_{2} \quad
    \mathrm{s.t.} \quad \langle\Tilde{g}, g_{j}\rangle \: \geq 0 \: \: \: \forall j \in [\mathcal{B}]
\end{align*}

\begin{figure}[t]
\centering
\begin{subfigure}[t]{0.23\linewidth}
\centering
\includegraphics[width=0.99\columnwidth]{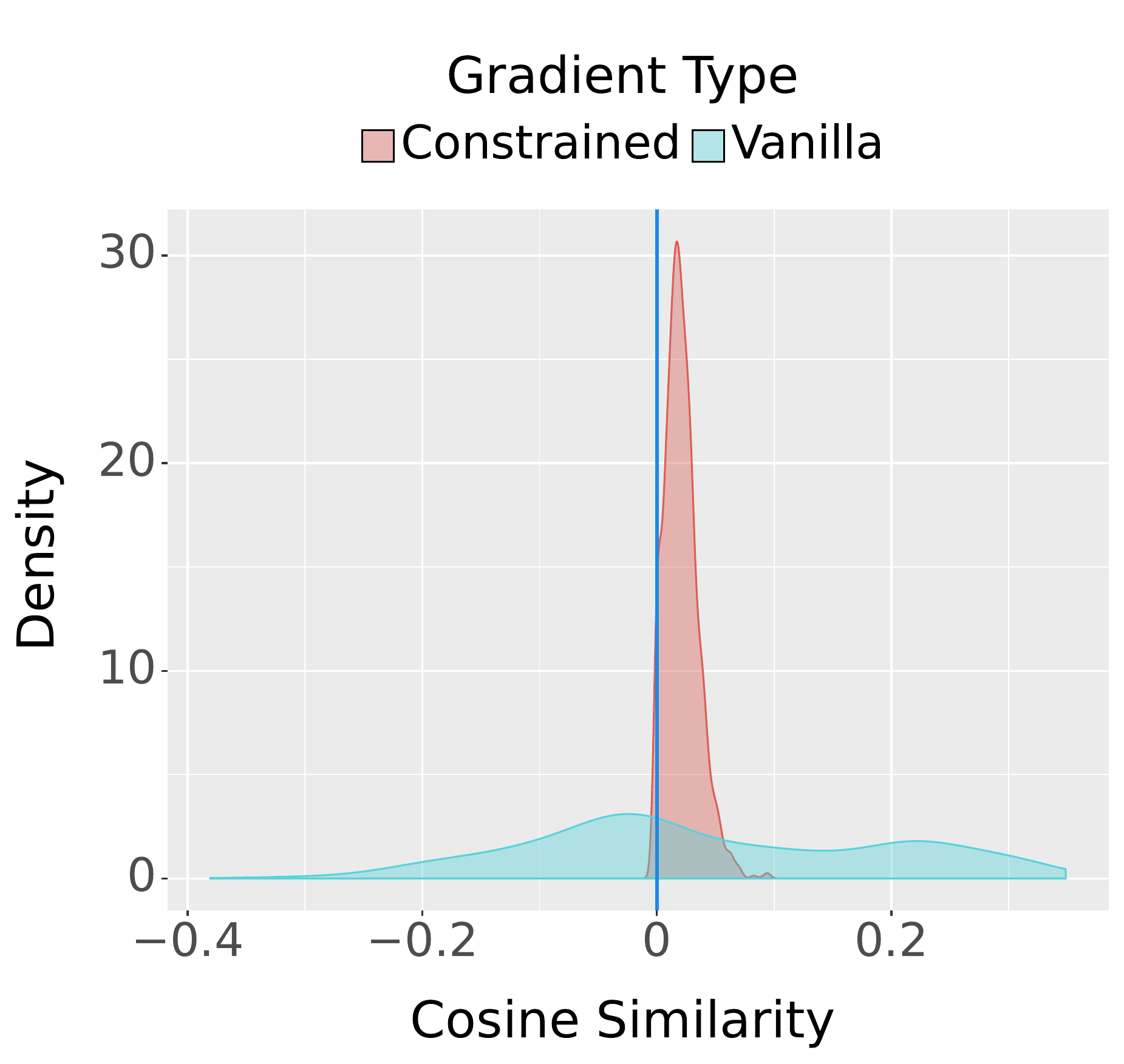}
\caption{}
\label{fig:gradient_similarity_distribution}
\end{subfigure}
\hfill
\begin{subfigure}[t]{0.23\linewidth}
\centering
\includegraphics[width=0.99\columnwidth]{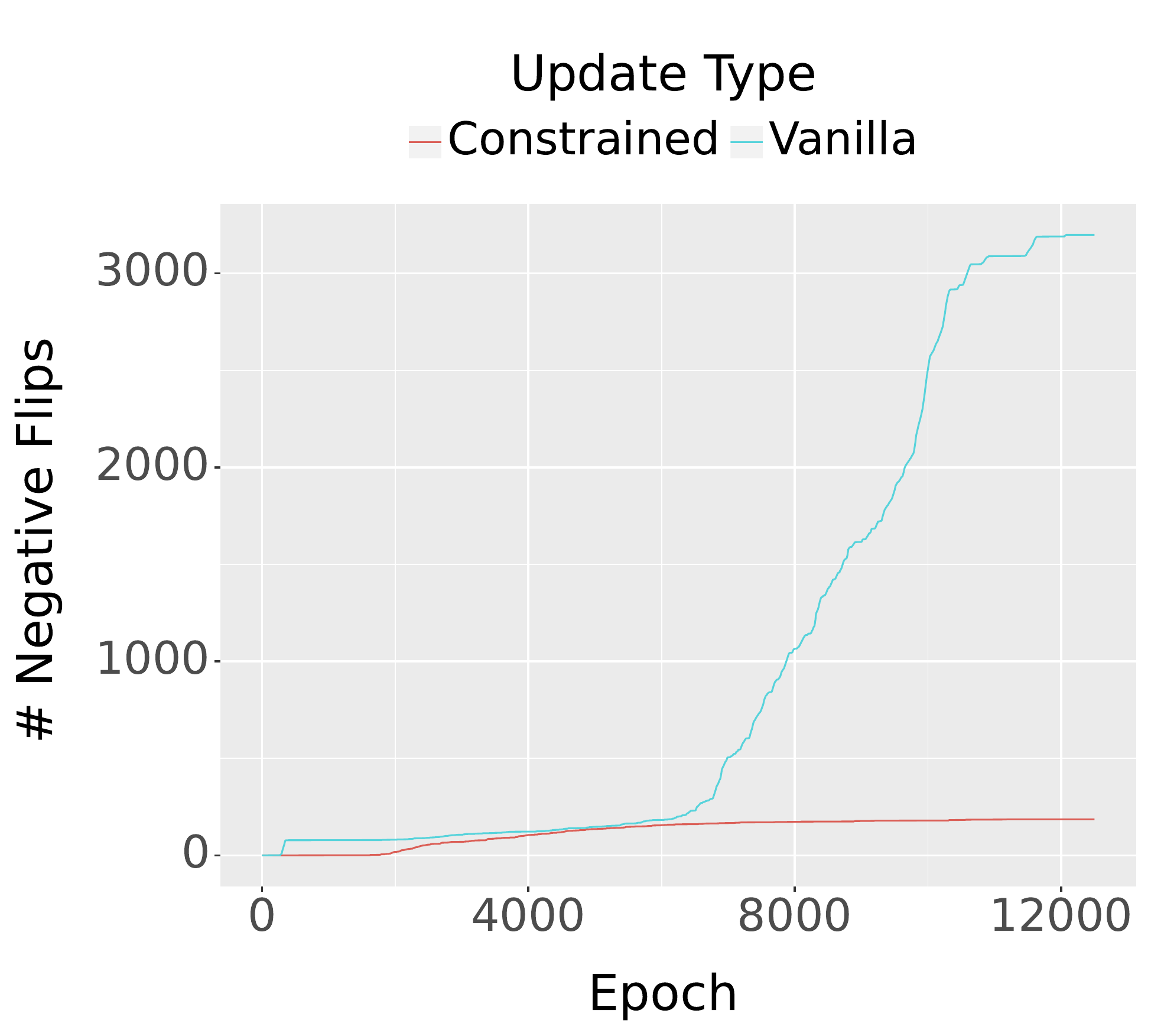}
\caption{}
\label{fig:nfr_evolution}
\end{subfigure}
\hfill
\begin{subfigure}[t]{0.23\linewidth}
\centering
\includegraphics[width=0.99\columnwidth]{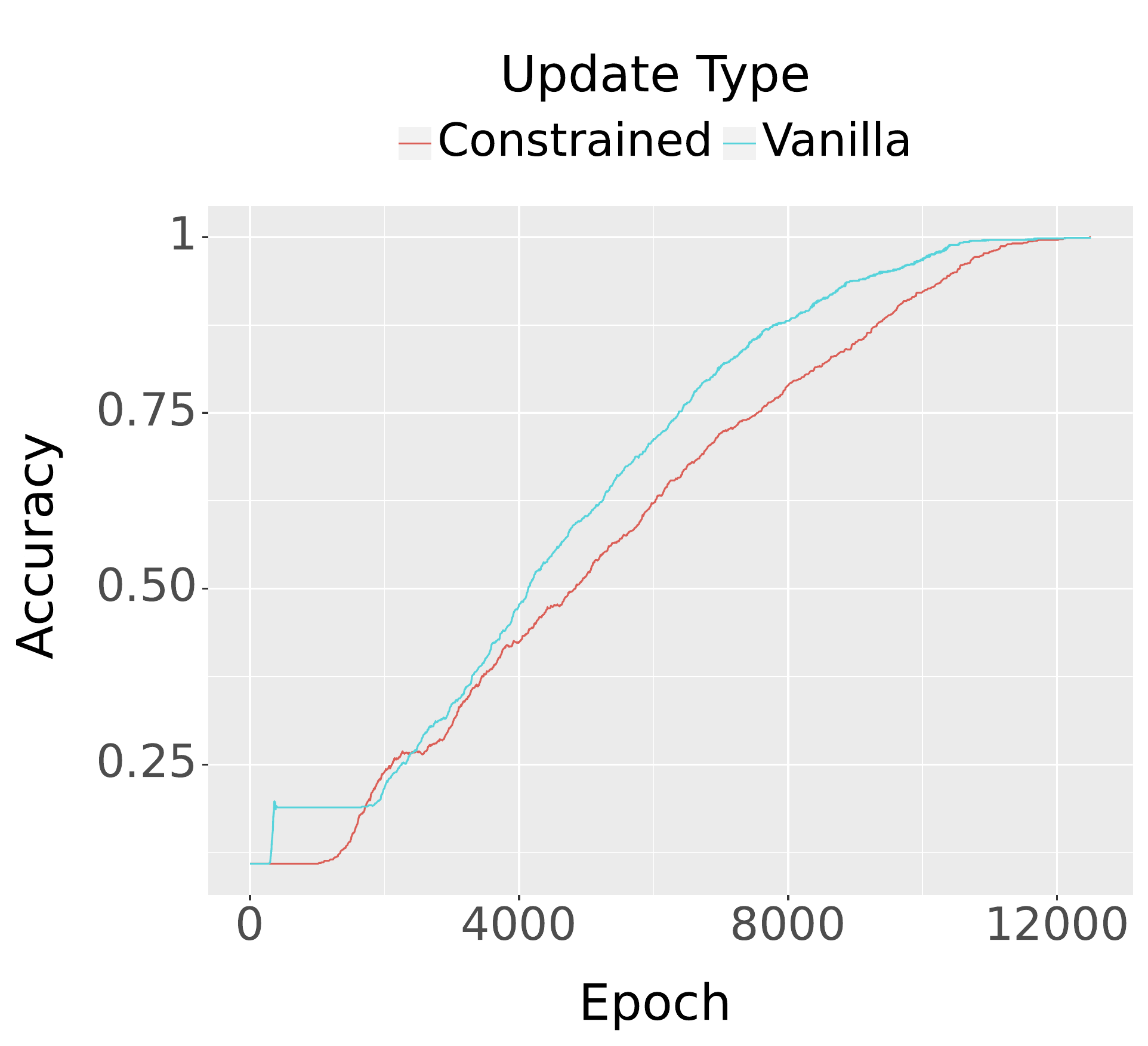}
\caption{}
\label{fig:svhn_low_curvature_accuracy}
\end{subfigure}
\hfill
\begin{subfigure}[t]{0.23\linewidth}
\centering
\includegraphics[width=0.99\columnwidth]{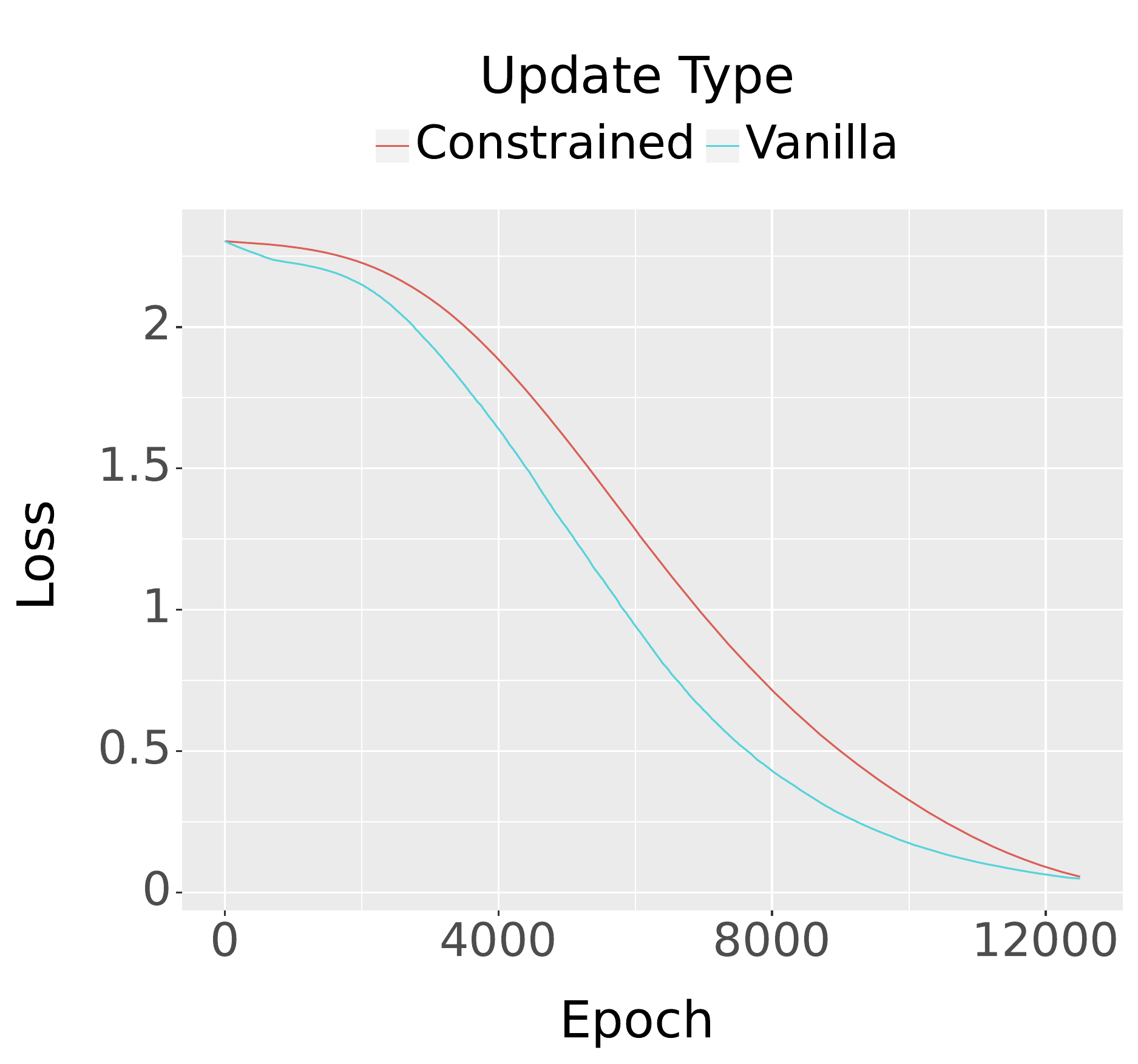}
\caption{}
\label{fig:svhn_loss}
\end{subfigure}
\caption{a) Gradient cosine similarity at a late epoch when training a LeNet model on a random subset of 1000 samples from SVHN. The average batch gradient when using regular gradient descent is incompatible with many per-sample gradients (blue area). Using constrained optimization, we can find a direction that is compatible with all per-sample gradients (red area). b) The cumulative number of negative flips while training until 100\% accuracy. The constrained optimization approach is able to eliminate many negative flips, confirming that some negative flips are a result of incompatible gradients. c) and d) show that both approaches achieve 100\% accuracy and near-zero loss by the final epoch of training.}
\label{fig:gem_flips}
\end{figure}

\noindent Solving the dual of this quadratic programming problem is more efficient in many cases since the optimization is done over the number of samples in the training set $\mathcal{B}$ instead of being over the number of parameters in the model which can easily reach millions for moderate size CNNs (details in appendix \ref{sec:dual_qp}). We perform experiments on a 1000 sample subset of SVHN using a learning rate of 0.0005 to show how effective the above formulation is at reducing negative flips compared to regular gradient descent. Results for additional hyperparameter settings can be found in appendix \ref{sec:dual_qp}. Note that the above formulation can only limit all incompatible samples when doing full-batch gradient descent, which is the setting we consider in our experiments.
Figure \ref{fig:gradient_similarity_distribution} shows that this constrained gradient descent is very effective at eliminating nearly all incompatible samples. Figure \ref{fig:svhn_low_curvature_accuracy} shows that both the constrained and vanilla optimization gradient descent are able to achieve 100\% training accuracy, and loss decreases monotonically so the source of negative flips cannot be attributed to using a learning rate that is too large. The two gradient descent variants make a similar number of NFs until roughly epoch 6000, at which point the number of NFs made by vanilla gradient descent increases rapidly. Initially when accuracy is low, incompatible gradients mainly cause benign flips since most samples are incorrectly predicted. As the set of correctly predicted samples becomes larger, incompatible gradients have an increased potential for causing NFs, thus explaining the increase in NFs by vanilla gradient descent once a high enough accuracy is reached. Some NFs persist even when using constrained gradient descent suggesting that incompatible gradients are not the only explanation for NFs. Indeed, it is possible for an NF to occur while the cross-entropy loss on a sample decreases. This is because cross-entropy loss uses only the predicted probability for the ground truth class. If the predicted probability for another class $j$ is close to that for the ground truth class $y$ ($\psi(f(x))_{y} - \max_{j \neq y} \psi(f(x))_{j} < \epsilon$), then it is possible for the class $j$ output to increase more than class $y$, thus changing ranking and causing an NF. 
Since neural networks are biased towards learning simple functions early on during training \cite{Nakkiran2019-sg, Hu2020-nh}, negative flips during this early stage may be inevitable. NFs on complex or ambiguous samples may also be inevitable since such samples can require memorization to be predicted correctly, and this memorization is at odds with the generalizable features learned from easier samples. 

In summary, the performance of a prediction matching churn reduction method, of which distillation is a prime example, is limited by both an explicit stability-plasticity tradeoff, and prediction instability while training $\fnew$. 
In order to avoid these limitations, we propose allowing $\fnew$ to learn unconstrained, keeping the new model's predictions when it is more likely to be correct than the base model, and reverting to the base model otherwise.

\section{Accumulated Model Combination (AMC)}
\label{sec:amc}
We begin by highlighting the challenges of using ensembles for churn reduction, then introduce the proposed method \amc{}. as well as some possible scores  that \amc{} can use to choose between models, and an alternative learning-based approach to \amc{}.
The ensemble approach for reducing churn requires that both the base and new model are ensembles: $\mathcal{F}_{\mathrm{base}} = \{\fbase^{(1)}, ..., \fbase^{(M)}\}$, $\mathcal{F}_{\mathrm{new}} = \{\fnew^{(1)}, ..., \fnew^{(M)}\}$, where $M$ is the number of models per ensemble, and inference is done by averaging the logits of the ensemble predictions $\mathcal{F}(x) = \frac{1}{M} \sum_{i=1}^{M} f^{(i)}(x)$.
Since ensembles result in both a more accurate base model and new model, most of the churn reduction advantages come from the increase in accuracy, with some additional benefit coming from less variance in predictions as shown by \cite{Zhao2022-uo}. Using ensembles for churn reduction has two major challenges which limit its feasibility in practice. First, it increases both training and inference costs by a factor of $M$ which can be prohibitively expensive. Second, since the base model needs to be an ensemble, it is not possible to apply this approach to a deployed model. 

\label{sec:stacking_definition}

\noindent To bypass these limitations, and reduce churn even further, we introduce \emph{\amc{}} (see appendix figure \ref{fig:stacking} for illustration) : an approach that fuses the output of $\fbase$ and $\fnew$ using a meta model $\psi$ to generate the final prediction. There are two categories of approaches for $\psi$: choosing between the outputs of $\fbase$ and $\fnew$, and generating a new output using $\fbase(x)$ and $\fnew(x)$ as input features.

\subsection{Choosing Between $\fbase$ and $\fnew$}
Let $S$ be a set of scoring functions $s: \mathcal{H} \times \mathbb{R}^{d} \rightarrow \mathbb{R}$ where $\mathcal{H}$ is the set of all models we consider, and define
\begin{align*}
\psi(x; \fbase, \fnew) := \begin{cases} 
      \fnew(x) & \bigwedge_{s \in S} [s(\fnew, x) > s(\fbase, x)] \\
      \fbase(x) & \mathrm{otherwise} \\
      \end{cases}
\end{align*}

\noindent By choosing between $\fbase$ and $\fnew$, the meta-model $\psi$ cannot make additional negative flips by definition, so churn is strictly reduced. Since $\fbase$ and $\fnew$ are not required to be ensembles, this approach trains $M-1$ models fewer than the ensemble approach at each update. Most of all, it allows for churn reduction even for an already deployed base model. We analyze the case where there is a single score based on prediction confidence and prove that $\psi$ in this case reduces churn while maintaining or improving accuracy relative to $\fnew$. $X, Y$ are capitalized to indicate that they are random variables rather than particular observations jointly distributed according to $p(x,y)$.
\begin{proposition}
Assume $\fbase$ and $\fnew$ are perfectly calibrated such that $\mathbf{Pr}(\hat{Y} = Y | \hat{P} = p) = p, \forall p \in [0, 1]$ where $\hat{Y} = \argmax_{k} f(X)_{k}$ (hard prediction) and $\hat{P} = \max_{k} \phi(f(X))_{k}$ (predicted probability). If $s = \max_{k} \phi(f(x))_{k}$ so that $\psi$ is also perfectly calibrated, then $\acc(\psi) \geq \acc(\fnew)$.
\end{proposition}
\begin{proof}
See appendix \ref{sec:proposition_proof}.
\end{proof}

\noindent It is easy to observe that  $\crel(\fbase, \psi) \leq \crel(\fbase, \fnew)$ (see eq \ref{eqn:relevant_churn}) since negative flips can only be reduced by using $\fbase$ for some samples instead of $\fnew$ which is what $\psi$ does. Thus, \amc{} can reduce $\crel$ without decreasing accuracy, hence it bypasses the stability-plasticity tradeoff. We refer to the above $s$ as 
\conf{} since it chooses the model with highest prediction confidence.
\conf{} has limited churn reduction capability even for perfectly calibrated models. 
Namely, it is possible that $\fbase$ is correct and $\fnew$ is wrong even when $s(\fnew, x) > s(\fbase, x)$. The expected number of irreducible NFs when using \conf{} is then the sum of the probability that for a given NF, $\fbase$ is right and $\fnew$ is wrong
\begin{align*}
\sum_{x \in B^{-}} [\underbrace{s(\fbase, x)}_{\fbase \mathrm{correct}} \underbrace{(1 - s(\fnew, x))}_{\fnew \mathrm{incorrect}}]
\end{align*}

\noindent where $B^{-}$ is the original set of NFs made by $\fnew$. To address this limitation, we propose using additional scores to provide further information about the correctness of $\fbase$ and $\fnew$. One such score is discussed below, followed by the alternate version of \amc{} based on generating new predictions.

\paragraph{Average Confidence}
Prediction stability over time when training $\fbase$ and $\fnew$ is used to create a score that complements \conf{}. 
Stability has been shown to be a proxy for sample difficulty \cite{Toneva2018-js, Maini2022-ei}, so it may help in reducing the NFs that cannot be identified by Conf. We define \avgconf{} as the average confidence of a model across training epochs for the final predicted class $s = \frac{1}{T} \sum_{t=1}^{T} \phi(f^{t}(x))_{\hat{y}}$
where $\hat{y} = \argmax_{k} f^{T}(x)$, $T$ is the number of epochs, and $f^{t}$ is a saved model checkpoint at epoch $t$. Unlike works that use stability measures for characterizing datapoint difficulty, \avgconf{} used for \amc{} does not require setting a threshold to define what stable and unstable is since the comparison being made is a relative one between $\fbase$ and $\fnew$. Computing the exact \avgconf{} for a test sample would remove the computational advantage of \amc{} since it requires storing $T$ checkpoints and performing inference with them. We show how to overcome this limitation by computing an approximate \text{AvgConf} for test samples in appendix \ref{sec:knn_avgconf} where the complexity is shown to be subsumed by the inference cost of deep networks. 

\subsection{Generating New Predictions Using $\fbase$ and $\fnew$}
\label{sec:learned_combination}
An alternative to choosing between model outputs is to learn a meta-model that makes predictions using the outputs of $\fbase$ and $\fnew$. This is known as stacking and it is able to correct the mistakes of both $\fbase$ and $\fnew$. Stacking has been around for decades \cite{Wolpert1992-xp}, though it has not been explored for churn reduction \cite{Ting_undated-af}.
A model $h^{*}:\mathbb{R}^{k} \times \mathbb{R}^{k} \rightarrow \mathbb{R}^{k}$ is learned to maximize accuracy on the validation set
\begin{gather*}
h^{*} = \argmin_{h \in \mathcal{H}} \frac{1}{|\mathcal{D}_{\mathrm{val}}|} \sum_{(x, y) \in \mathcal{D}_{\mathrm{val}}} \ell(h(\fbase(x), \fnew(x)), y) \qquad \; \; \psi(x; \fbase, \fnew) = h^{*}(\fbase(x), \fnew(x))
\end{gather*}

\noindent where $\mathcal{H}$ is the hypothesis space for the learned combination. We use 5-fold cross validation to select the best $h$. This approach is referred to as \amc{} Learned throughout the text. Choices of $h$ investigated include logistic regression, random forest, and gradient boosted decision trees. Hyperparameter choices are listed in appdendix \ref{sec:best_learned_model}. Note that this formulation is general and allows for model types which make hard or soft predictions to be used.

A possible limitation of \amc{} Learned is that unlike the score-based version of \amc, it is capable of introducing additional errors relative to $\fnew$. To ensure that \amc{} Learned reduces negative flips beyond the increase in accuracy that it provides, we propose using a distillation-based objective similar to eq \ref{eq:vanilla_distillation}  when learning $h$
\begin{align*}
    h^{*} = \min_{h \in \mathcal{H}} \; &\frac{1 - \alpha}{|\mathcal{D}_{\mathrm{val}}|} \sum_{(x, y) \in \mathcal{D}_{\mathrm{val}}} \ell (\phi(h(\fbase(x), \fnew(x))), y) \: \: + \\
    &\frac{\alpha}{|\mathcal{D}_{\mathrm{val}}|} \sum_{(x, y) \in \mathcal{D}_{\mathrm{val}}} \ell (\phi(h(\fbase(x), \fnew(x))), \phi(\fbase(x)))
\end{align*}

\noindent This restricts the model class to neural networks. Details about the architecture used can be found in the appendix \ref{sec:amc_distill_details}. This version of \amc{} is referred to as \amc{} Distill in the text.

\section{Results}
\label{sec:churn_results}
\paragraph{Data}
Our objective is to demonstrate a practical deployment scenario where a model is trained as soon as we accumulate sufficient data. This enables us to immediately begin making predictions during deployment (provide utility), and model updates are performed as new data is gathered. Such an approach is preferable to delaying the model's availability to users until a large-scale dataset has been collected.
We largely follow the experimental setup of \cite{Jiang2021-ti} and focus on benchmark computer image classification datasets, as well as text and tabular classification datasets. Data splits for training/validation/update can be found in appendix \ref{sec:training_details}. The split sizing was chosen such that $\fnew$ has a clear accuracy increase over $\fbase$ justifying a model update. All datasets used have a pre-defined test partition which is used for final evaluation. The following datasets and architectures are used
\begin{enumerate}
    \item CIFAR10, CIFAR100 \cite{Krizhevsky_undated-eb}, STL10 \cite{Coates2011-um} using ResNet18 \cite{He2015-ov}, and FairFace \cite{Karkkainen2019-co} using ResNet50.
    \item MNIST \cite{lecun-mnisthandwrittendigit-2010}, EMINST \cite{Cohen2017-vc}, KMNIST \cite{Clanuwat2018-md}, FashionMNIST \cite{Xiao2017-am}, SVHN \cite{Netzer2011-gf} using LeNet-5 \cite{Lecun1998-ll}
    \item AG News \cite{Zhang2015-lu}, IMDB \cite{Maas2011-ub} text classification using a Transformer architecture. Adult Income Prediction \cite{Dua:2019}, and Human Activity Recognition (HAR) \cite{Dua:2019} using a fully connected architecture.
\end{enumerate}

\paragraph{Training:} Early stopping like in \cite{Jiang2021-ti} is used to avoid having to find an optimal fixed number of epochs for each dataset/method combination which is complicated by the fact that $\fnew$ has access to more data and would require a different number of epochs than $\fbase$. $\fbase$ is trained on $\dtrain$, and $\fnew$ is trained from scratch on $\dtrain \cup \dupdate$. More training details can be found in appendix \ref{sec:training_details}. All experiments are done with 10 random seeds, and the mean accuracy/churn is reported in the tables.

\subsection{SOTA Method Comparison}
\label{sec:sota_comparison}
\citeauthor{Jiang2021-ti} investigated several churn reduction baselines and found that distillation outperforms all of them. Results for some of these baselines are included in the appendix due to space limitations, with distillation and ensembles being the focus in this section, the latter of which is known to be the SOTA churn reduction method \cite{Yan2020-hg,Zhao2022-uo}. The no regularization baseline which trains $\fnew$ from a random initialization is denoted by Cold, and training $\fnew$ using $\fbase$ as the initialization is denoted by Warm Start. 
The distillation approach from \cite{Jiang2021-ti} shown in Section \ref{sec:problem_setup} is used as the SOTA distillation approach to compare against (Distill), with the same search space for the hyperparameter $\alpha \in \{ 0.1, 0.2, ..., 0.9\}$ as in their paper for a fair comparison. For Ensemble, $M \in \{3, 5, 7 \}$ is considered as a larger number of models results in diminishing variance reduction relative to the increase in training costs. Finally, there are 5 versions of \amc{} considered: confidence score only (\amc{} Conf), AvgConf score only (\amc{} AvgConf), conjunction of Conf and AvgConf (\amc{} Combined), learning to generate a new output (\amc{} Learned), and the version of Learned with a distillation term in the loss (\amc{} Distill). Only \amc{} Learned and Distill require hyperparameter tuning, namely a search over the space of models used to combine the outputs of $\fbase$ and $\fnew$. The standard experimental setup in the churn reduction literature is to perform just a single model update, hence the experiments follow this design. Further description of baselines is found in appendix \ref{sec:training_details}.

\begin{table*}
\caption{Comparison of churn reduction methods. \amc{} outperforms both distillation and ensembles across all datasets in reducing $\crel$ (lower is better). Results are reported as a percentage.}
\begin{center}
\scalebox{0.73}{
\begin{tabular}{lcccc|ccccc}\toprule
Dataset & \multicolumn{1}{c}{Cold} & \multicolumn{1}{p{1.2cm}}{\centering Warm \\ Start} & \multicolumn{1}{c}{Distill} & \multicolumn{1}{c|}{Ensemble} & \multicolumn{1}{p{1.2cm}}{\centering \amc{} \\ Conf} & \multicolumn{1}{p{1.5cm}}{\centering \amc{} \\ Avg Conf} & \multicolumn{1}{p{1.2cm}}{\centering \amc{} \\ Combined} & \multicolumn{1}{p{1.5cm}}{\centering \amc{} \\ Learned}  & \multicolumn{1}{p{1.5cm}}{\centering \amc{} \\ Distill}
\\\midrule
CIFAR10      &   6.449 &      5.406 &        4.354 &    2.176 &    2.526 &        3.048 &        2.017 &       2.618 &                 \textbf{0.632} \\
CIFAR100     &   9.914 &        NaN &        5.814 &    4.113 &    4.092 &          NaN &        \textbf{3.357} &       7.622 &                   NaN \\
FairFace     &  12.454 &        NaN &        7.241 &    5.514 &    5.784 &        7.343 &        5.076 &       7.202 &                 \textbf{2.635} \\
FashionMNIST &   3.537 &        NaN &        2.778 &    1.166 &    1.312 &        1.281 &        0.784 &       1.573 &                 \textbf{0.624} \\
EMNIST       &   2.852 &        NaN &        1.877 &    0.972 &    1.091 &        1.198 &        \textbf{0.598} &       1.900 &                 1.133 \\
KMNIST       &   2.382 &      1.934 &        1.593 &    0.752 &    0.807 &        0.862 &        \textbf{0.450} &       0.946 &                 0.541 \\
MNIST        &   0.536 &        NaN &        0.359 &    0.153 &    0.194 &        0.184 &        \textbf{0.099} &       0.202 &                 0.141 \\
SVHN         &   6.415 &        NaN &        3.976 &    1.759 &    1.945 &        2.269 &        1.369 &       2.241 &                 \textbf{0.315} \\
STL10        &  12.391 &     10.177 &        6.592 &    5.186 &    4.971 &        7.006 &        4.445 &       6.448 &                 \textbf{2.226} \\ 
\midrule
Adult        &   2.780 &        NaN &        1.255 &    1.340 &    1.197 &        1.364 &        \textbf{0.711} &       1.874 &                 \textbf{0.818} \\
HAR          &   0.852 &      0.558 &        0.594 &    0.321 &    \textbf{0.352} &        0.470 &        \textbf{0.221} &       0.455 &                 0.465 \\
\midrule
AG-News      &   3.149 &      1.720 &        1.945 &    1.163 &    1.174 &        1.449 &        0.943 &       1.712 &                 \textbf{0.717} \\
IMDB         &   6.378 &      3.935 &        4.950 &    3.239 &    2.450 &        3.094 &        \textbf{2.150} &       4.286 &                 \textbf{2.019} \\
\bottomrule
\end{tabular}
}
\end{center}
\label{table:main_results}
\end{table*}

\begin{figure}[H]
\centering
\begin{subfigure}{0.24\textwidth}
\centering
\includegraphics[width=\textwidth]{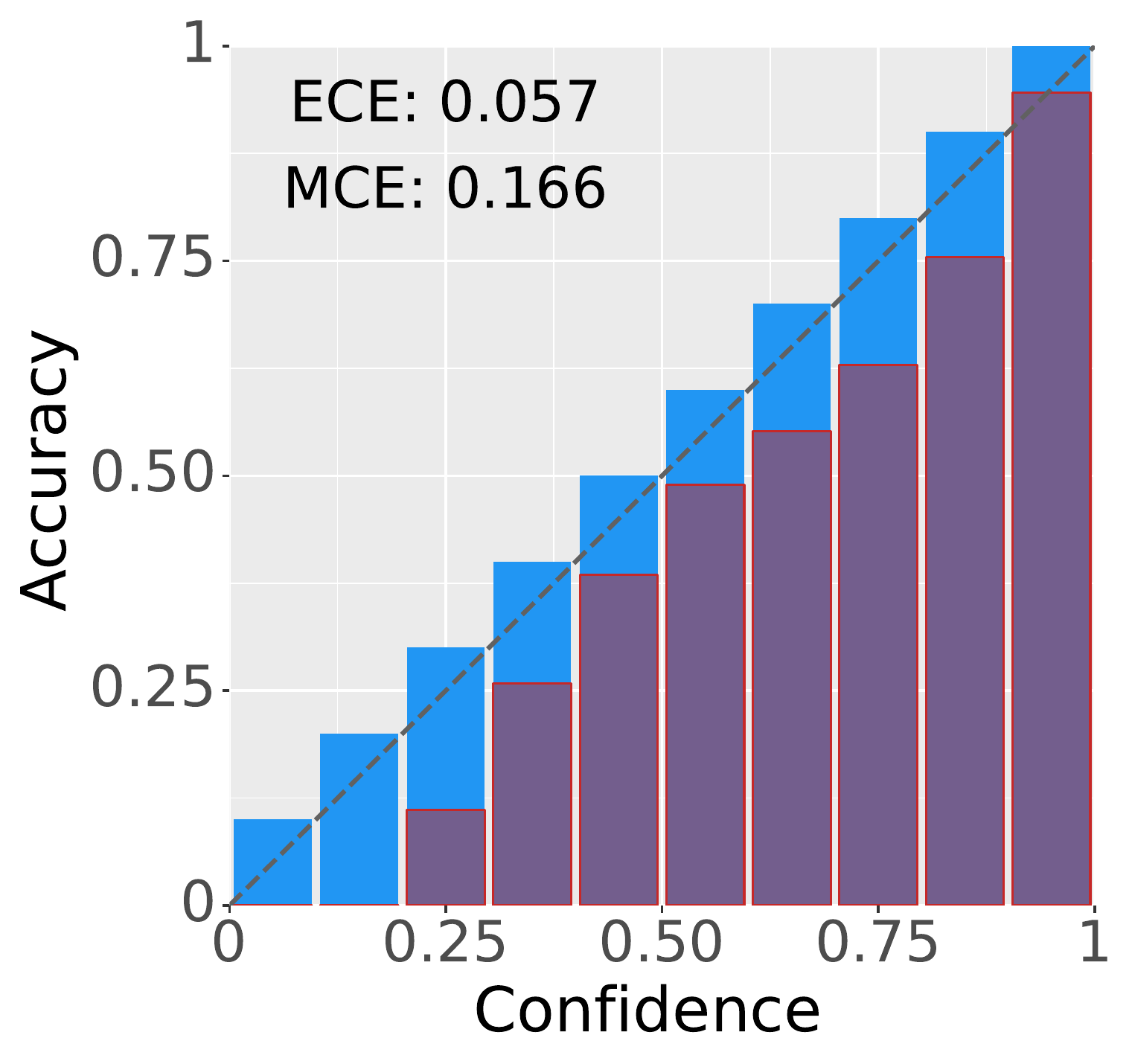}
\caption{$\fbase$ Uncalibrated}
\end{subfigure}
\begin{subfigure}{0.24\textwidth}
\centering
\includegraphics[width=\textwidth]{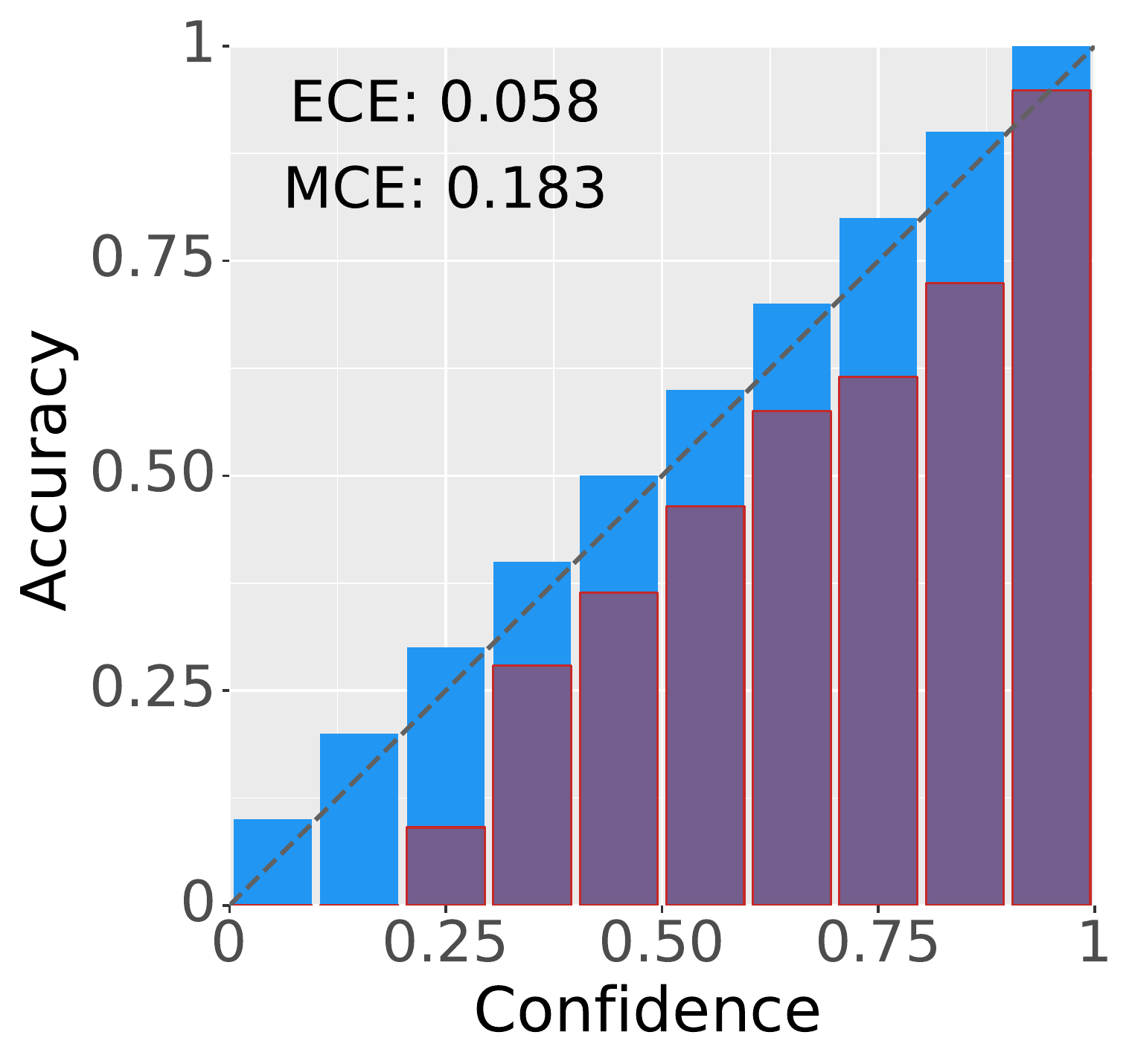}
\caption{$\fnew$ Uncalibrated}
\end{subfigure}
\begin{subfigure}{0.24\textwidth}
\centering
\includegraphics[width=\textwidth]{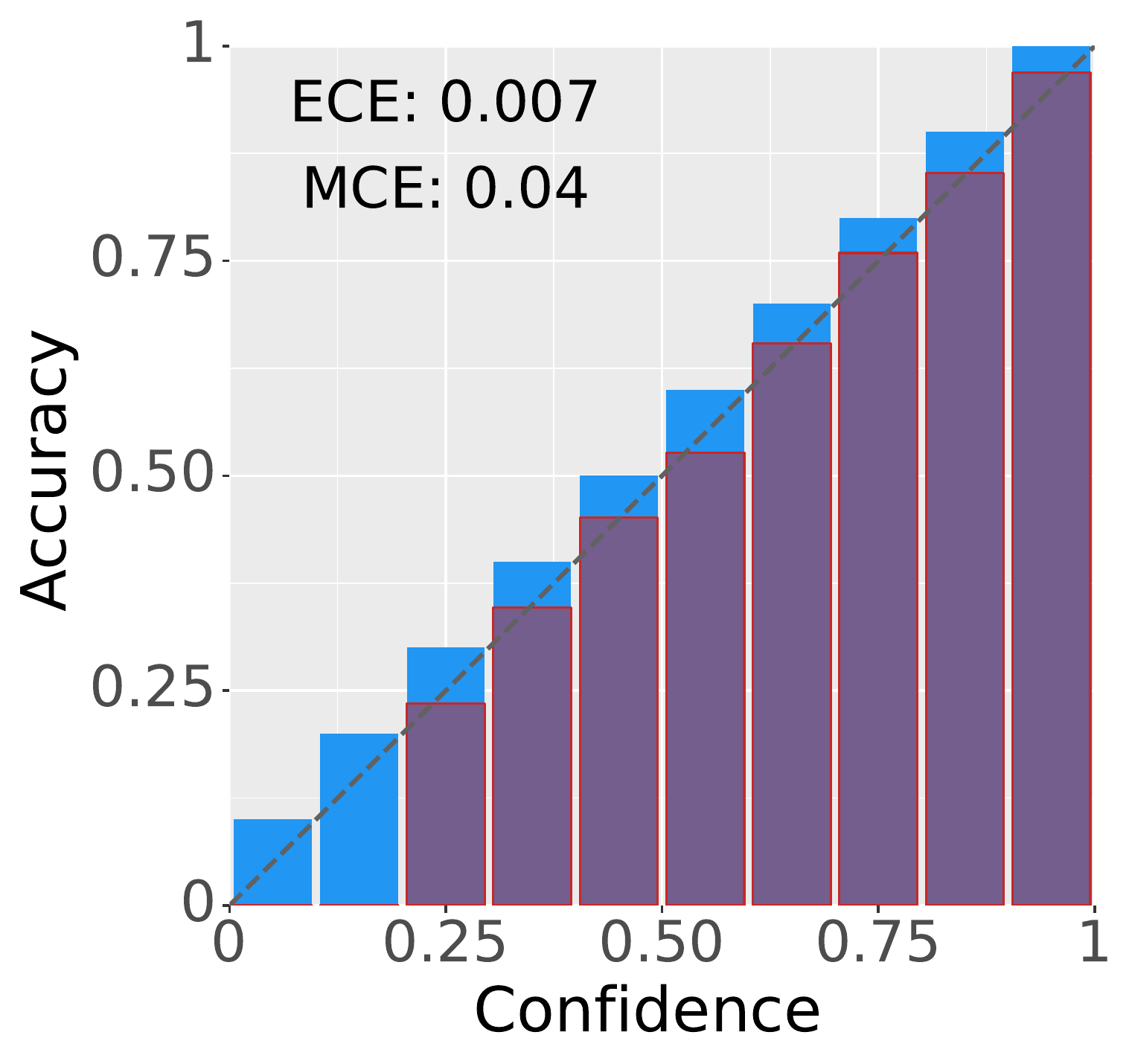}
\caption{$\fbase$ Calibrated}
\end{subfigure}
\begin{subfigure}{0.24\textwidth}
\centering
\includegraphics[width=\textwidth]{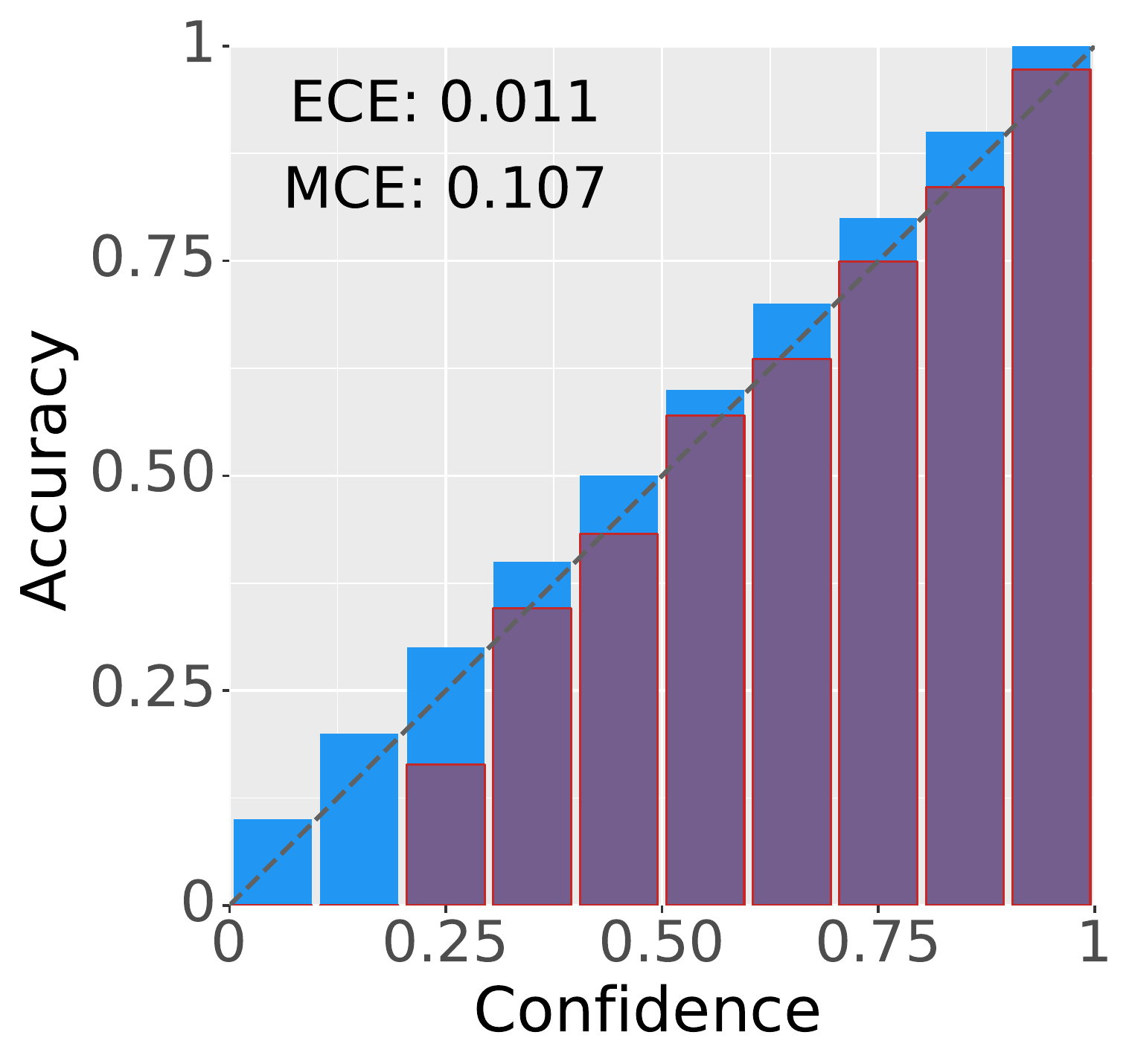}
\caption{$\fnew$ Calibrated}
\end{subfigure}
\caption{Calibration of ResNet18 models trained on CIFAR10.}
\label{fig:calibration}
\end{figure}

The negative flip rate ($\crel$) of the considered methods is reported in Table \ref{table:main_results}. The best performing method for each dataset is bolded. If the observed difference in $\crel$ is not statistically significant (one-sided t-test with $\alpha=0.05$), both methods are highlighted. In all cases, \amc{} outperforms both distillation and ensembles by a significant margin. All methods achieve accuracy as high as no regularization (Cold), or have NA entries otherwise implying that the method would not be used for that dataset since even if it may have good churn reduction ability, the sacrifice in accuracy is not acceptable. This is the same approach that \cite{Jiang2021-ti} take to reporting results (churn at cold accuracy), otherwise reporting both accuracy and churn in the same table is distracting and makes it unclear which method should be preferred. We report method accuracy in appendix table \ref{table:main_accuracy}. Note that \amc{} achieves more churn reduction than Ensemble at a fraction of the cost, in this case $\approx$7x less memory, training compute cost, and inference cost. The best version of \amc{} varies across datasets, though it is clear that \amc{} Combined and \amc{} Distill are the top two methods. Due to limited space, confidence intervals are not included in Table \ref{table:main_results}, but boxplots can be found in Appendix \ref{sec:performance_stats}. Interestingly, \amc{} Conf and \amc{} AvgConf are not particularly effective on their own, yet the combination of the two results in better churn reduction than Ensemble. The combined score results in fewer positive flips since the conjunction of conditions will cause $\fbase$ to be selected more often. However, the reduction in positive flips is similar to that of negative flips, which allows for accuracy to be largely maintained as seen in appendix table \ref{table:main_accuracy}. Appendix \ref{sec:ood_scores} investigates the effectiveness of several OOD detection scores in choosing between model outputs. \amc{} Learned is the least effective version at reducing churn for most datasets even though it is the most accurate version (appendix table \ref{table:main_accuracy}). This shows that our contribution of applying distillation in addition to stacking models (AMC Distill) is fundamental in achieving superior churn reduction since maximizing only for accuracy introduces additional negative flips (appendix \ref{sec:flip_counts}).

The ensemble size in table \ref{table:main_results} was limited to 7 as it is unlikely that any practical deployment scenario of a machine learning would allow for scaling the number of models past this limit. However, it is important to verify if there exists a number of models $M$ such that Ensemble matches the churn reduction ability of \amc{} given possible future developments in making ensembles more efficient. Due to the high cost of these experiments, we focus on CIFAR10, FairFace, and SVHN where there the churn reduction gap between Ensemble and \amc{} is greatest. Table \ref{table:ensemble_limit} shows the diminishing churn reduction ability of Ensemble as the number of models increases, revealing that \amc{} is still superior even when $M=17$. This demonstrates that the fundamentally different way in which \amc{} reduces churn compared to other methods cannot be surpassed by naively increasing computational resources, further emphasizing the need for our novel approach. 

\subsection{Role of Calibration}
\amc{} Conf works best under the assumption that predictions are calibrated, so we investigate if further churn reduction is achievable by improving calibration. The calibration of both models before and after temperature scaling can be seen in Figure \ref{fig:calibration} for ResNet18 models on CIFAR10. Both expected calibration error (ECE) and maximum calibration error (MCE) are significantly reduced through temperature scaling \cite{Guo2017-eh}. Surprisingly, this does not help reduce churn, or improve model accuracy. Accuracy without scaling is 84.17\% and after scaling it is 84.08\% while $\crel$ with scaling is 2.17\% without scaling and 2.19\% with scaling. Appendix table \ref{table:calibration_ranking_changes} shows that calibration results in a total of just 324 total changes in ranking between $\fbase$ and $\fnew$, of which 275 are benign, 20 are good (now choosing the correct model), and 29 are bad (now choosing the incorrect model). This behavior is explained by both $\fbase$ and $\fnew$ being systematically overconfident in their predictions prior to temperature scaling, so improved calibration does not affect the ranking in prediction confidence between the models for many samples. Appendix \ref{sec:nfr_confidence_distribution} shows the distribution of prediction confidence on negative flips, focusing on samples that $\fnew$ predicts with higher confidence than $\fbase$ and thus cannot be eliminated using Conf. Perfect churn reduction can thus occur only when using a score that sometimes chooses $\fbase$ even when $\max_{k} \phi(\fbase(x))_{k} < \max_{k} \phi(\fnew(x))_{k}$. This is precisely what the AvgConf score enables when used with Conf in \amc{} Combined. The synergy of Conf with OOD scores is examined in appendix \ref{sec:ood_scores} where none of the considered scores are as compatible as AvgConf.

\begin{table*}
\caption{Analysis of varying number of ensemble models $M$. Even impractically large values of $M$ fail to match the churn reduction ability of \amc.}
\begin{center}
\scalebox{0.8}{
\begin{tabular}{lccccccccc|c}\toprule
Dataset & \multicolumn{1}{c}{Cold} & \multicolumn{1}{p{1.2cm}}{\centering Ens \\ x3} & \multicolumn{1}{p{1cm}}{\centering Ens \\ x5} & \multicolumn{1}{p{1cm}}{\centering Ens \\ x7} & \multicolumn{1}{p{1cm}}{\centering Ens \\ x9} & \multicolumn{1}{p{1cm}}{\centering Ens \\ x11} & \multicolumn{1}{p{1cm}}{\centering Ens \\ x13} & \multicolumn{1}{p{1cm}}{\centering Ens \\ x15} & \multicolumn{1}{p{1cm}|}{\centering Ens \\ x17}  & \multicolumn{1}{p{1cm}}{\centering \amc{} \\ Distill}
\\\midrule
CIFAR10      &        6.449 &        3.401 &        2.652 &        2.176 &        1.926 &         1.727 &         1.613 &         1.515 &         1.452 &      \textbf{0.632} \\
FairFace     &       12.454 &        7.115 &        5.904 &        5.514 &        5.065 &         5.047 &         5.019 &         5.010 &         4.995 &      \textbf{2.635} \\
SVHN         &        6.415 &        2.892 &        2.173 &        1.759 &        1.518 &         1.392 &         1.304 &         1.207 &         1.110 &      \textbf{0.315} \\
\bottomrule
\end{tabular}
}
\end{center}
\label{table:ensemble_limit}
\end{table*}

\section{Discussion}
By showing that prediction instability during training may be infeasible to eliminate, and exists in part due to incompatibility between gradients, we motivated the need for a churn reduction method that does not rely on the new model having to match the predictions of the base model. We showed that \amc{} is capable of bypassing the stability-plasticity tradeoff by reverting to the base model when necessary for stability, and letting the new model learn unconstrained for maximum plasticity. While the performance advantage over distillation and ensembles is clear, the cost of inference increases by a factor of 2 since both the current model at time $t$, and the previous model $t-1$ are used for inference. In many cases this is a satisfiable requirement, and companies such as Tesla already use "shadow mode" which simultaneously runs both the old and new model version to compare their predictions for comprehensive evaluation \cite{Templeton2019-ps}. Crucially, \amc{} can be applied to models that are already deployed, whereas the churn reduction ability of ensemble requires the deployed model to be an ensemble. Preferring higher accuracy vs. lower churn is something that is application specific. For online ML APIs, user trust may not be as affected by churn as it would be in medical or financial applications. Therefore, the appropriateness of churn at cold accuracy as a metric is determined by the extent to which model utility depends on user trust. Overall, \amc{} gives ML practitioners an easy to implement method for maintaining user trust throughout model deployment. 

%% file: supp.tex
\appendix
\section{Training Details and Description of Baselines Evaluated}
\label{sec:training_details}
We use the Adam optimizer with a learning rate of 0.001 for both $\fbase$ and $\fnew$, and a batch size of 32. Data augmentation in the form of random horizontal flips and crops is used for the CIFAR10, CIFAR100, and STL10 datasets \citep{Shorten2019-ez}. Every set of experiments is run with 10 random seeds to obtain a reliable estimate of average accuracy and $\crel$. We note that since we are randomly splitting the original training sets into train/validation/update, the aim is not to reach SOTA accuracy for the respective tasks, but rather to investigate the churn reduction ability of the methods. Table \ref{table:data_splits} shows how the training set is split into train/validation/update for all datasets.

Experiments required several hundred GPU hours and were performed on a cluster of machines with NVIDIA T4 GPUs. PyTorch was used for all experiments which were logged using Weights and Biases.

\begin{table}[H]
\begin{center}
    \caption{Data splits for all datasets investigated. Note that in some cases a random subset of the original training data is split into train/validation/update to ensure that updating improves accuracy.}
    \begin{tabular}{lccc}\toprule
    Dataset           & Train & Validation & Update \\\midrule
    MNIST             & 20000 & 4000 & 10000  \\
    EMNIST            & 20000 & 4000 & 10000  \\
    KMNIST            & 20000 & 4000 & 10000  \\
    FashionMNIST      & 36000 & 12000 & 12000   \\
    SVHN              & 20000 & 4000 & 10000  \\
    CIFAR10           & 30000 & 10000 & 10000   \\
    CIFAR100          & 30000 & 10000 & 10000   \\
    STL10             & 3000 & 1000 & 1000   \\
    FairFace          & 15000 & 5000 & 15000   \\
    AG-News           & 10000 & 5000 & 10000   \\
    IMDB              & 5000 & 5000 & 15000   \\
    Adult             & 5000 & 1000 & 5000 \\
    HAR               & 3000 & 1000 & 3000 \\
    \bottomrule
    \end{tabular}
    \label{table:data_splits}
\end{center}
\end{table}

\paragraph{No Regularization (Cold)}
$\fbase$ is learned on the original training data as follows
\begin{align*}
    \thetabase = \argmin_{\theta} \frac{1}{|\dtrain|} \sum_{(x, y) \in \dtrain} \ell (\phi(f(x; \theta)), y)
\end{align*}

$\fnew$ is naively trained from a random initialization independently of $\fbase$ using the additional data $\data = \dtrain \cup \dupdate$
\begin{align*}
    \thetanew = \argmin_{\theta} \frac{1}{|\data|} \sum_{(x, y) \in \data} \ell (\phi(f(x; \theta)), y)
\end{align*}

This leads to an upper bound on churn that we aim to reduce.

\paragraph{Warm Start}
Instead of training $\fnew$ from a random initialization, it is initialized with the parameters of $\fbase$. This reduces churn by biasing the parameters of $\fnew$ to be closer to $\fbase$. This baseline is not always an option as on some datasets it results in worse accuracy than training from scratch using no regularization which is not a tradeoff we are willing to make. \citet{Ash2020-jn} have also observed that warm starts lead to worse generalization performance than retraining from scratch.

\paragraph{Distillation}
$\fnew$ is learned using a loss that is a combination of standard cross entropy loss on ground truth one-hot labels, and distillation using the predicted probabilities of $\fbase$ as targets
\begin{align*}
    \thetanew = \argmin_{\theta} \frac{1 - \alpha}{|\mathcal{D}|} \sum_{(x, y) \in \mathcal{D}} \ell (\phi(f(x; \theta)), y) + \frac{\alpha}{|\mathcal{D}|} \sum_{(x, y) \in \mathcal{D}} \ell (\phi(f(x; \theta)), \phi(f(x; \thetabase)))
\end{align*}

where lower values of $\alpha$ place more emphasis on learning independently from $\data$, and higher values encourage matching the predictions of $\fbase$.

\paragraph{Ensemble}
The base model is itself an ensemble  $\mathcal{F}_{\mathrm{base}} = \{\fbase^{(1)}, ..., \fbase^{(M)}\}$ where the parameters of each individual model $\thetabase^{(i)}, i \in [M]$ are learned as in the No Regularization baseline from different random initializations. 

The new model is also an ensemble  $\mathcal{F}_{\mathrm{new}} = \{\fnew^{(1)}, ..., \fnew^{(M)}\}$ where the parameters of each individual model $\thetanew^{(i)}, i \in [M]$ are learned as in the No Regularization baseline from different random initializations. 

Inference is done by averaging model logits via

\begin{align*}
    \mathcal{F}(x) = \frac{1}{M} \sum_{i=1}^{M} f^{(i)}(x)
\end{align*}

\paragraph{Focal Loss}
Similar to distillation except that the distillation loss target depends on the correctness of the base model. $\fnew$ is learned using the following loss
\begin{gather*}
    \thetanew = \argmin_{\theta} \frac{1 - \alpha}{|\mathcal{D}|} \sum_{(x, y) \in \mathcal{D}} \ell (\phi(f(x; \theta)), y) + \frac{\alpha}{|\mathcal{D}|} \sum_{(x, y) \in \mathcal{D}} \ell (\phi(f(x; \theta)), t(x, y)) \\
    t(x, y) = \begin{cases} 
      \phi(f(x; \thetabase)) & \mathrm{if} \; \; \sigma(f(x; \thetabase)) = y \\
      \epsilon e_{y} & \mathrm{otherwise} \\
   \end{cases}
\end{gather*}

where $e_{y}$ is a one-hot vector representation of the label $y$.

\section{More Performance Stats}
\label{sec:performance_stats}
\begin{figure}[H]
\centering
\includegraphics[width=0.99\linewidth]{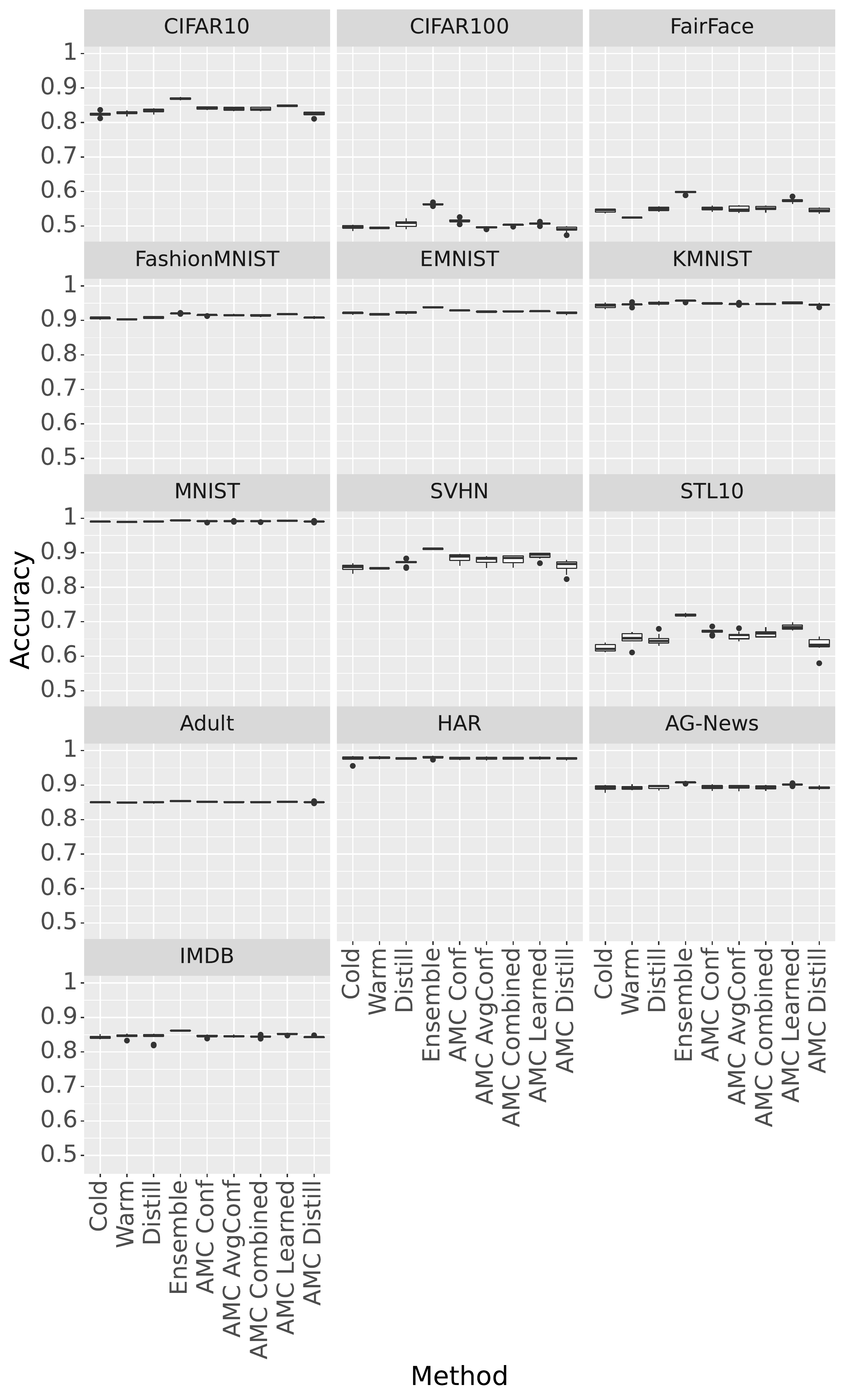}
\caption{Accuracy of methods on all datasets. Ensemble gives the best accuracy as expected, albeit at a large computational cost.}
\label{fig:accuracy_boxplot}
\end{figure}

\begin{figure}[H]
\centering
\includegraphics[width=0.99\linewidth]{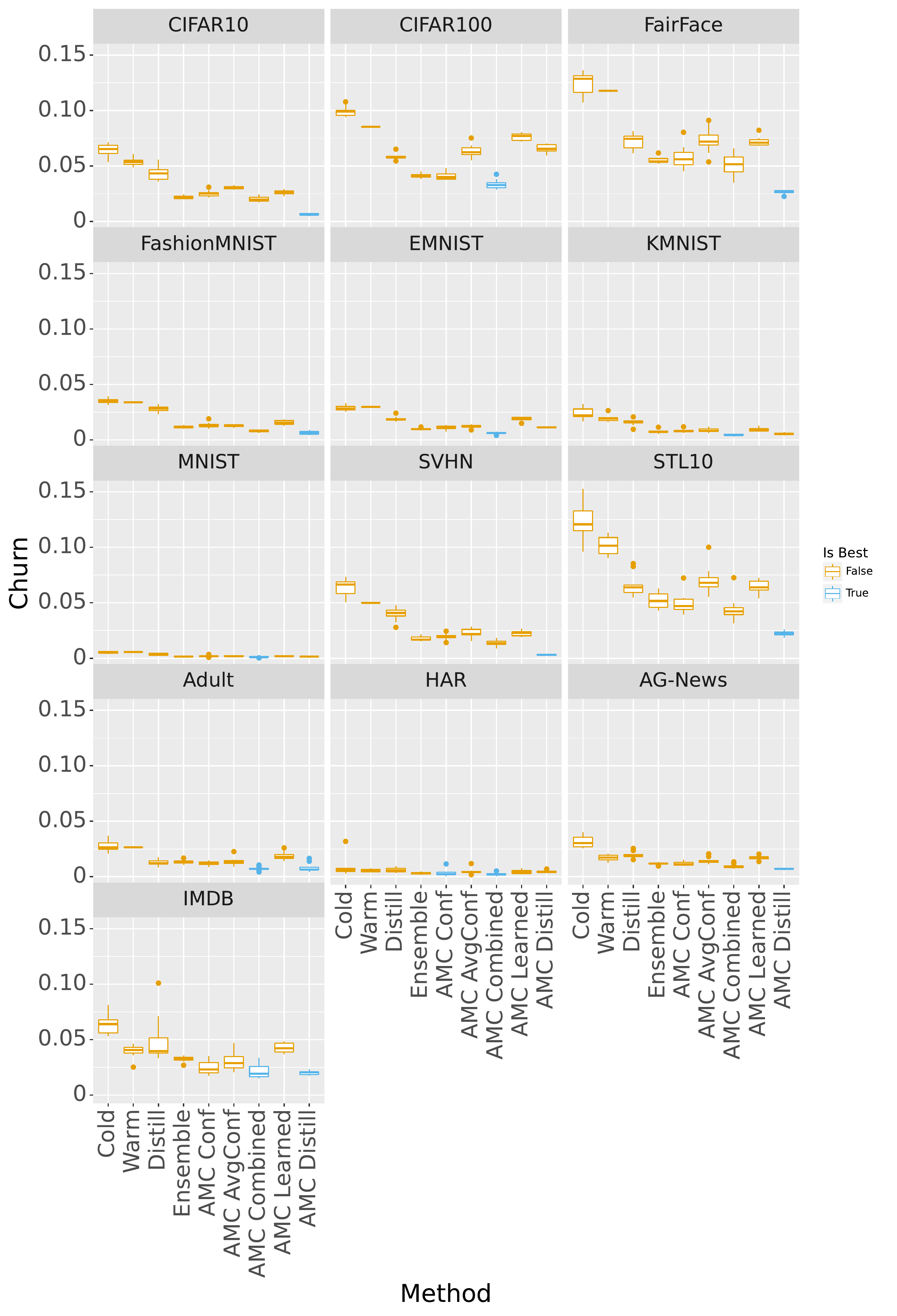}
\caption{Churn of methods on all datasets. \amc{} reduces churn the most.}
\label{fig:churn_boxplot}
\end{figure}

Table \ref{table:main_accuracy} shows the accuracy of the various churn reduction methods we consider. Ensemble achieves the best accuracy as expected. \amc{} matches/exceeds Cold accuracy on all datasets, such that there is no sacrifice in performance when using it. This result combined with its superior churn reduction ability confirms that it bypasses the stability-plasticity tradeoff. Figures \ref{fig:accuracy_boxplot} and \ref{fig:churn_boxplot} show ranges of accuracy and churn values respectively via boxplots. It is important to note that $\fbase$ for Ensemble is an ensemble itself, so it has higher initial accuracy compared to all other methods prior to an update using extra data. This makes it difficult to directly compare the accuracy of $\fnew$ between Ensemble and the remaining methods since it is the only method we consider which also affects the training/inference of $\fbase$.

\begin{table*}[h!]
\caption{Accuracy of churn reduction methods. \amc{} matches/exceeds Cold accuracy on all datasets as do ensembles, though \amc{} is much more efficient.}
\begin{center}
\scalebox{0.75}{
\begin{tabular}{lcccc|ccccc}\toprule
Dataset & \multicolumn{1}{c}{Cold} & \multicolumn{1}{p{1.2cm}}{\centering Warm \\ Start} & \multicolumn{1}{c}{Distill} & \multicolumn{1}{c|}{Ensemble} & \multicolumn{1}{p{1.2cm}}{\centering \amc{} \\ Conf} & \multicolumn{1}{p{1.5cm}}{\centering \amc{} \\ Avg Conf} & \multicolumn{1}{p{1.2cm}}{\centering \amc{} \\ Combined} & \multicolumn{1}{p{1.5cm}}{\centering \amc{} \\ Learned}  & \multicolumn{1}{p{1.5cm}}{\centering \amc{} \\ Distill}
\\\midrule
CIFAR10      &      82.352 &     82.792 &       83.372 &   86.904 &   84.130 &       83.950 &       83.907 &      84.839 &                82.484 \\
CIFAR100     &      49.622 &     49.448 &       50.613 &   56.319 &   51.524 &       49.574 &       50.334 &      50.655 &                48.948 \\
FairFace     &      54.473 &     52.400 &       54.894 &   59.786 &   55.019 &       54.896 &       55.056 &      57.415 &                54.514 \\
FashionMNIST &      90.689 &     90.322 &       90.869 &   92.001 &   91.509 &       91.516 &       91.426 &      91.760 &                90.804 \\
EMNIST       &      98.990 &     98.937 &       99.036 &   99.363 &   99.126 &       99.139 &       99.130 &      99.230 &                99.031 \\
KMNIST       &      92.153 &     91.757 &       92.268 &   93.779 &   92.854 &       92.540 &       92.617 &      92.678 &                92.184 \\
MNIST        &      94.245 &     94.564 &       94.963 &   95.664 &   94.923 &       94.717 &       94.765 &      95.105 &                94.488 \\
SVHN         &      85.645 &     85.487 &       87.171 &   91.167 &   88.456 &       87.849 &       88.024 &      89.146 &                86.099 \\
STL10        &      62.387 &     65.154 &       64.762 &   71.940 &   67.174 &       65.935 &       66.558 &      68.504 &                63.328 \\
\midrule
Adult        &      85.006 &     84.961 &       85.022 &   85.384 &   85.180 &       85.074 &       85.066 &      85.180 &                85.045 \\
HAR          &      97.673 &     97.954 &       97.745 &   98.078 &   97.761 &       97.778 &       97.766 &      97.830 &                97.693 \\
\midrule
AG-News      &      89.192 &     89.233 &       89.447 &   90.750 &   89.455 &       89.405 &       89.330 &      90.183 &                89.217 \\
IMDB         &      84.266 &     84.630 &       84.359 &   86.144 &   84.584 &       84.574 &       84.444 &      85.202 &                84.406 \\
\bottomrule
\end{tabular}
}
\end{center}
\label{table:main_accuracy}
\end{table*}

\newpage

\newpage

\section{Best Learner for Learned Model}
\label{sec:best_learned_model}
Tables \ref{table:learned_churn} and \ref{table:learned_accuracy} compares the effectiveness of logistic regression, random forest, and gradient boosting models as a trainable meta-model $\psi$. Logistic regression does the best on all datasets except CIFAR100 suggesting there is no advantage to learning a non-linear function that combines the outputs of $\fnew$ and $\fbase$. This observation is in accordance with the stacking literature where the meta-model is usually naive Bayes or logistic regression.

\paragraph{Logistic Regression Hyperparameters}
We search over different values of the L2-regulariazation parameter $\lambda \in [\mathrm{1e-4}, \mathrm{1e-3}, \mathrm{1e-2}, \mathrm{1e-1}]$

\paragraph{Random Forest Hyperparameters}
We consider random forests having $[50, 100, 250, 500]$ trees.

\paragraph{Gradient Boosting}
We consider $[50, 100, 250, 500]$ stages of gradient boosting.

\begin{table}[ht]
\caption{Churn of logistic regression, random forest, and gradient boosting models for\amc{} learned}
\begin{center}
\begin{tabular}{lccc}\toprule
\multicolumn{1}{c}{Dataset} & \multicolumn{1}{c}{Logistic Regression} & \multicolumn{1}{c}{Random Forest} & \multicolumn{1}{c}{Gradient Boosting} \\\midrule
MNIST           &  0.202 & 0.271 &  0.417 \\   
EMNIST          &  1.900 & 2.029 &  2.784 \\   
KMNIST          &  0.946 & 1.293 &  1.859 \\   
FashionMNIST    &  1.573 & 1.795 &  1.914 \\   
SVHN            &  2.241 & 2.447 &  3.171 \\   
CIFAR10         &  2.618 & 3.102 &  3.341 \\   
CIFAR100        & 11.714 & 7.622 & 15.310 \\   
STL10           &  6.448 & 7.005 &  8.411 \\   
\bottomrule
\end{tabular}
\end{center}
\label{table:learned_churn}
\end{table}

\begin{table}[ht]
\caption{Accuracy of logistic regression, random forest, and gradient boosting models for Learned Model \amc{}}
\begin{center}
\begin{tabular}{lccc}\toprule
\multicolumn{1}{c}{Dataset} & \multicolumn{1}{c}{Logistic Regression} & \multicolumn{1}{c}{Random Forest} & \multicolumn{1}{c}{Gradient Boosting} \\\midrule
MNIST           &    99.230 &   99.190 &   99.063 \\ 
EMNIST          &    92.678 &   92.547 &   91.836 \\ 
KMNIST          &    95.105 &   94.715 &   94.161 \\ 
FashionMNIST    &    91.760 &   91.547 &   91.449 \\ 
SVHN            &    89.146 &   88.712 &   88.044 \\ 
CIFAR10         &    84.839 &   84.512 &   84.281 \\ 
CIFAR100        &    46.067 &   50.655 &   40.694 \\ 
STL10           &    68.504 &   67.349 &   65.855 \\ 
\bottomrule
\end{tabular}
\end{center}
\label{table:learned_accuracy}
\end{table}

\section{\amc{} Distill Details}
\label{sec:amc_distill_details}
We search over two options for the best \amc{} Distill architecture:
\begin{itemize}
    \item Fully connected network with no hidden layers
    \item Single layer fully connected network with 100 hidden units and ReLU hidden activation
\end{itemize}

We find that architectures with higher capacity are too prone to overfitting for the task of generating meta-predictions, hence the limited capacity of the considered architectures. Adam with a learning rate of 0.001 along with a batch size of 32 and early stopping with a patience of 5 epochs is used to learn $\psi$. Similar to the standard distillation baseline, we search over $\alpha \in \{ 0.1, 0.2, ..., 0.9\}$.

\section{OOD Detection Scores}
\label{sec:ood_scores}
Given the strong churn reduction baseline that \amc{} Conf provides, and the improved results when combined with AvgConf, we investigate if OOD detection scores are even more effective at choosing the model most likely to be correct.

\paragraph{Entropy:}
An alternative to prediction confidence is entropy which captures not only the probability for the predicted class, but also the uncertainty among the remaining classes. We use the negative entropy as the score for \amc{}
\begin{align*}
    H(x; f) &= - \sum_{c=1}^{k} \phi(f(x))_{c} \mathrm{log} (\phi(f(x))_{c}) \\
    s(f, x) &= - H(x; f)
\end{align*}

\paragraph{Energy:}
\citet{Liu2020-nj} show that the Helmholtz free-energy of a neural network's predictions can be used to distinguish between in-distribution (ID) and out-of-distribution (OOD) samples. This could also be useful when choosing between which model to use for a given output. We use the negative energy as the score for \amc{}:
\begin{align*}
    E(x; f) &= - \mathrm{log} \sum_{c=1}^{k} \mathrm{e}^{(f_{c}(x))} \\
    s(f, x) &= - E(x; f)
\end{align*}

\paragraph{KL-Div:}
The KL-divergence between a model's output probabilities and the uniform distribution has been observed to be larger for ID data compared to OOD data \citep{Huang2021-rm}. While correlated with prediction confidence, this captures uniformity among remaining classes and prefers that the remaining probability is concentrated among a few classes. We thus choose the model that has the highest such KL-divergence for a given sample
\begin{gather*}
    \mathbf{u} = \left[\frac{1}{k}, ..., \frac{1}{k} \right] \in \mathbb{R}^{k} \; \; \; \; \; \; \; \; s(f, x) = \dkl(\mathbf{u} || f(x)) 
\end{gather*}

\paragraph{Gradnorm:}
\citep{Huang2021-rm} showed that the norm of the gradient of the above KL-Div is an even more effective OOD score. 

\begin{gather*}
    \mathbf{u} = \left[\frac{1}{k}, ..., \frac{1}{k} \right] \in \mathbb{R}^{k} \; \; \; \; \; \; \; \; s(f, x) = || \nabla_{\theta} \dkl(\mathbf{u} || f(x)) ||_{1}
\end{gather*}

\begin{figure}[H]
\hfill
\begin{subfigure}[t]{0.24\textwidth}
\centering
\includegraphics[width=\textwidth]{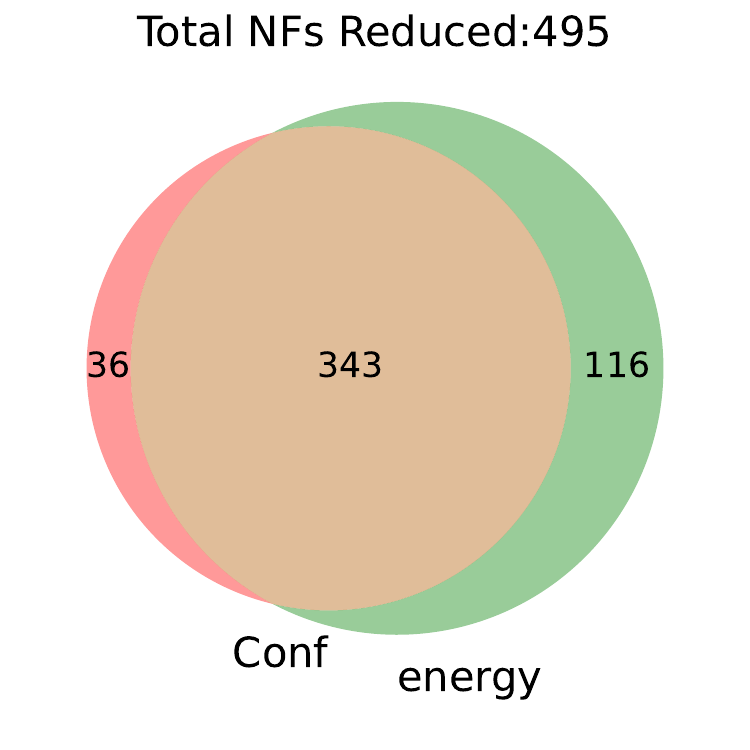}
\caption{Energy NFs}
\end{subfigure}
\begin{subfigure}[t]{0.24\textwidth}
\centering
\includegraphics[width=\textwidth]{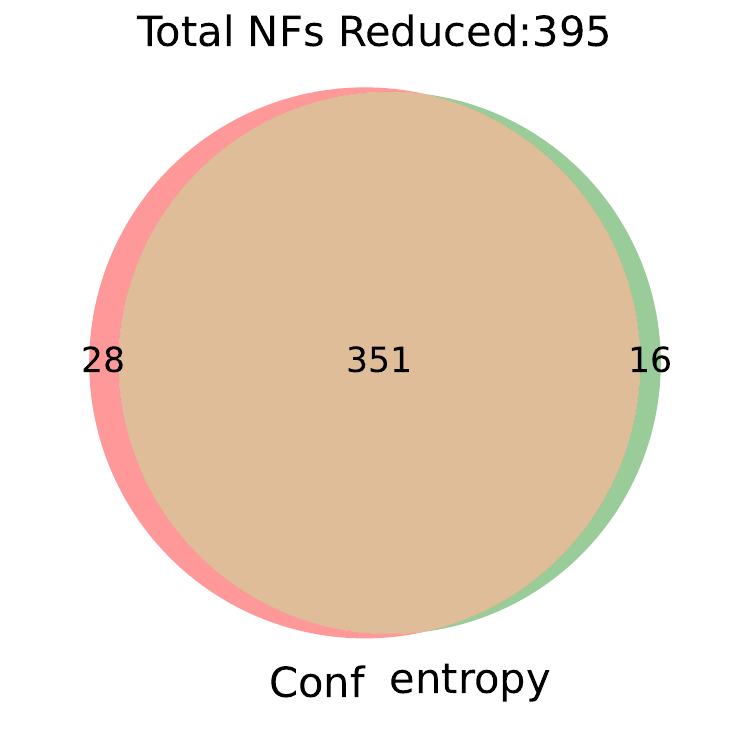}
\caption{Entropy NFs}
\end{subfigure}
\begin{subfigure}[t]{0.24\textwidth}
\centering
\includegraphics[width=\textwidth]{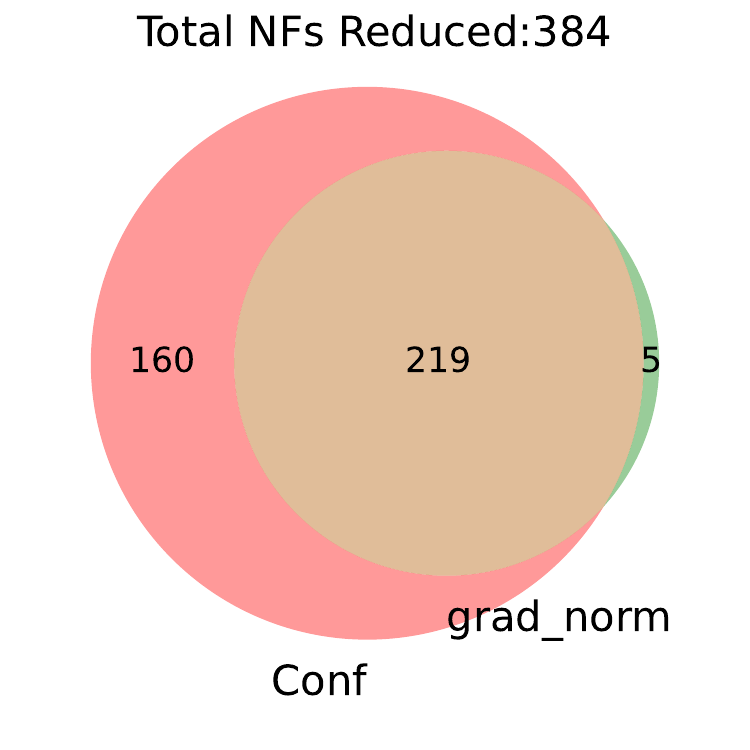}
\caption{GradNorm NFs}
\end{subfigure}
\begin{subfigure}[t]{0.24\textwidth}
\centering
\includegraphics[width=\textwidth]{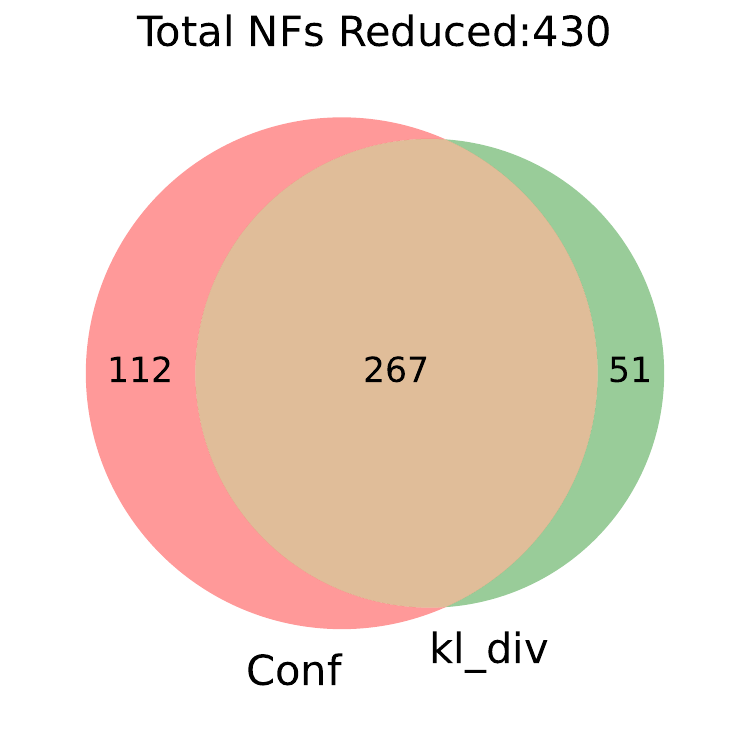}
\caption{KL-Div NFs}
\end{subfigure}
\vskip\baselineskip
\begin{subfigure}[t]{0.24\textwidth}
\centering
\includegraphics[width=\textwidth]{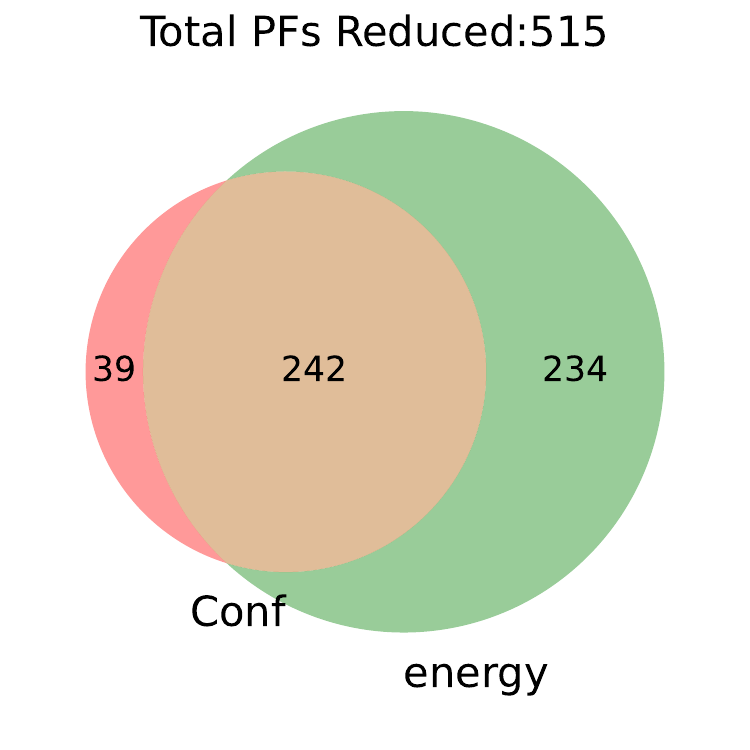}
\caption{Energy PFs}
\end{subfigure}
\begin{subfigure}[t]{0.24\textwidth}
\centering
\includegraphics[width=\textwidth]{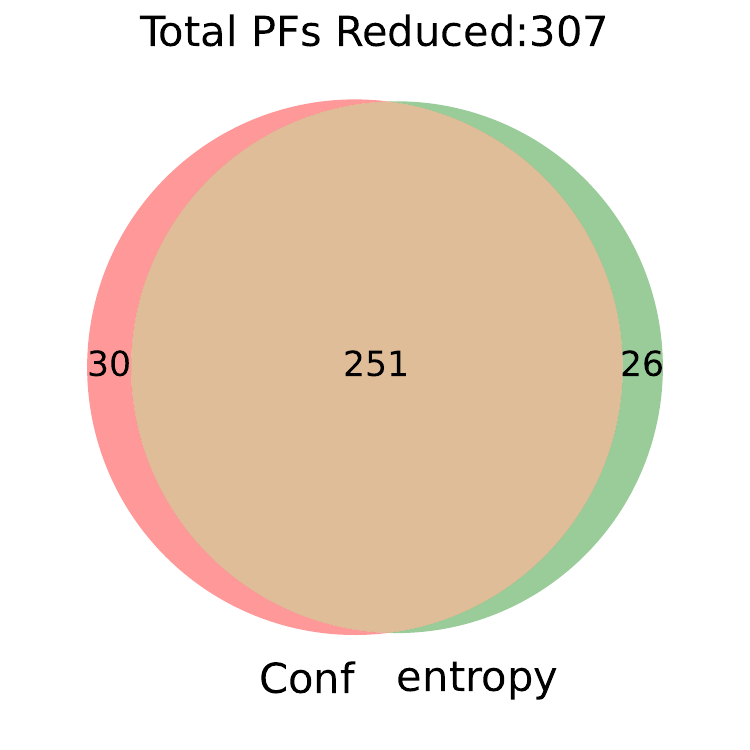}
\caption{Entropy PFs}
\end{subfigure}
\begin{subfigure}[t]{0.24\textwidth}
\centering
\includegraphics[width=\textwidth]{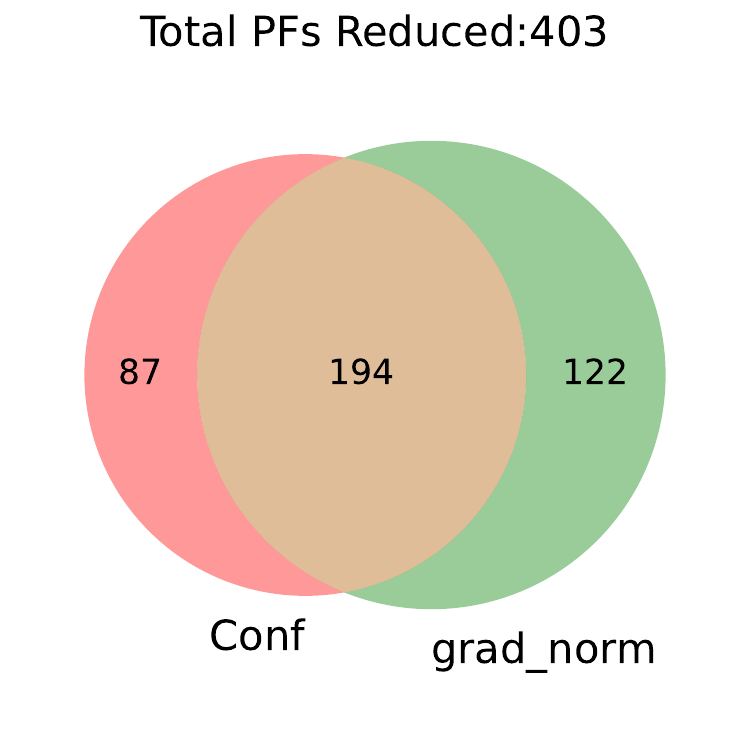}
\caption{GradNorm PFs}
\end{subfigure}
\begin{subfigure}[t]{0.24\textwidth}
\centering
\includegraphics[width=\textwidth]{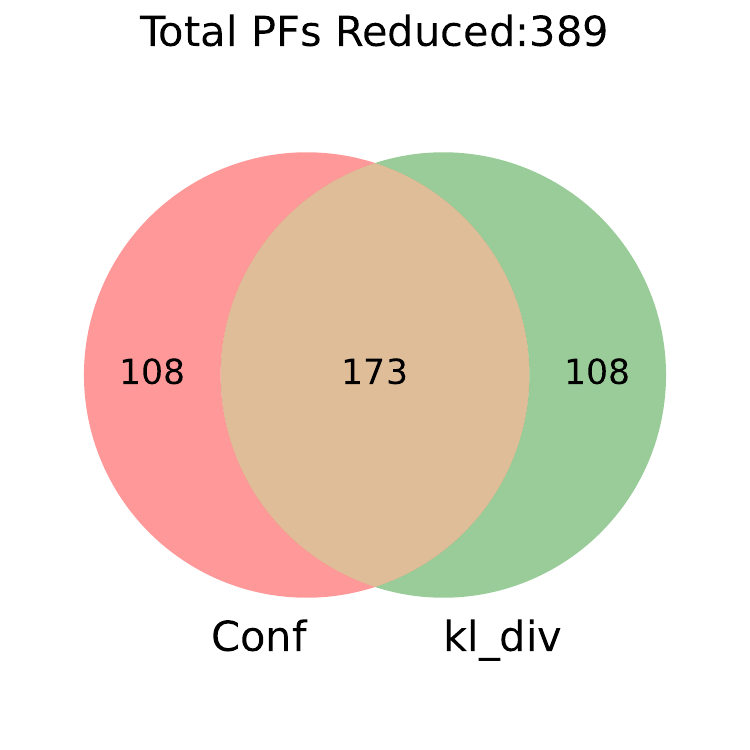}
\caption{KL-Div PFs}
\end{subfigure}
\caption{Overlap in NFs and PFs reduced for between various scores and Conf on CIFAR10.}
\label{fig:cifar10_score_overlap}
\end{figure}

Table \ref{table:other_scores_churn} and \ref{table:other_scores_accuracy} show that on a single score-basis, Conf is the most effective score for both reducing churn and maintaining accuracy across nearly all datasets. This is surprising since Entropy or KL-Div include information from all model outputs, not only the predicted class. It is possible that although these alternative OOD scores are not effective on their own, they might complement Conf well similar to Avgconf. Figure \ref{fig:cifar10_score_overlap} investigates this possibility by examining the overlap between Conf and other scores for both positive and negative flips. Entropy and Conf are very similar having high overlap in both the NFs and PFs that they reduce, so they do not make  an effective combination. Energy has less overlap with Conf for negative flips, but for these additional NFs removed it removes twice the number of positive flips which results in a significant drop in accuracy such that this would violate the requirement of matching cold accuracy. KL-Div exhibits similar behavior, and the set of NFs reduced by GradNorm is only 5 samples away from being a strict subset of Conf. These results suggest that OOD detection scores are not suited for choosing between two models, and this is true even when the scores for the base and new model are scaled to have the same range. 

\begin{table}[H]
\caption{Churn of various other scores compared to Conf. amc{} using Conf is best on almost all datasets except KMNIST.}
\begin{center}
\begin{tabular}{lccccc}\toprule
\multicolumn{1}{c}{Dataset} & \multicolumn{1}{c}{\conf{}} & \multicolumn{1}{c}{KL-Div} & \multicolumn{1}{c}{Entropy} & \multicolumn{1}{c}{Energy} & \multicolumn{1}{c}{Gradnorm} \\\midrule
MNIST           &  0.194 &   0.195 &  0.188 &  0.254 &    0.615 \\
EMNIST          &  1.091 &   1.101 &  1.196 &  1.373 &    2.824 \\
KMNIST          &  0.807 &   0.831 &  0.718 &  1.485 &    2.467 \\
FashionMNIST    &  1.312 &   1.353 &  1.367 &  1.901 &    3.642 \\
SVHN            &  1.945 &   2.111 &  2.041 &  2.443 &    2.805 \\
CIFAR10         &  2.526 &   2.711 &  2.667 &  3.371 &    4.841 \\
CIFAR100        &  4.092 &   4.636 &    NaN &  5.473 &    5.704 \\
STL10           &  4.971 &   5.298 &  5.837 &  7.032 &    5.950 \\
\bottomrule
\end{tabular}
\end{center}
\label{table:other_scores_churn}
\end{table}

\begin{table}[H]
\caption{Accuracy of various other scores compared to Conf. amc{} using Conf is best on almost all datasets. The fact that Conf }
\begin{center}
\begin{tabular}{lccccc}\toprule
\multicolumn{1}{c}{Dataset} & \multicolumn{1}{c}{Conf} & \multicolumn{1}{c}{Entropy} & \multicolumn{1}{c}{Energy} & \multicolumn{1}{c}{KL-Div} & \multicolumn{1}{c}{Gradnorm} \\\midrule
MNIST           &     99.126 &   99.124 &   99.025 &   99.042 &   98.915 \\
EMNIST          &     92.854 &   92.837 &   92.271 &   92.223 &   91.928 \\
KMNIST          &     94.923 &   94.888 &   94.489 &   94.357 &   94.072 \\
FashionMNIST    &     91.509 &   91.503 &   90.991 &   90.949 &   90.601 \\
SVHN            &     88.456 &   88.254 &   87.057 &   87.563 &   87.561 \\
CIFAR10         &     84.130 &   84.051 &   83.294 &   83.316 &   82.877 \\
CIFAR100        &     51.524 &   51.113 &   49.712 &   50.066 &   50.365 \\
STL10           &     67.174 &   67.068 &   65.606 &   65.608 &   65.631 \\
\bottomrule
\end{tabular}
\end{center}
\label{table:other_scores_accuracy}
\end{table}

\begin{figure}
\hfill
\begin{subfigure}[t]{0.24\textwidth}
\centering
\includegraphics[width=\textwidth]{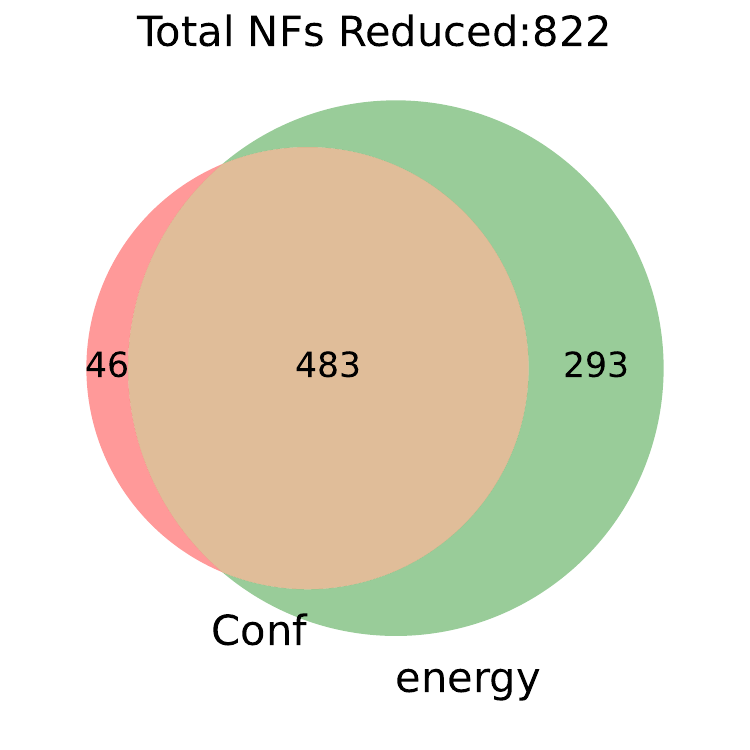}
\caption{Energy NFs}
\end{subfigure}
\begin{subfigure}[t]{0.24\textwidth}
\centering
\includegraphics[width=\textwidth]{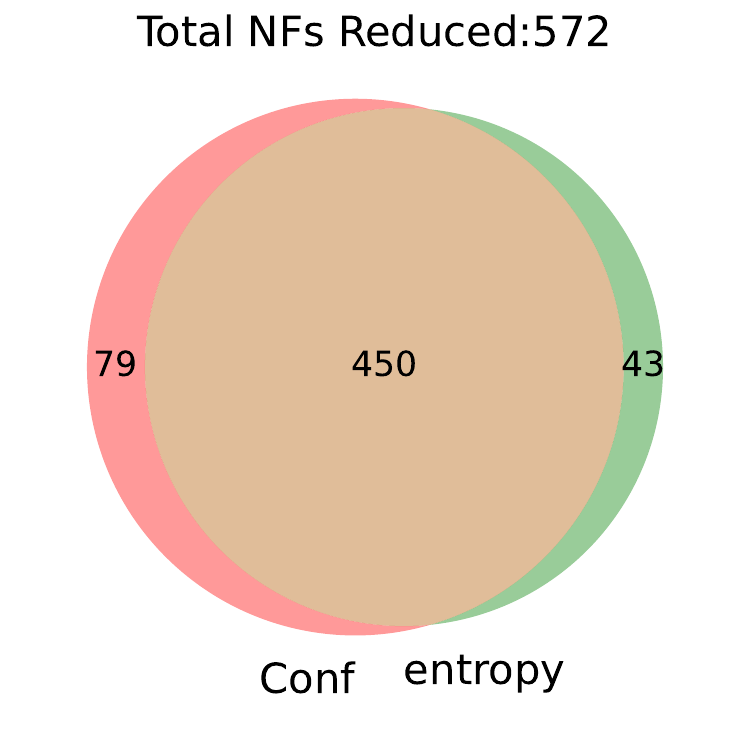}
\caption{Entropy NFs}
\end{subfigure}
\begin{subfigure}[t]{0.24\textwidth}
\centering
\includegraphics[width=\textwidth]{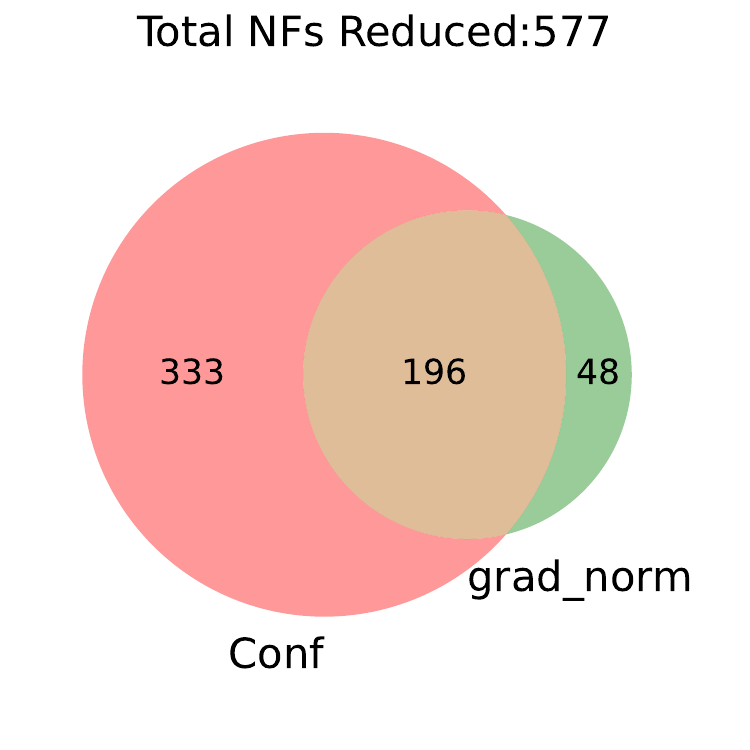}
\caption{GradNorm NFs}
\end{subfigure}
\begin{subfigure}[t]{0.24\textwidth}
\centering
\includegraphics[width=\textwidth]{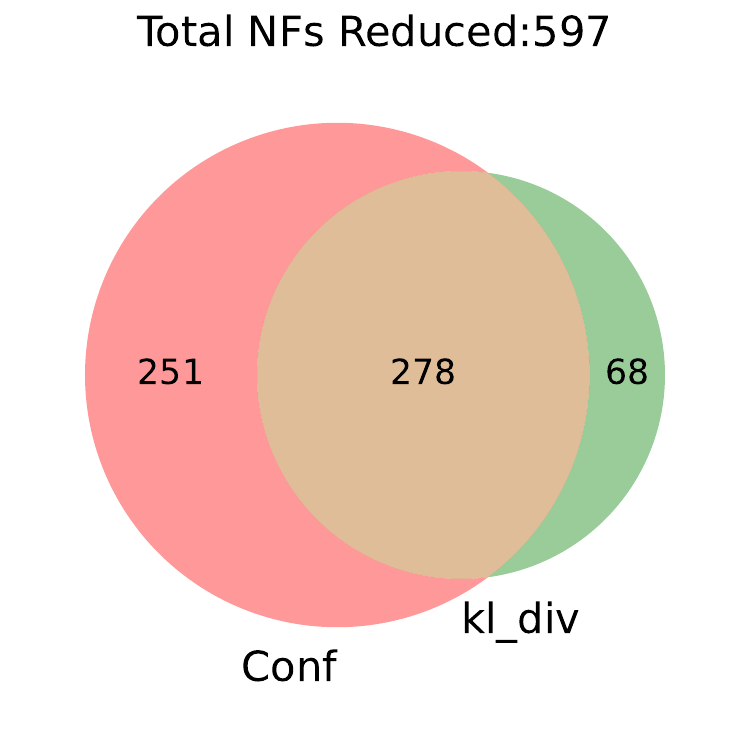}
\caption{KL-Div NFs}
\end{subfigure}
\vskip\baselineskip
\begin{subfigure}[t]{0.24\textwidth}
\centering
\includegraphics[width=\textwidth]{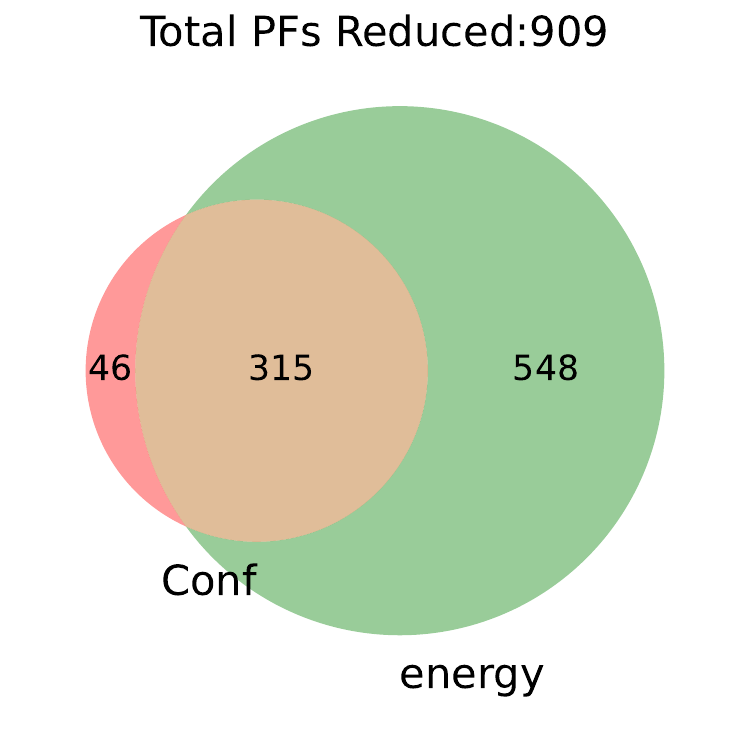}
\caption{Energy PFs}
\end{subfigure}
\begin{subfigure}[t]{0.24\textwidth}
\centering
\includegraphics[width=\textwidth]{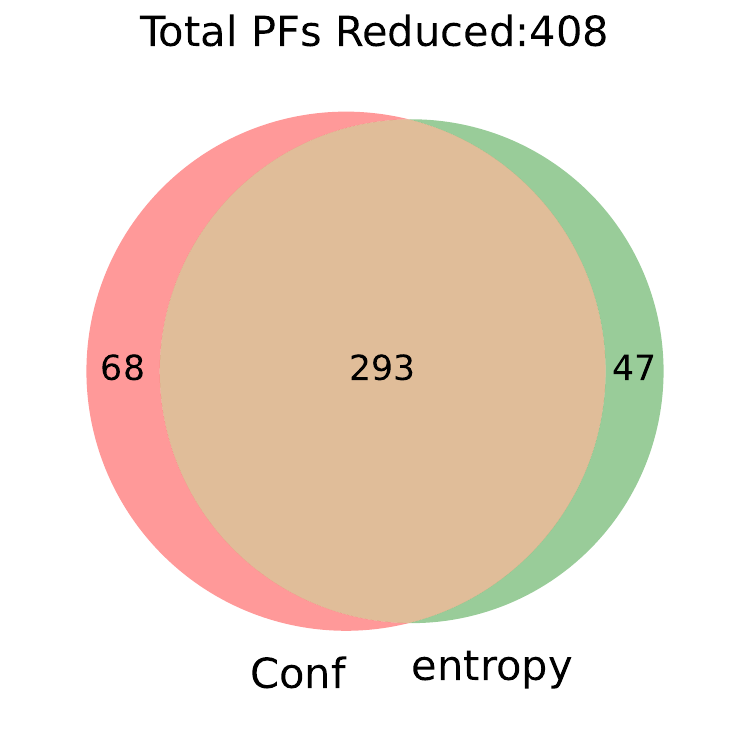}
\caption{Entropy PFs}
\end{subfigure}
\begin{subfigure}[t]{0.24\textwidth}
\centering
\includegraphics[width=\textwidth]{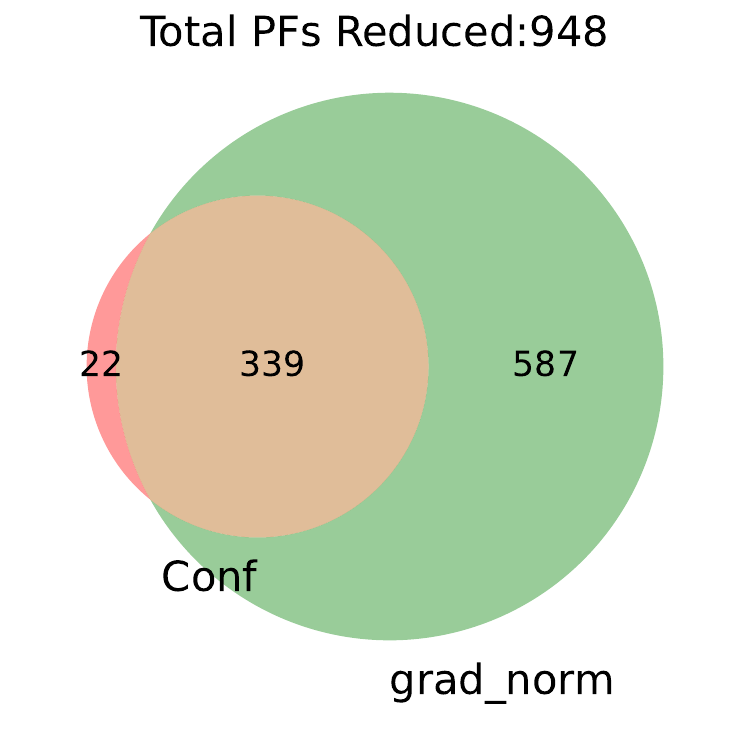}
\caption{GradNorm PFs}
\end{subfigure}
\begin{subfigure}[t]{0.24\textwidth}
\centering
\includegraphics[width=\textwidth]{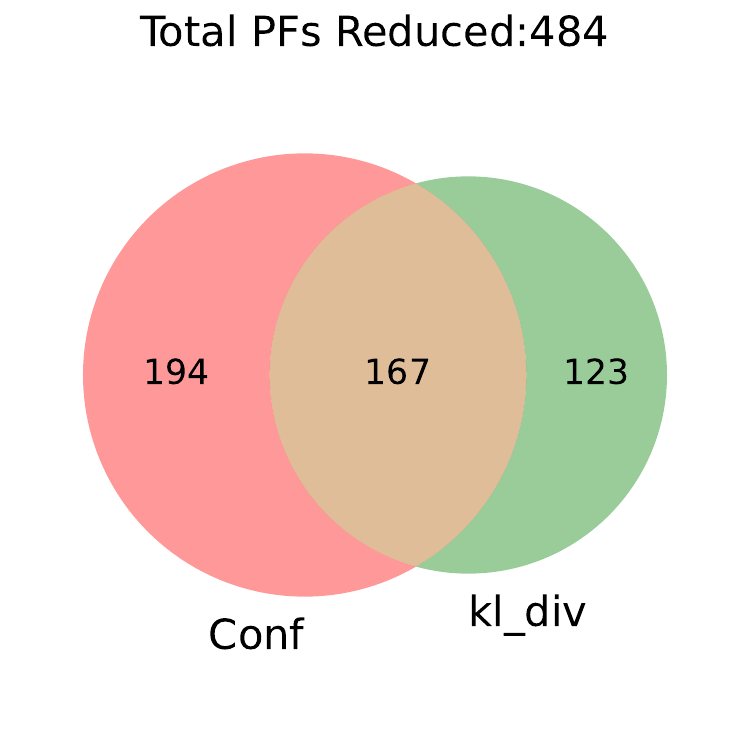}
\caption{KL-Div PFs}
\end{subfigure}
\caption{Overlap in NFs and PFs reduced for between various scores and Conf on CIFAR100.}
\label{fig:cifar100_score_overlap}
\end{figure}

\begin{figure}
\hfill
\begin{subfigure}[t]{0.24\textwidth}
\centering
\includegraphics[width=\textwidth]{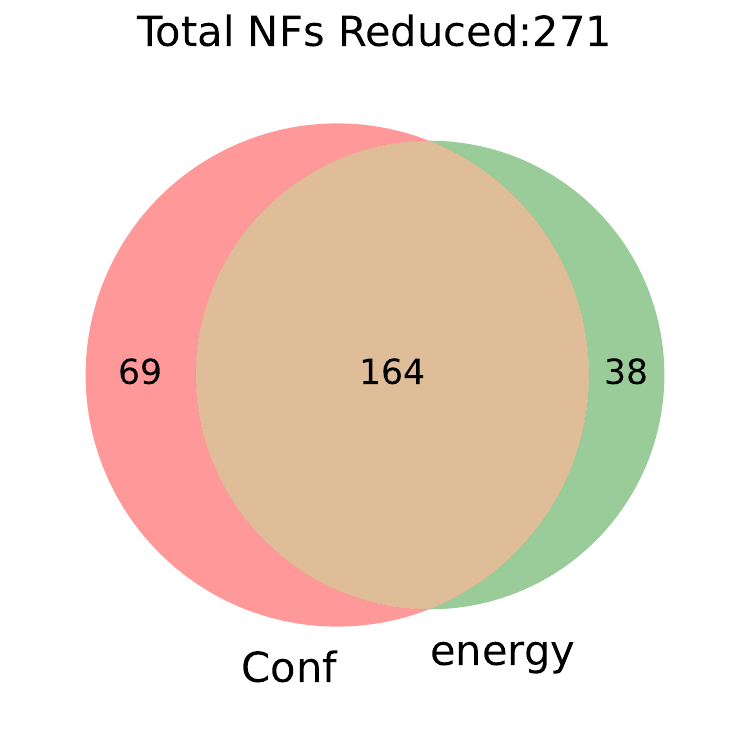}
\caption{Energy NFs}
\end{subfigure}
\begin{subfigure}[t]{0.24\textwidth}
\centering
\includegraphics[width=\textwidth]{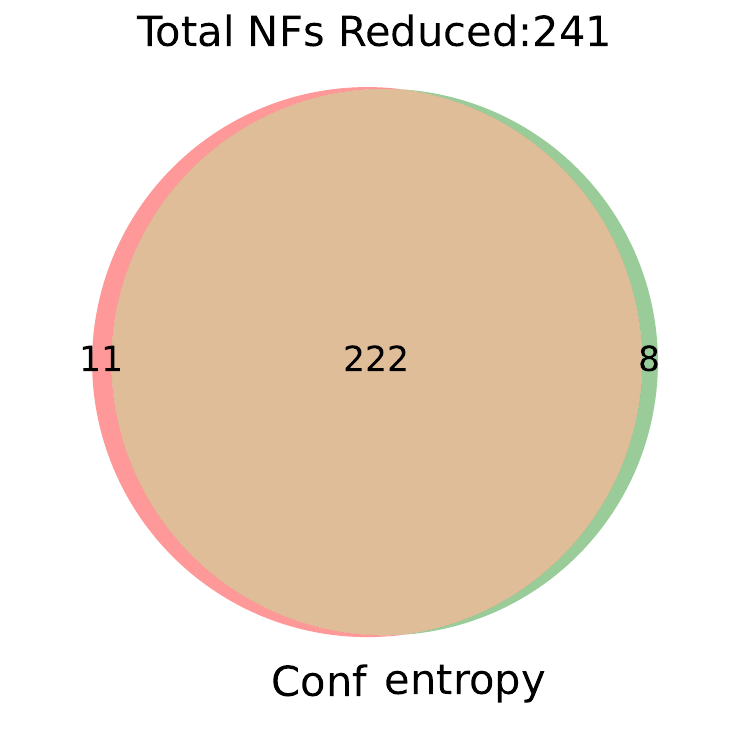}
\caption{Entropy NFs}
\end{subfigure}
\begin{subfigure}[t]{0.24\textwidth}
\centering
\includegraphics[width=\textwidth]{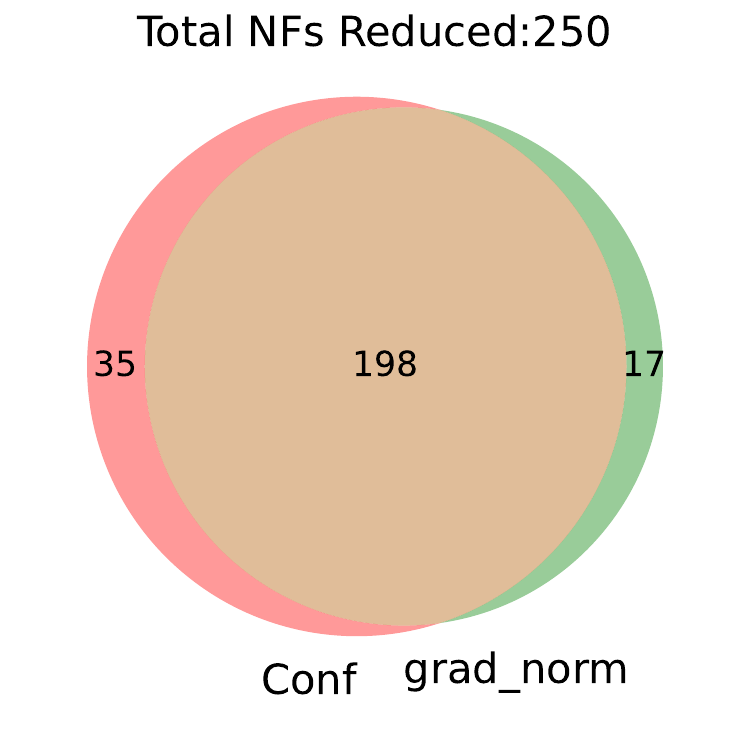}
\caption{GradNorm NFs}
\end{subfigure}
\begin{subfigure}[t]{0.24\textwidth}
\centering
\includegraphics[width=\textwidth]{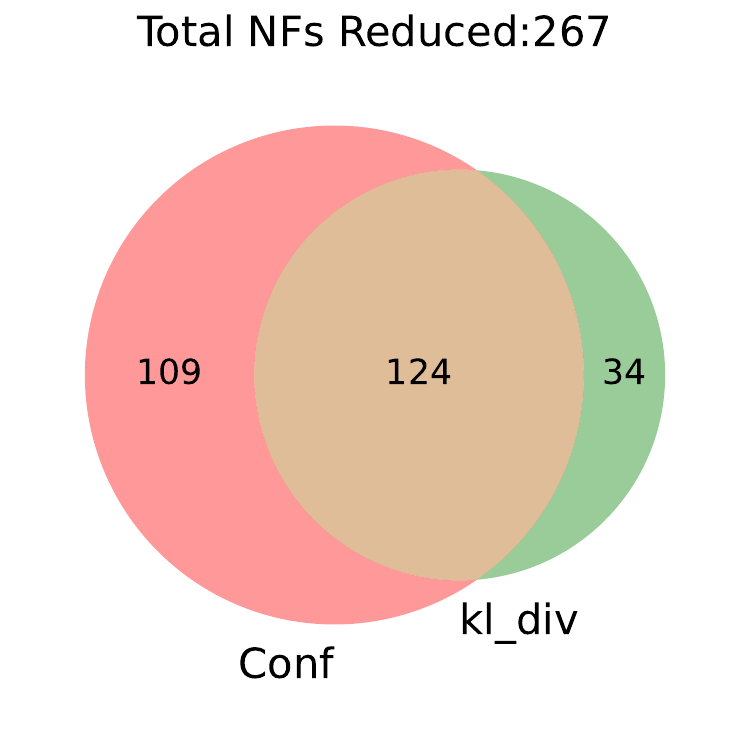}
\caption{KL-Div NFs}
\end{subfigure}
\vskip\baselineskip
\begin{subfigure}[t]{0.24\textwidth}
\centering
\includegraphics[width=\textwidth]{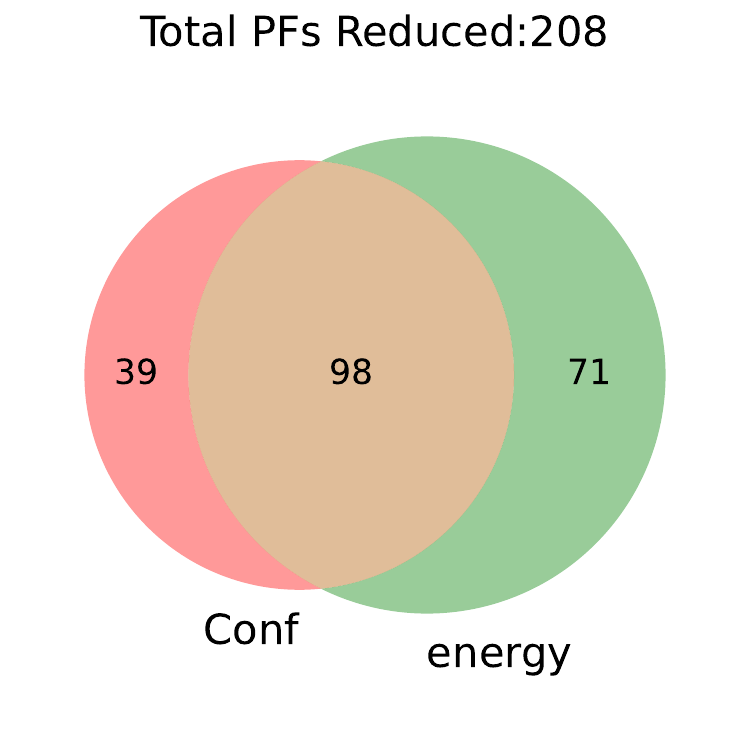}
\caption{Energy PFs}
\end{subfigure}
\begin{subfigure}[t]{0.24\textwidth}
\centering
\includegraphics[width=\textwidth]{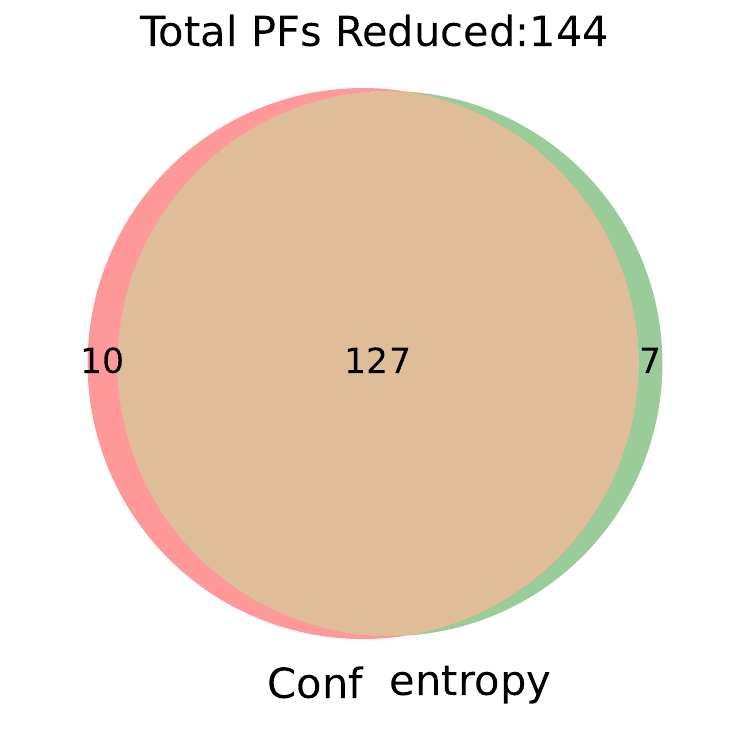}
\caption{Entropy PFs}
\end{subfigure}
\begin{subfigure}[t]{0.24\textwidth}
\centering
\includegraphics[width=\textwidth]{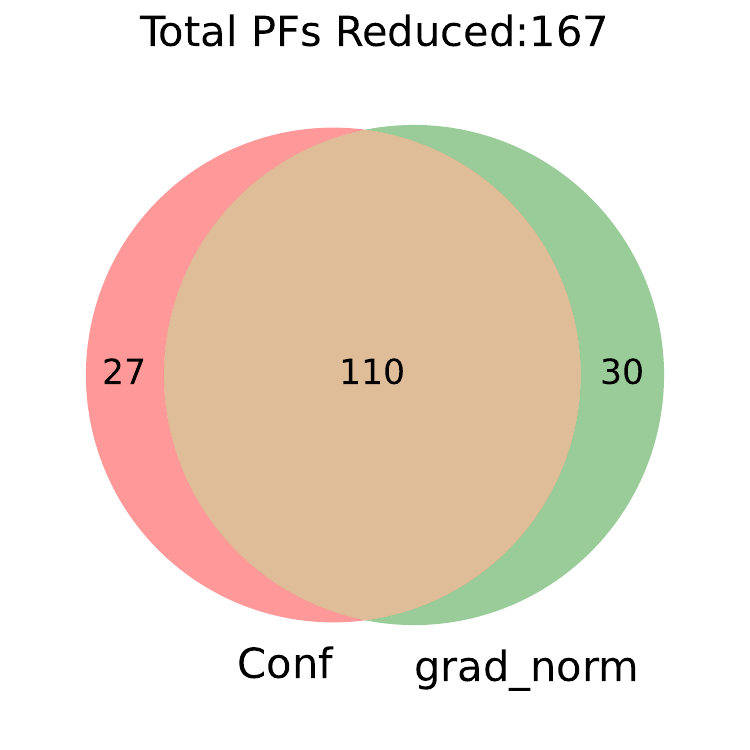}
\caption{GradNorm PFs}
\end{subfigure}
\begin{subfigure}[t]{0.24\textwidth}
\centering
\includegraphics[width=\textwidth]{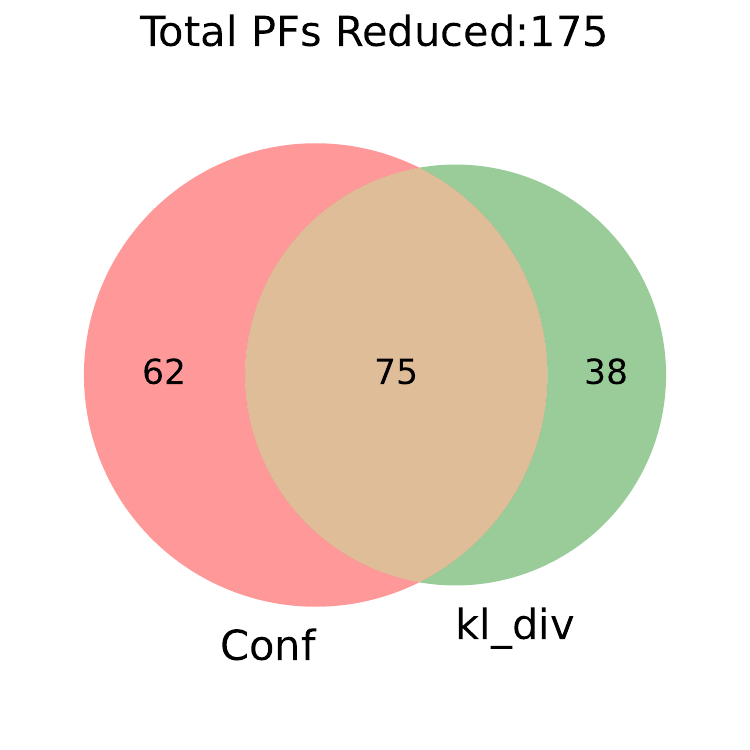}
\caption{KL-Div PFs}
\end{subfigure}
\caption{Overlap in NFs and PFs reduced for between various scores and Conf on FashionMNIST.}
\label{fig:fashion_mnist_score_overlap}
\end{figure}

\section{Additional Calibration Figures}
\label{sec:additional_calibration_figures}

\subsubsection*{SVHN}
\begin{figure}[H]
\centering
\begin{subfigure}[t]{0.24\textwidth}
\centering
\includegraphics[width=0.99\linewidth]{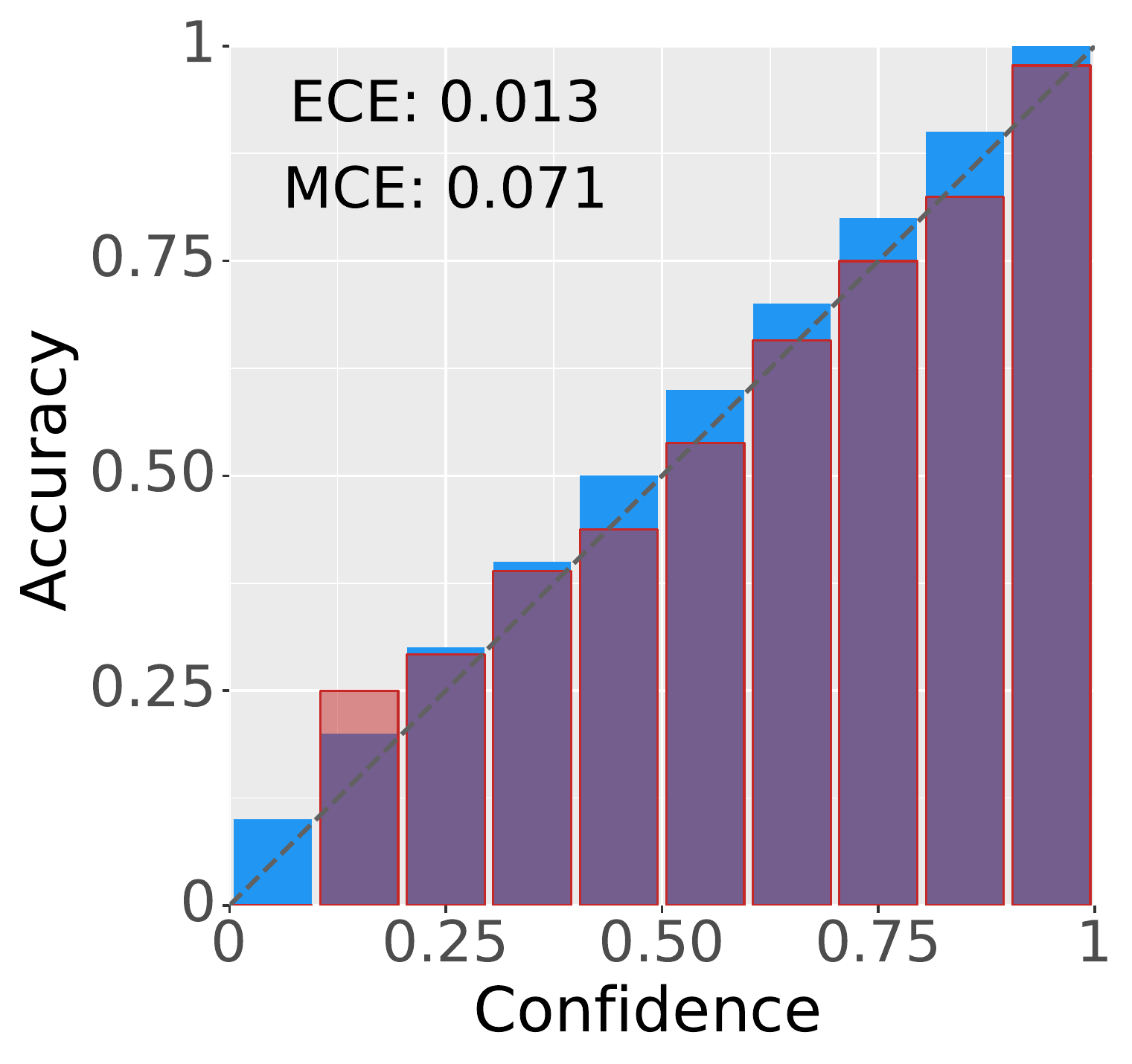}
\caption{$\fbase$ Uncalibrated}
\end{subfigure}
\hfill
\begin{subfigure}[t]{0.24\textwidth}
\centering
\includegraphics[width=0.99\linewidth]{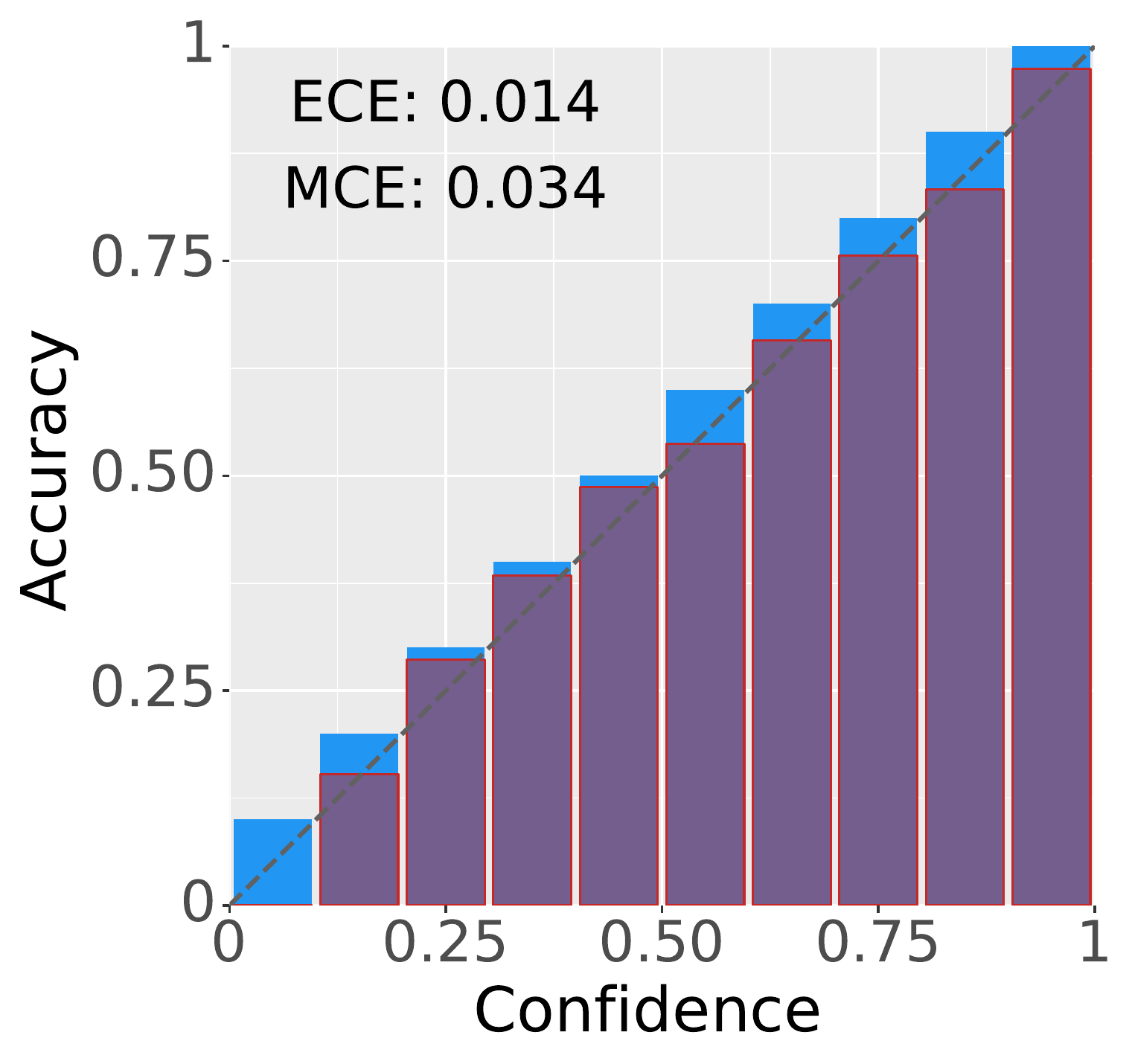}
\caption{$\fnew$ Uncalibrated}
\end{subfigure}
\hfill
\begin{subfigure}[t]{0.24\textwidth}
\centering
\includegraphics[width=0.99\linewidth]{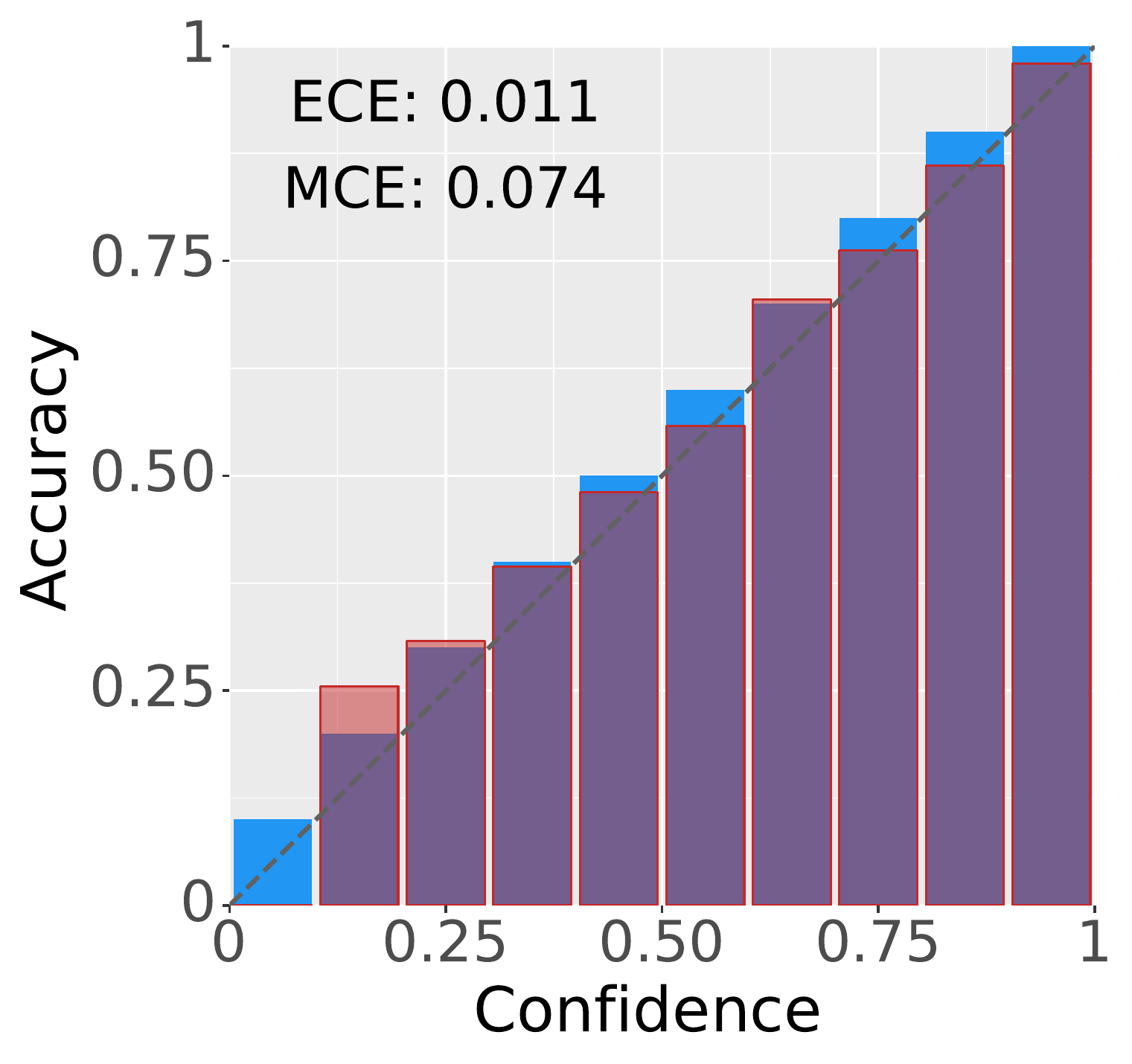}
\caption{$\fbase$ Calibrated}
\end{subfigure}
\hfill
\begin{subfigure}[t]{0.24\textwidth}
\centering
\includegraphics[width=0.99\linewidth]{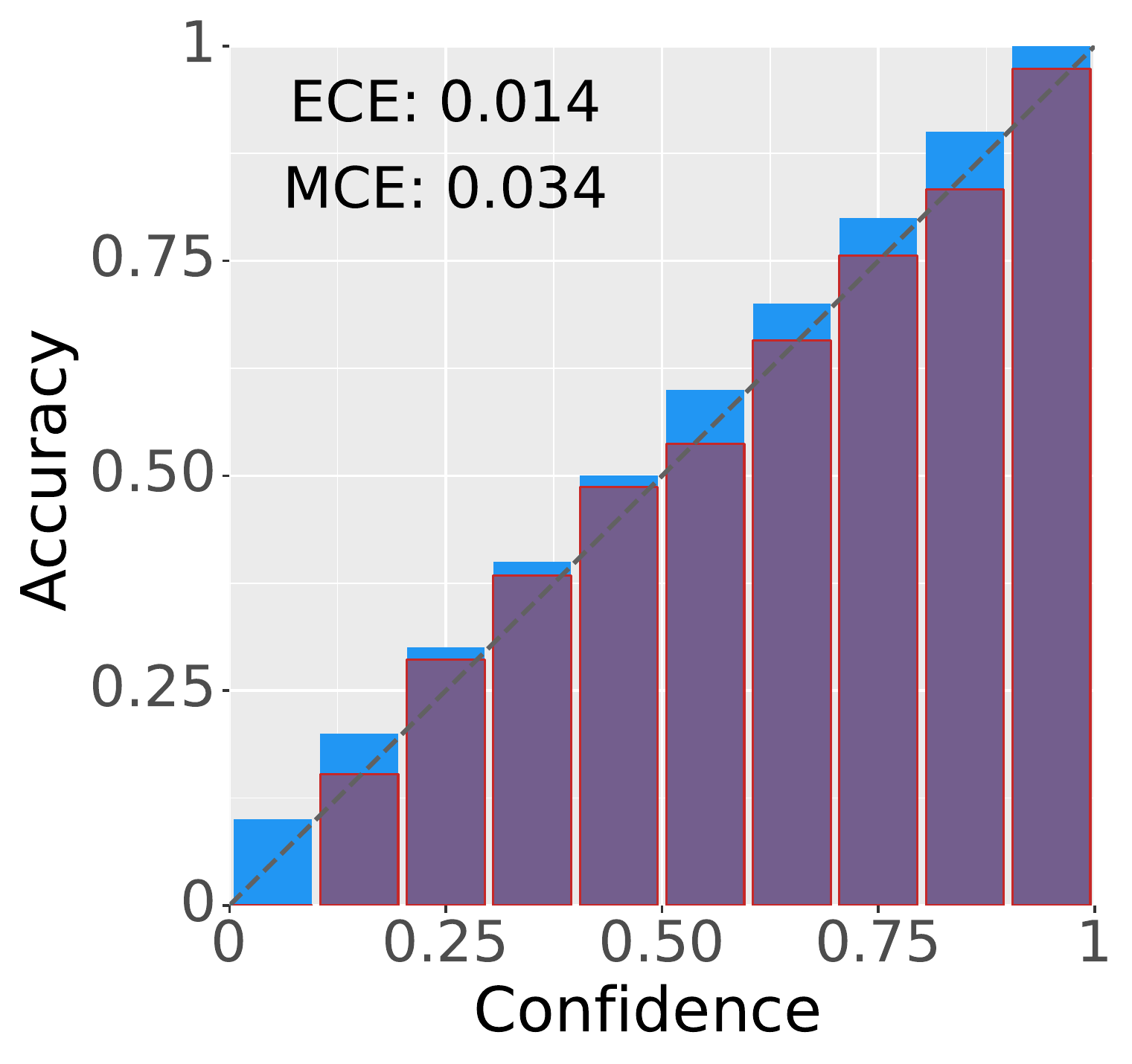}
\caption{$\fnew$ Calibrated}
\end{subfigure}
\hfill
\caption{Calibration of LeNet models trained on SVHN.}
\label{fig:calibration_svhn}
\end{figure}

\begin{table}[htbp]
    \centering
    \caption{Changes in prediction confidence ranking due to temperature scaling for LeNet models on SVHN.}
    \begin{tabular}{lccc}\toprule
              Switch To & Benign & Good & Bad \\\midrule
    $\fbase$          & 0 & 0 & 0  \\
    $\fnew$           & 1587 & 37 & 35   \\
    \bottomrule
    \end{tabular}
    \label{table:calibration_ranking_changes_svhn}
\end{table}

\subsubsection*{FashionMNIST}
\begin{figure}[H]
\centering
\begin{subfigure}[t]{0.24\textwidth}
\centering
\includegraphics[width=0.99\linewidth]{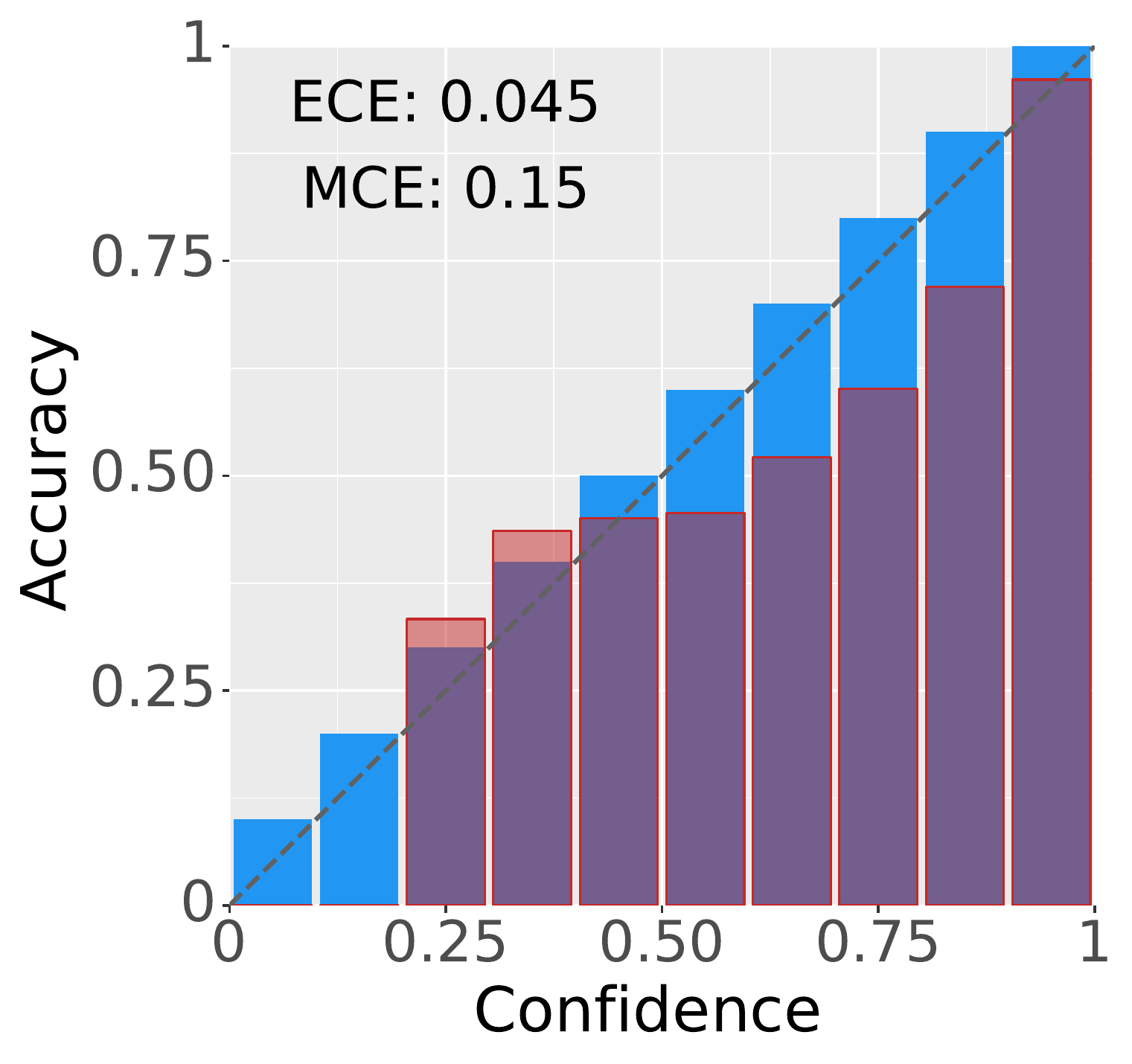}
\caption{$\fbase$ Uncalibrated}
\end{subfigure}
\hfill
\begin{subfigure}[t]{0.24\textwidth}
\centering
\includegraphics[width=0.99\linewidth]{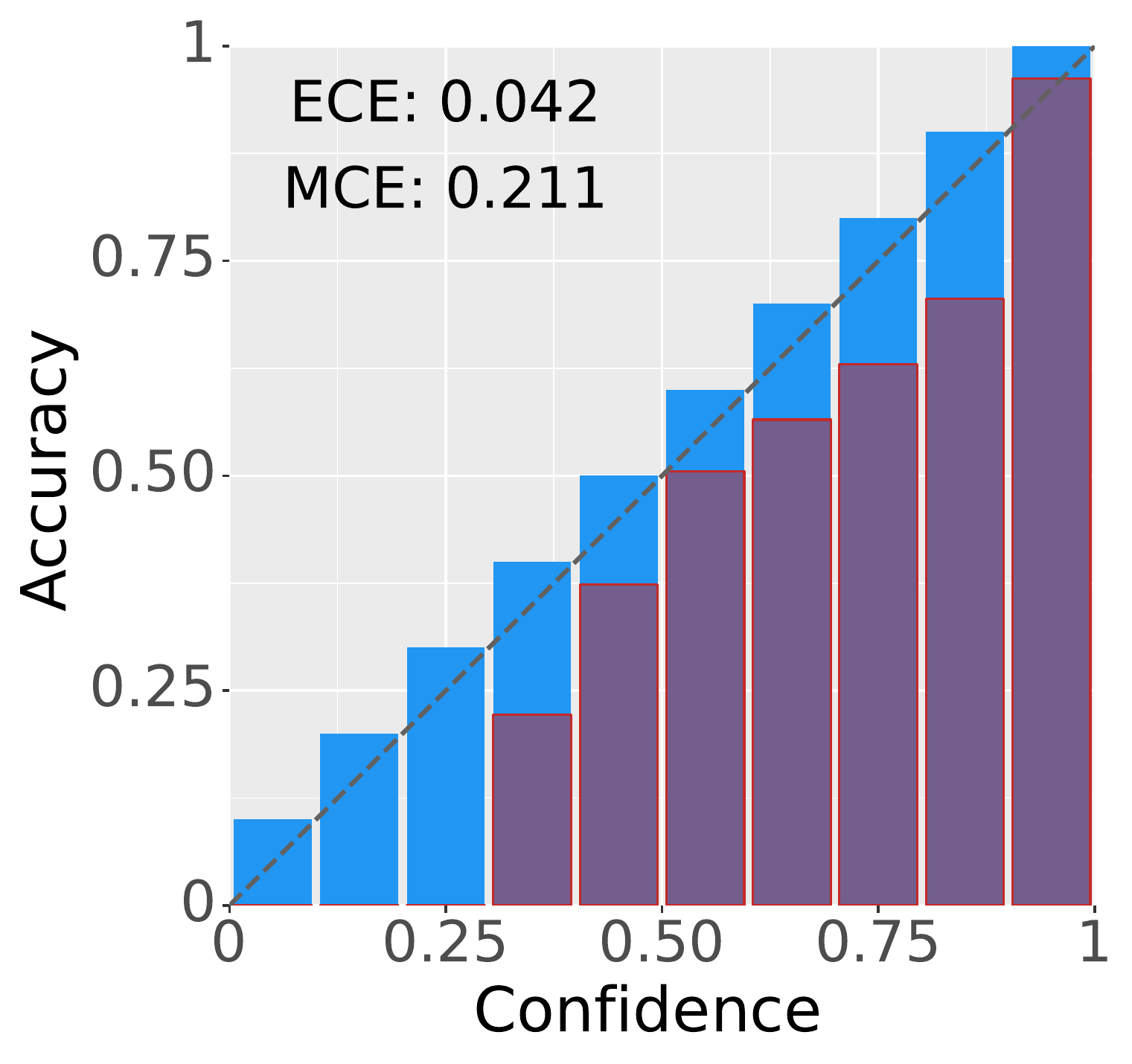}
\caption{$\fnew$ Uncalibrated}
\end{subfigure}
\hfill
\begin{subfigure}[t]{0.24\textwidth}
\centering
\includegraphics[width=0.99\linewidth]{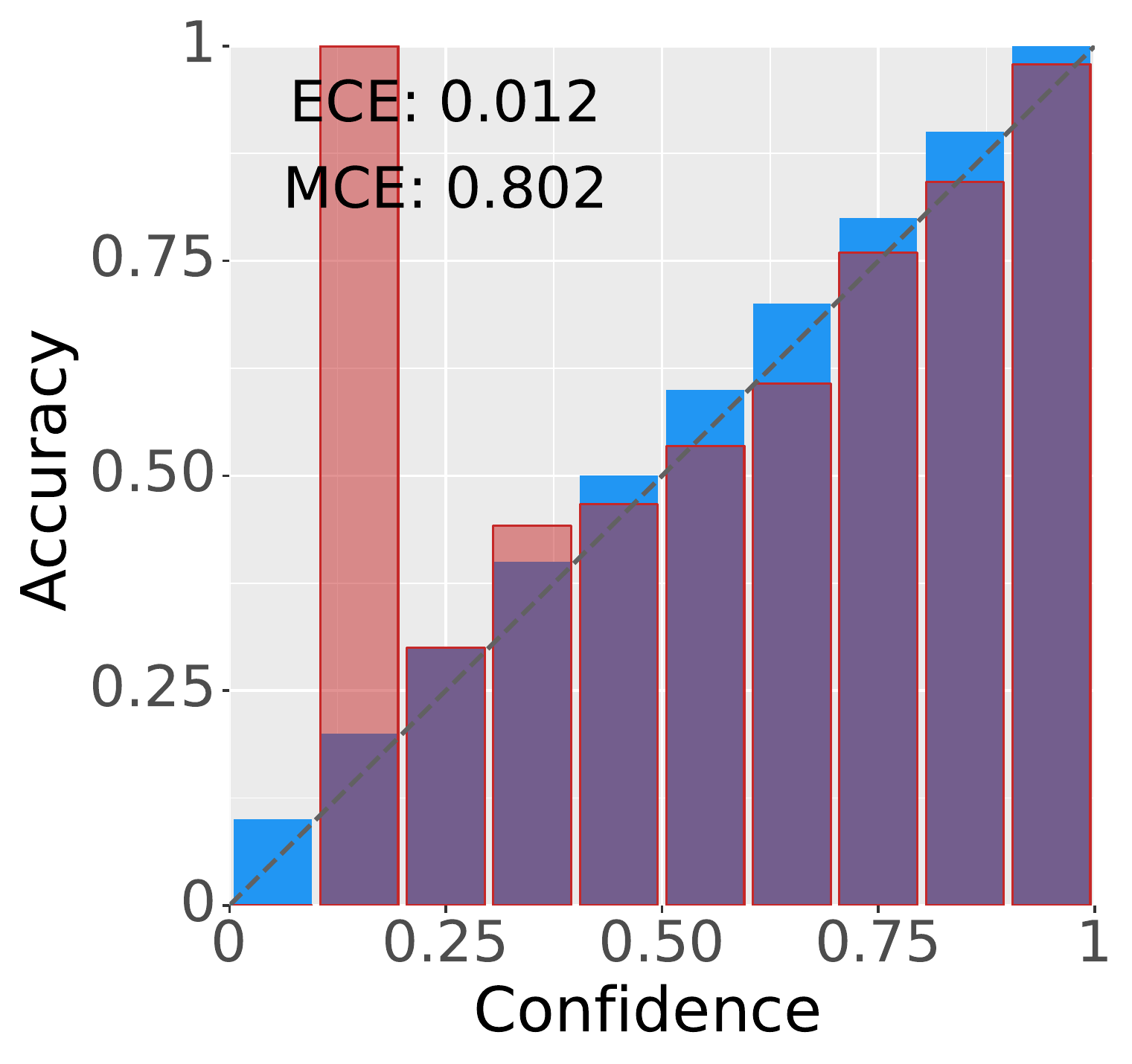}
\caption{$\fbase$ Calibrated}
\end{subfigure}
\hfill
\begin{subfigure}[t]{0.24\textwidth}
\centering
\includegraphics[width=0.99\linewidth]{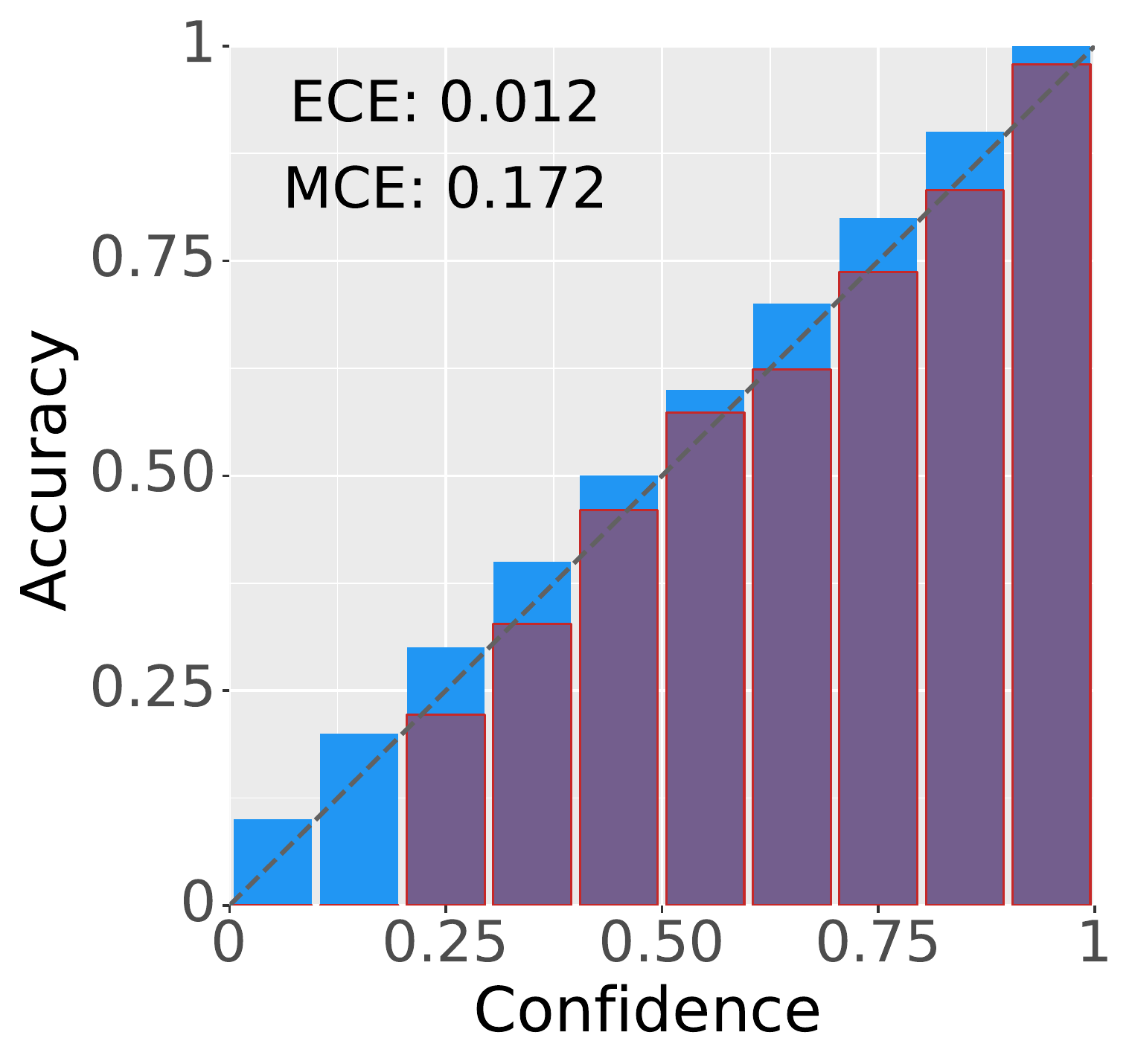}
\caption{$\fnew$ Calibrated}
\end{subfigure}
\hfill
\caption{Calibration of LeNet models trained on FashionMNIST.}
\label{fig:calibration_fashion_mnist}
\end{figure}

\begin{table}[htbp]
    \centering
    \caption{Changes in prediction confidence ranking due to temperature scaling for LeNet models on FashionMNIST.}
    \begin{tabular}{lccc}\toprule
              Switch To & Benign & Good & Bad \\\midrule
    $\fbase$          & 601 & 2 & 3  \\
    $\fnew$           & 118 & 12 & 10   \\
    \bottomrule
    \end{tabular}
    \label{table:calibration_ranking_changes_fashion_mnist}
\end{table}

\begin{table}[H]
\begin{center}
    \caption{Changes in prediction confidence ranking due to temperature scaling for a ResNet18 model on CIFAR10.}
    \begin{tabular}{lccc}\toprule
              Switch To & Benign & Good & Bad \\\midrule
    $\fbase$          & 234 & 9 & 18  \\
    $\fnew$           & 41 & 11 & 11   \\
    \bottomrule
    \end{tabular}
    \label{table:calibration_ranking_changes}
\end{center}
\end{table}
\newpage 
\section{Prediction Confidence on Negative Flips}
\label{sec:nfr_confidence_distribution}
\begin{figure}[H]
\centering
\includegraphics[width=0.7\linewidth]{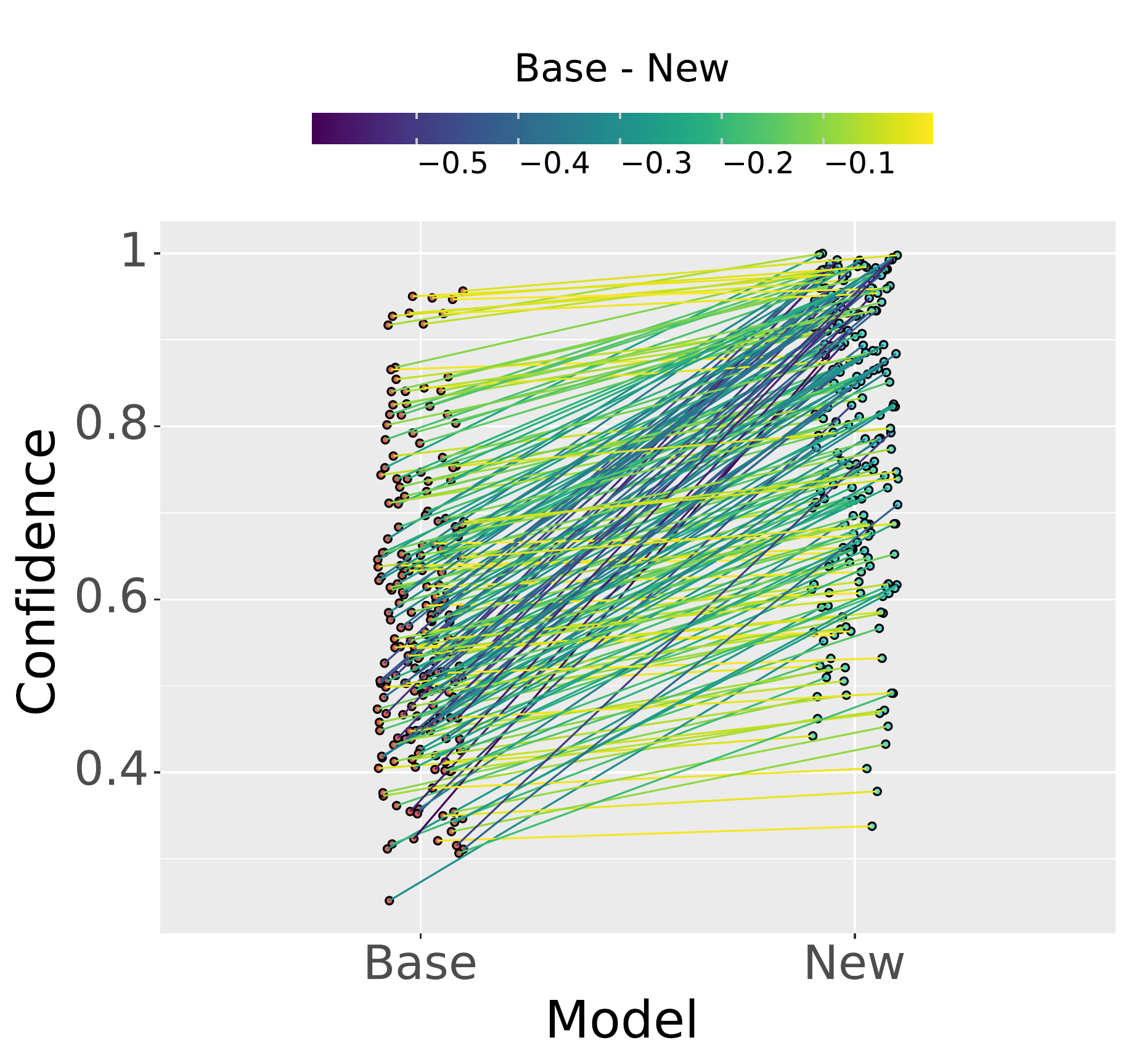}
\caption{Paired prediction confidence of the predicted class for the base and new model on negative flips where the new model is more confident than the base model. The color indicates the difference in prediction confidence between the base and new model. ResNet18 model trained on CIFAR10 where the base model is trained on 30000 samples, and the new model is trained on an extra 10000 samples from scratch with no churn reduction regularization. These negative flips cannot be reduced further by \amc{} Conf since the new model predicts higher confidence than the base model but is incorrect.}
\label{fig:paired_plot}
\end{figure}

To get a better understanding of why calibrating models post-hoc with temperature scaling does not improve the effectiveness of the Conf score, we visualize the paired prediction confidence of $\fbase$ and $\fnew$ on negative flips where $\max_{k} \fnew(x)_{k} > \max_{k} \fbase(x)_{k}$ (Figure \ref{fig:paired_plot}). If temperature scaling does not make $\fnew$ less confident on average, or $\fbase$ more confident on average, or both, these negative flips cannot be further reduced by Conf. Since both $\fbase$ and $\fnew$ are both overconfident as was shown in the main text, temperature scaling does not change the ranking in prediction confidence on many samples, hence why negative flips are not reduced further. Moreover, even if $\fbase$ and $\fnew$ are perfectly calibrated, 0 negative flips cannot be achieved by the Conf score since if for example $\max_{k} \fbase(x)_{k} = 0.5$ and $\max_{k} \fnew(x)_{k} = 0.51$, then the probability of choosing the correct model is essentially a coin flip. Thus, full negative flip reduction would require occasionally choosing the lower confidence model.

\newpage
\section{\amc{} Conceptual Figure}
\begin{figure}[H]
\begin{center}
\begin{tikzpicture}[x=1cm,y=1cm]
    
    \begin{scope}[local bounding box=scope1]
        \message{^^JNeural network without text}
          \readlist\Nnod{2,3,1} 
          
          \message{^^J  Layer}
          \foreachitem \N \in \Nnod{ 
            \def\lay{\Ncnt} 
            \pgfmathsetmacro\prev{int(\Ncnt-1)} 
            \message{\lay,}
            \foreach \i [evaluate={\y=\N/2-\i; \x=\lay; \n=\nstyle;}] in {1,...,\N}{ 
              
              \node[node \n] (N\lay-\i) at (\y,\x) {};
              
              \ifnum\lay>1 
                \foreach \j in {1,...,\Nnod[\prev]}{ 
                  \draw[connect arrow,white,line width=1.2] (N\prev-\j) -- (N\lay-\i);
                  \draw[connect arrow] (N\prev-\j) -- (N\lay-\i);
                }
              \fi 
              
            }
          }
            \node[above=1.3,align=center,outer sep=0] (f1) at (N2-2)  {$\fbase$};
            \node[draw, trapezium, above=0.9cm of f1] (agg) {$\fbase(x)$};
            
            \draw (f1.north west) -- (agg.bottom side);
            \draw (f1.north east) -- (agg.bottom side);
            \node[above=0.5cm of agg] {\large initial deployment};
    \end{scope} 
\end{tikzpicture}
\qquad
\begin{tikzpicture}[x=1cm,y=1cm]
    
    \begin{scope}[local bounding box=scope1]
        \message{^^JNeural network without text}
          \readlist\Nnod{2,3,1} 
          
          \message{^^J  Layer}
          \foreachitem \N \in \Nnod{ 
            \def\lay{\Ncnt} 
            \pgfmathsetmacro\prev{int(\Ncnt-1)} 
            \message{\lay,}
            \foreach \i [evaluate={\y=\N/2-\i; \x=\lay; \n=\nstyle;}] in {1,...,\N}{ 
              
              \node[node \n] (N\lay-\i) at (\y,\x) {};
              
              \ifnum\lay>1 
                \foreach \j in {1,...,\Nnod[\prev]}{ 
                  \draw[connect arrow,white,line width=1.2] (N\prev-\j) -- (N\lay-\i);
                  \draw[connect arrow] (N\prev-\j) -- (N\lay-\i);
                }
              \fi 
              
            }
          }
          \node[above=1.3,align=center,outer sep=0] (f1) at (N2-2)  {$\fbase$};
    \end{scope} 
    \begin{scope}[shift={(4cm, 0)}]
        \message{^^JNeural network without text}
          \readlist\Nnod{2,3,1} 
          
          \message{^^J  Layer}
          \foreachitem \N \in \Nnod{ 
            \def\lay{\Ncnt} 
            \pgfmathsetmacro\prev{int(\Ncnt-1)} 
            \message{\lay,}
            \foreach \i [evaluate={\y=\N/2-\i; \x=\lay; \n=\nstyle;}] in {1,...,\N}{ 
              
              \node[node \n] (NR\lay-\i) at (\y,\x) {};
              
              \ifnum\lay>1 
                \foreach \j in {1,...,\Nnod[\prev]}{ 
                  \draw[connect arrow,white,line width=1.2] (NR\prev-\j) -- (NR\lay-\i);
                  \draw[connect arrow] (NR\prev-\j) -- (NR\lay-\i);
                }
              \fi 
              
            }
          }
          \node[above=1.3,align=center,outer sep=0] at (NR2-2) (fm) {$\fnew$};
    \end{scope} 
    
  \node[draw, trapezium, above=1.1cm of f1] at ($(f1)!0.5!(fm)$) (agg) {$\psi(x; \fbase, \fnew)$};
  \draw (f1.north west) -- (agg.bottom side);
  \draw (fm.north east) -- (agg.bottom side);
  \node[above=0.5cm of agg] {\large update};
\end{tikzpicture}
\end{center}
\caption{\amc{} for reducing churn. A combination of the base and new model's outputs are used to generate the final prediction.}
\label{fig:stacking}
\end{figure}
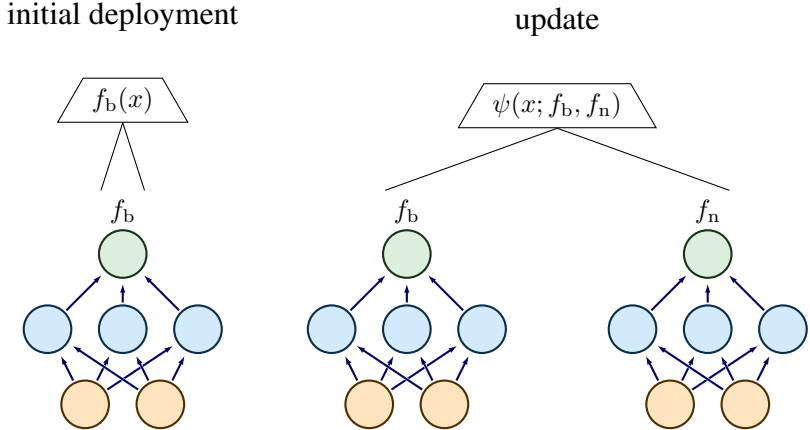

\section{Low Curvature Training}
\begin{figure}[H]
\centering
\includegraphics[width=0.6\linewidth]{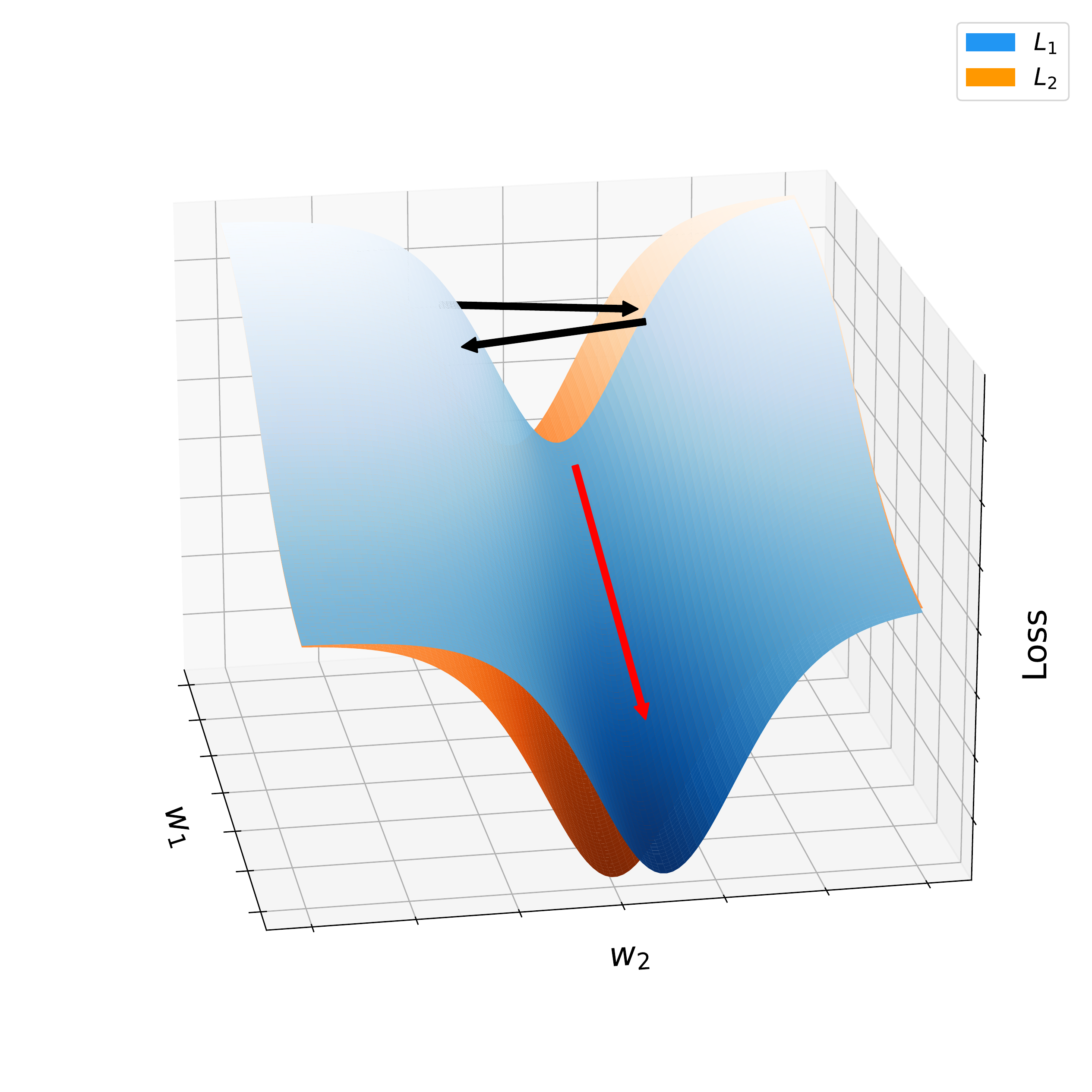}
\caption{Illustration of loss functions for two different samples. Moving the high curvature direction of the black arrows can cause prediction flips minimizing $L_{1}$ in the direction of $w_{2}$ eventually ends up increasing $L_{2}$. However, moving in the low curvature direction of the red arrow enables use to decrease both $L_{1}$ and $L_{2}$ monotonically.}
\label{fig:curvature_hypothesis}
\end{figure}

One explanation for per-sample gradients pointing in opposite direction to the average batch gradient is high curvature directions in the loss landscape. Figure \ref{fig:curvature_hypothesis} shows how if the total loss is a sum of 2 per-sample losses, moving in directions of high curvature can result in decreasing the loss on one sample at the cost of increasing it on the other sample. However, this can be mitigated by moving in the low curvature direction which monotonically decreases the loss on both samples. To perform gradient descent in a low curvature subspace, we use information from the Hessian to find a low curvature subspace to project the total gradient onto. Formally, the loss function Hessian is defined as 
\begin{align*}
    H(\theta)_{i,j} = \mathbb{E}_{x, y}\left[\frac{\partial^{2}}{\partial \theta_{i} \partial \theta_{j}} \ell(\phi(f(x, \theta)), y)\right]
\end{align*}

\noindent for which an Eigendecomposition $H(\theta) = Q \Lambda Q^{-1}$ can be found. Let $U = \mathrm{span}\{q_{1}, ..., q_{k}\}$ be the high curvature subspace spanned by the first k Eigenvectors of $Q$ corresponding to the top $k$ eigenvalues sorted in decreasing order. The average batch gradient is projected onto the low curvature subspace which is the orthogonal complement $U_{\perp}$. Namely
\begin{align*}
    \hat{g} = \mathrm{proj}_{U_{\perp}}g = g - \mathrm{proj}_{U} g
\end{align*}

\begin{figure}[H]
\centering
\begin{subfigure}[t]{0.45\linewidth}
\centering
\includegraphics[width=0.99\columnwidth]{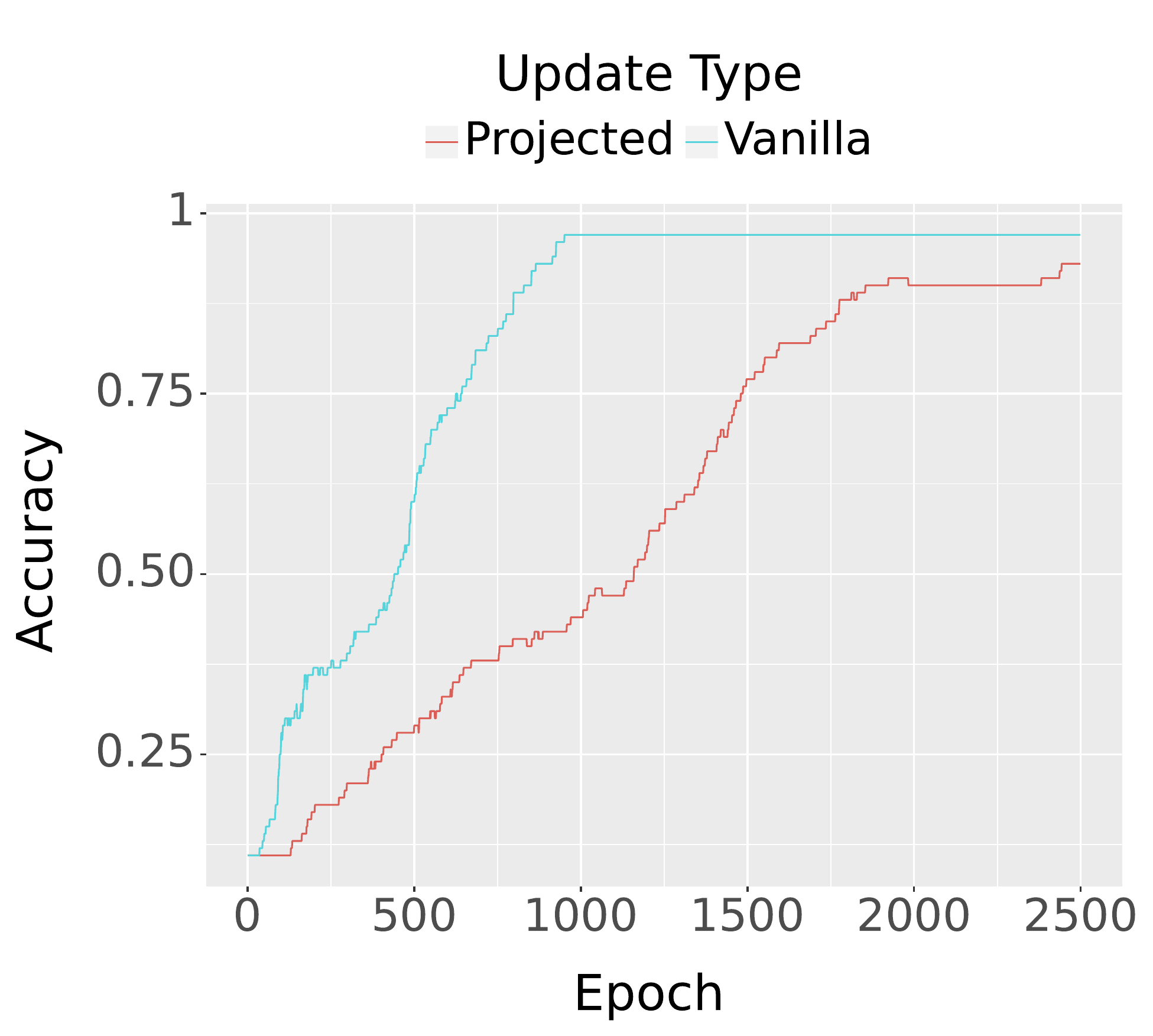}
\caption{}
\label{fig:mnist_low_curvature_accuracy}
\end{subfigure}
\hfill
\begin{subfigure}[t]{0.45\linewidth}
\centering
\includegraphics[width=0.96\columnwidth]{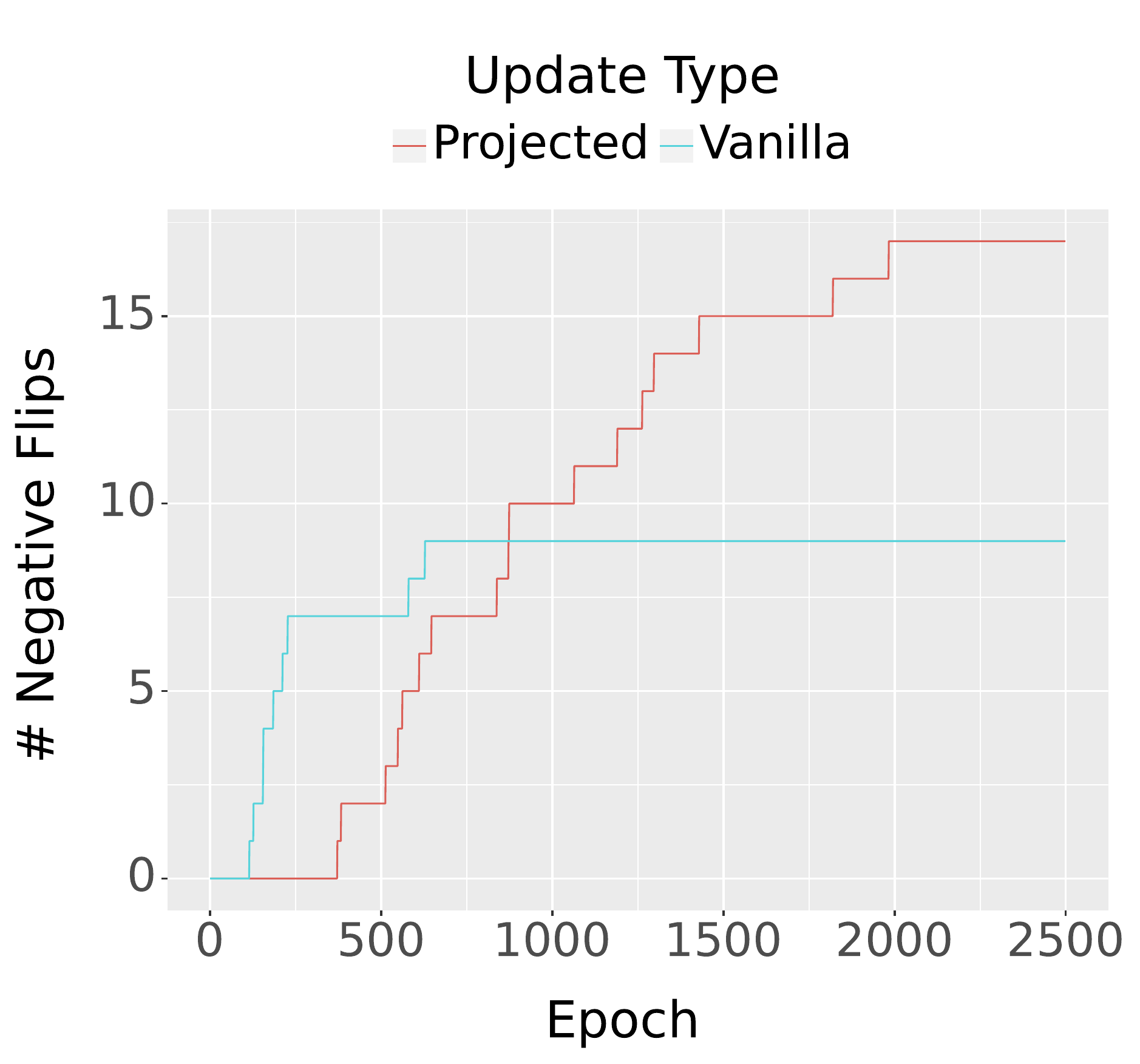}
\caption{}
\label{fig:mnist_low_curvature_nfr}
\end{subfigure}
\caption{a) Accuracy of regular vs. low-curvature projected gradient descent. A LeNet model was trained on a 100 sample subset of MNIST using full batch gradient descent with a learning rate of 0.005. Regular gradient descent is able to reach 100\% accuracy by the final epoch while the low-curvature version is not. b) The cumulative number of negative flips while training. The projected gradient  is able to eliminate some negative flips early on, though has more negative flips towards the end of training.}
\label{fig:mnist_low_curvature}
\end{figure}

\begin{figure}[H]
\centering
\includegraphics[width=0.55\linewidth]{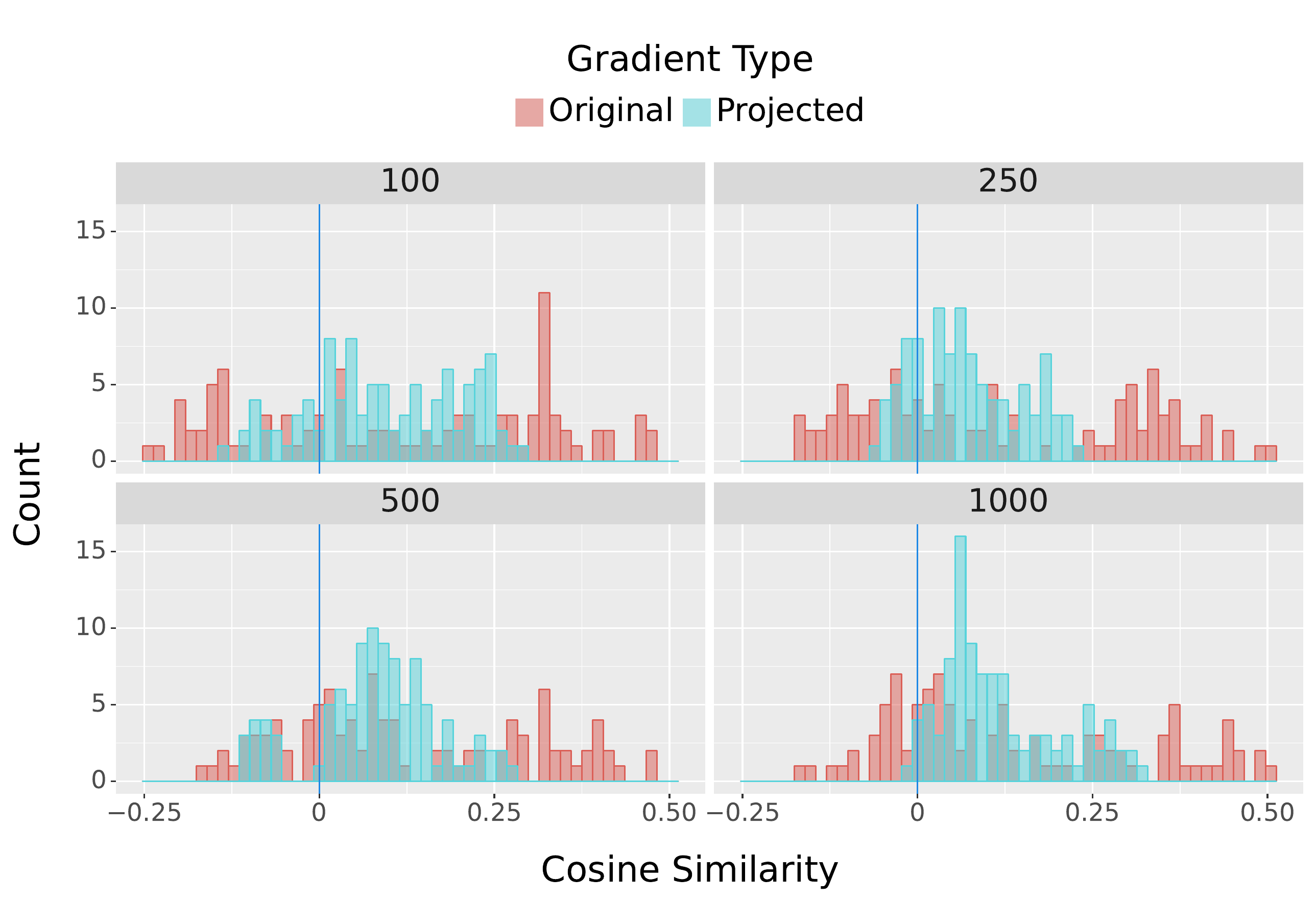}
\caption{Comparison of gradient cosine similarity distribution between regular gradient descent and low-curvature gradient descent. A LeNet model was trained on a 100 sample subset of MNIST using full batch gradient descent with a learning rate of 0.005. Headings indicate the epoch of the model checkpoint. Low-curvature gradient descent limits the most negative cosine similarities to some extent, though, does not entirely eliminate them. }
\label{fig:low_curvature_distribution}
\end{figure}

\noindent To demonstrate the feasibility of this approach in eliminating incompatible gradients, an experiment is performed on a 100 sample subset of MNIST data using full batch gradient descent such that minibatch ordering is not a source of stochasticity. Hessian Eigenvectors corresponding to largest Eigenvalues are computed using the PyHessian package \cite{Yao2019-ul}. The first $k=10$ Eigenvectors are used in the experiment. Figure \ref{fig:low_curvature_distribution} compares the cosine similarity between per sample gradients and the total gradient for both the standard and projected version. Some of the negative cosine similarities are eliminated by the projected gradient. However, this comes at the cost of reducing the maximum positive cosine similarity as well which means slower overall convergence on those samples. Furthermore, negative cosine similarities are not eliminated entirely, and the remaining samples which have negative cosine similarity with the projected gradient are still at risk of of a prediction flip. Larger values of $k \in (25, 50, 100)$ were investigated, but this made little difference in the elimination of incompatible gradients. Figure \ref{fig:mnist_low_curvature} shows that this method does little to reduce the number of prediction flips relative to standard gradient descent, so it has limited promise for attaining self-consistent training.

\section{Efficiently Reducing Incompatible Gradients}
\label{sec:dual_qp}
Here we describe how to more efficiently solve the quadratic programming problem 
\begin{align*}
    \min_{\Tilde{g}} \quad \frac{1}{2} &||g - \Tilde{g}||^{2}_{2} \quad
    \mathrm{s.t.} \quad \langle\Tilde{g}, g_{j}\rangle \: \geq 0 \: \: \: \forall j \in [\mathcal{B}]
\end{align*}

This can be restated as 
\begin{align}
\label{eq:gem_primal}
\begin{split}
    \min_{x} \quad &\frac{1}{2}x^{\top} x - g^{\top}x + \frac{1}{2} g^{\top}g \\
    \mathrm{s.t.} \quad &Gx \succeq 0
\end{split}
\end{align}

\noindent where $G=(g_{1}, ..., g_{|\mathcal{B}|})$. The $g^{\top}g$ term can be ignored since it is constant w.r.t. $x$. The Lagrangian is then
\begin{align*}
    L(x, \lambda) = \frac{1}{2} x^{\top}x - g^{\top}{x} - \lambda^{\top}Gx
\end{align*}

\noindent The infimum of $L(x, \lambda)$ is then found by setting $\nabla_{x} L(x, \theta)=0$ and solving for $x$ to obtain the dual function
\begin{align*}
    \nabla_{x} L(x, \theta) &= x - g - G^{\top}\lambda \\
    x^{*} &= g + G^{\top}\lambda \\
    D(\lambda) &= \frac{1}{2} (g + G^{\top}\lambda)^{\top} (g + G^{\top}\lambda) - g^{\top}(g + G^{\top}\lambda) - \lambda^{\top}G(g + G^{\top}\lambda) \\
    &= -\frac{1}{2} \lambda^{\top}GG^{\top}\lambda - g^{\top}G^{\top}\lambda
\end{align*}

\noindent where $D$ is used to denote the dual function since $g$ is already in use. This provides us a more efficient dual formulation of eq. \ref{eq:gem_primal}
\begin{align*}
    \min_{\lambda} \quad & \frac{1}{2} \lambda^{\top}GG^{\top}\lambda + g^{\top}G^{\top}\lambda \\
    \mathrm{s.t.} \quad & \lambda \succeq 0
\end{align*}

\noindent where $\lambda \in R^{|\mathcal{B}|}$ such that optimization is done over the number of samples in the training set $\mathcal{B}$ instead of being over the number of parameters in the model which can easily reach millions for moderate size CNNs. For large enough training sets, even this dual formulation becomes computationally impractical. 

\section{Self-Consistent Learning Details}
We trained LeNet models trained for 12500 epochs on a random 1000 sample subset of SVHN with a variety of learning rates (0.005, 0.001, 0.0005, 0.0001) using full batch versions of regular gradient descent and constrained gradient descent. We focus on results for a learning rate of 0.0005 as that achieved both 100\% accuracy, and a monotonically decreasing loss. Larger learning rates resulted in increases in loss at some epochs such that negative flips could be attributed to the learning rate being too large. The smaller learning rate
 of 0.0001 was not able to achieve perfect training accuracy, and there is an accuracy gap between constrained and regular descent methods, which makes it difficult to compare their cumulative negative flips. 
We also normalized the updates to have unit $\ell_{2}$ norm. This was done to make the comparison between the two methods as fair as possible, as we observed large differences in magnitude between the regular and constrained gradient directions. 

\begin{figure}[t]
\centering
\begin{subfigure}[t]{0.23\linewidth}
\centering
\includegraphics[width=0.99\columnwidth]{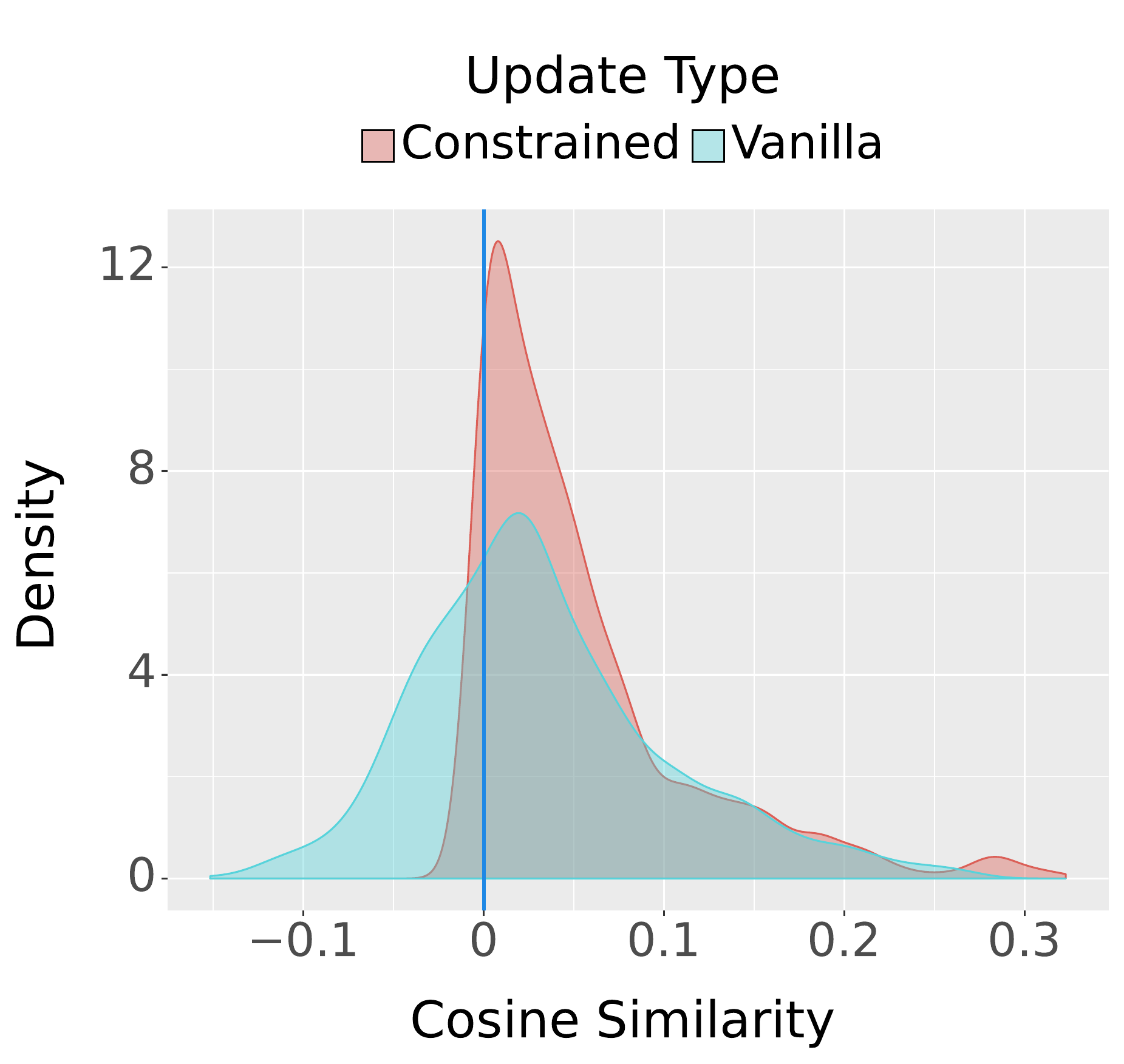}
\caption{}
\end{subfigure}
\hfill
\begin{subfigure}[t]{0.23\linewidth}
\centering
\includegraphics[width=0.99\columnwidth]{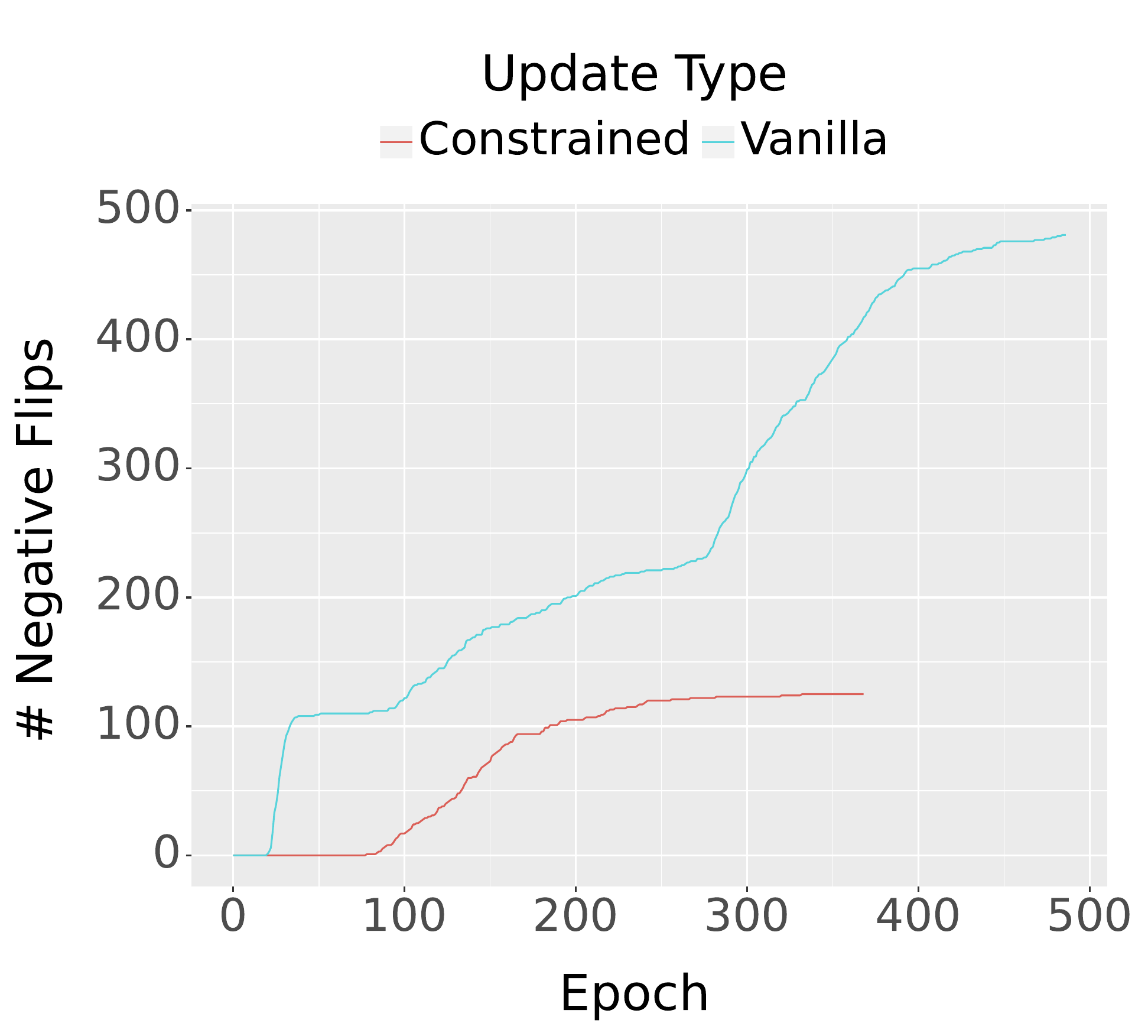}
\caption{}
\end{subfigure}
\hfill
\begin{subfigure}[t]{0.23\linewidth}
\centering
\includegraphics[width=0.99\columnwidth]{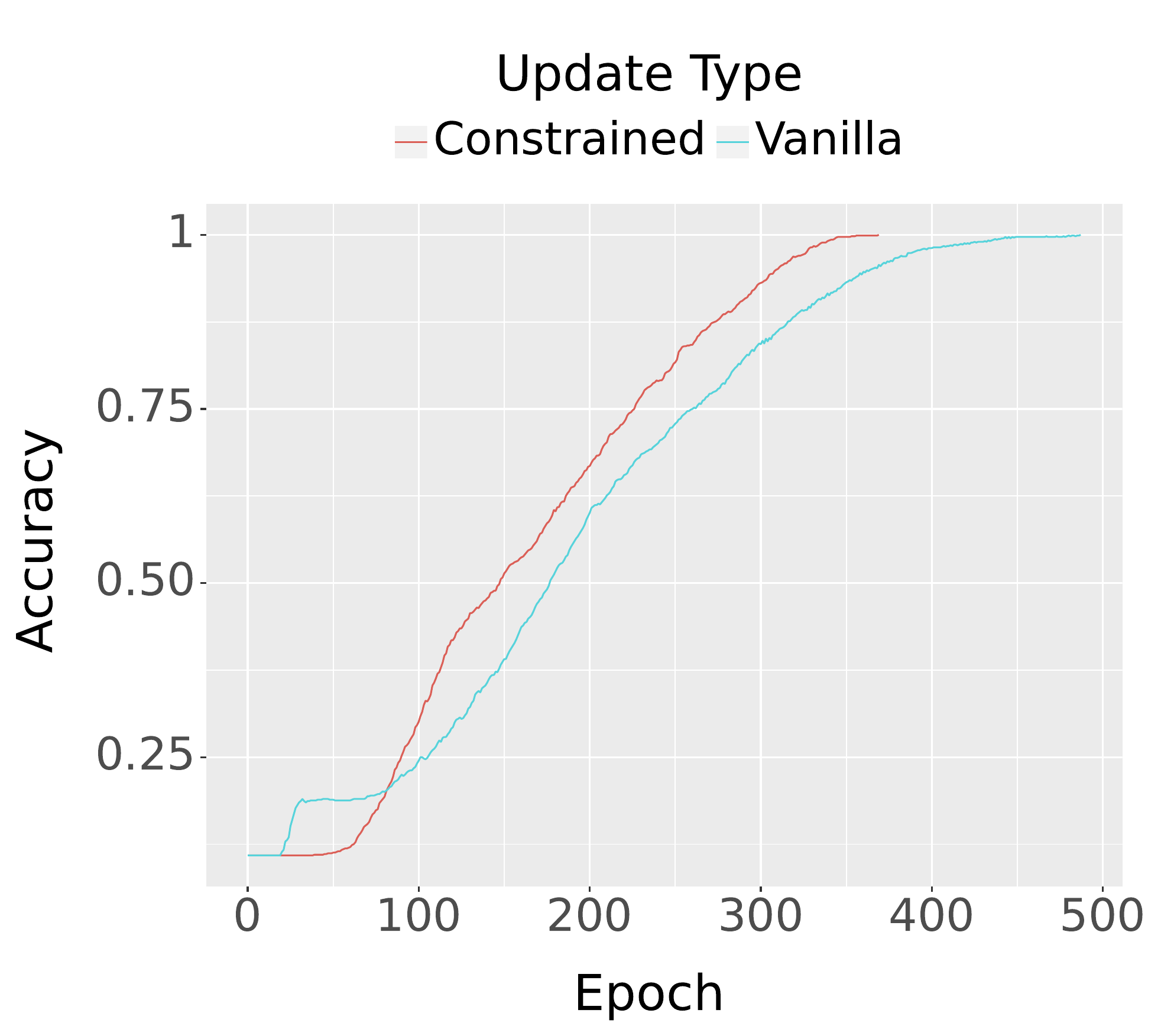}
\caption{}
\end{subfigure}
\hfill
\begin{subfigure}[t]{0.23\linewidth}
\centering
\includegraphics[width=0.99\columnwidth]{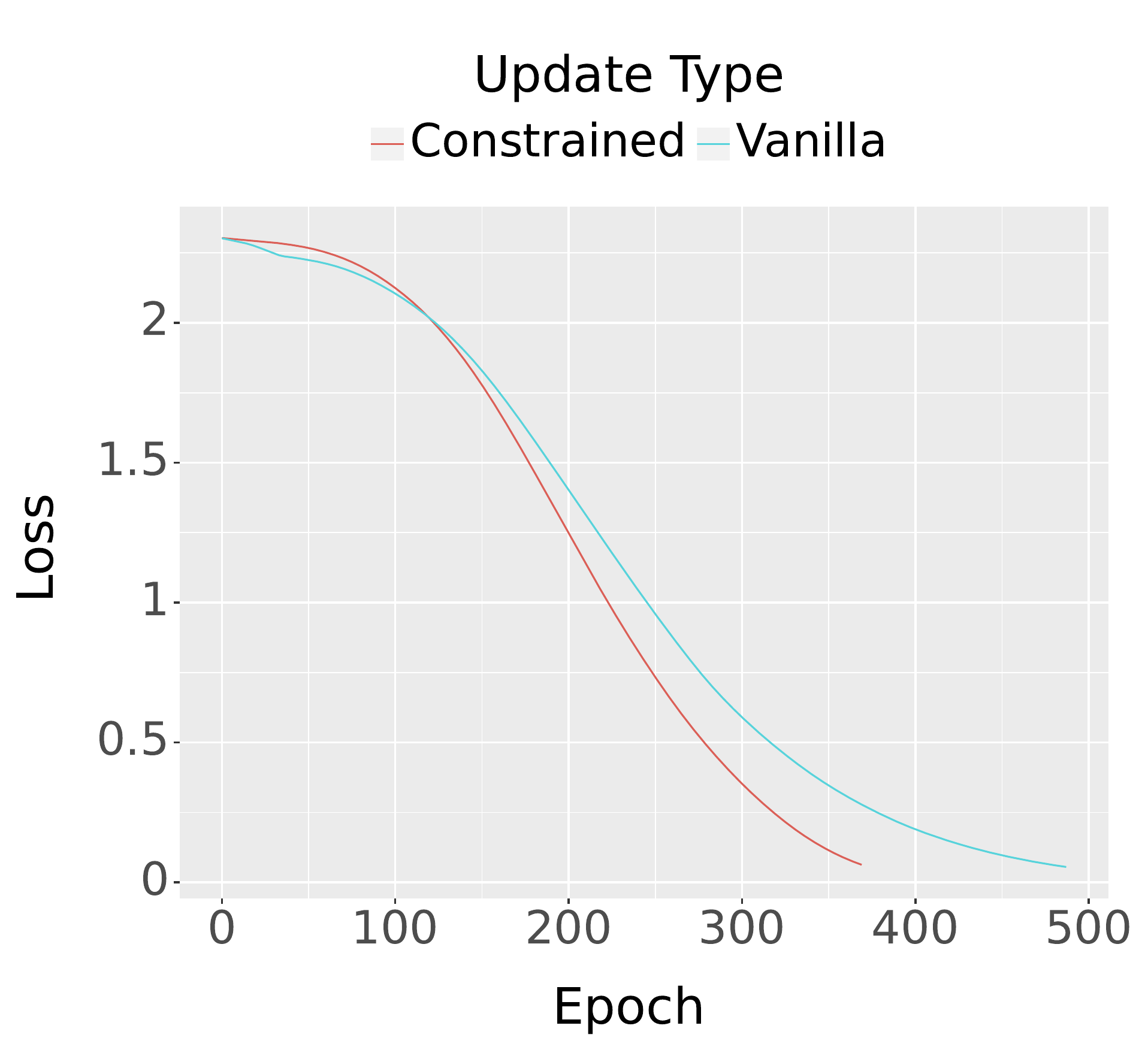}
\caption{}
\end{subfigure}
\caption{a) Gradient cosine similarity at a mid-stage epoch when training a LeNet model on a random subset of 1000 samples from SVHN using Adam with learning rate 0.0001. The average batch gradient when using regular gradient descent is incompatible with many per-sample gradients (blue area). Using constrained optimization, we can find a direction that is compatible with all per-sample gradients (red area). b) The cumulative number of negative flips while training until 100\% accuracy. The constrained optimization approach is able to eliminate many negative flips, confirming that some negative flips are a result of incompatible gradients. c) and d) show that both approaches achieve 100\% accuracy and near-zero loss by the final epoch of training.}
\label{fig:adam_quadprog_flips}
\end{figure}

\begin{figure}[t]
\centering
\begin{subfigure}[t]{0.23\linewidth}
\centering
\includegraphics[width=0.99\columnwidth]{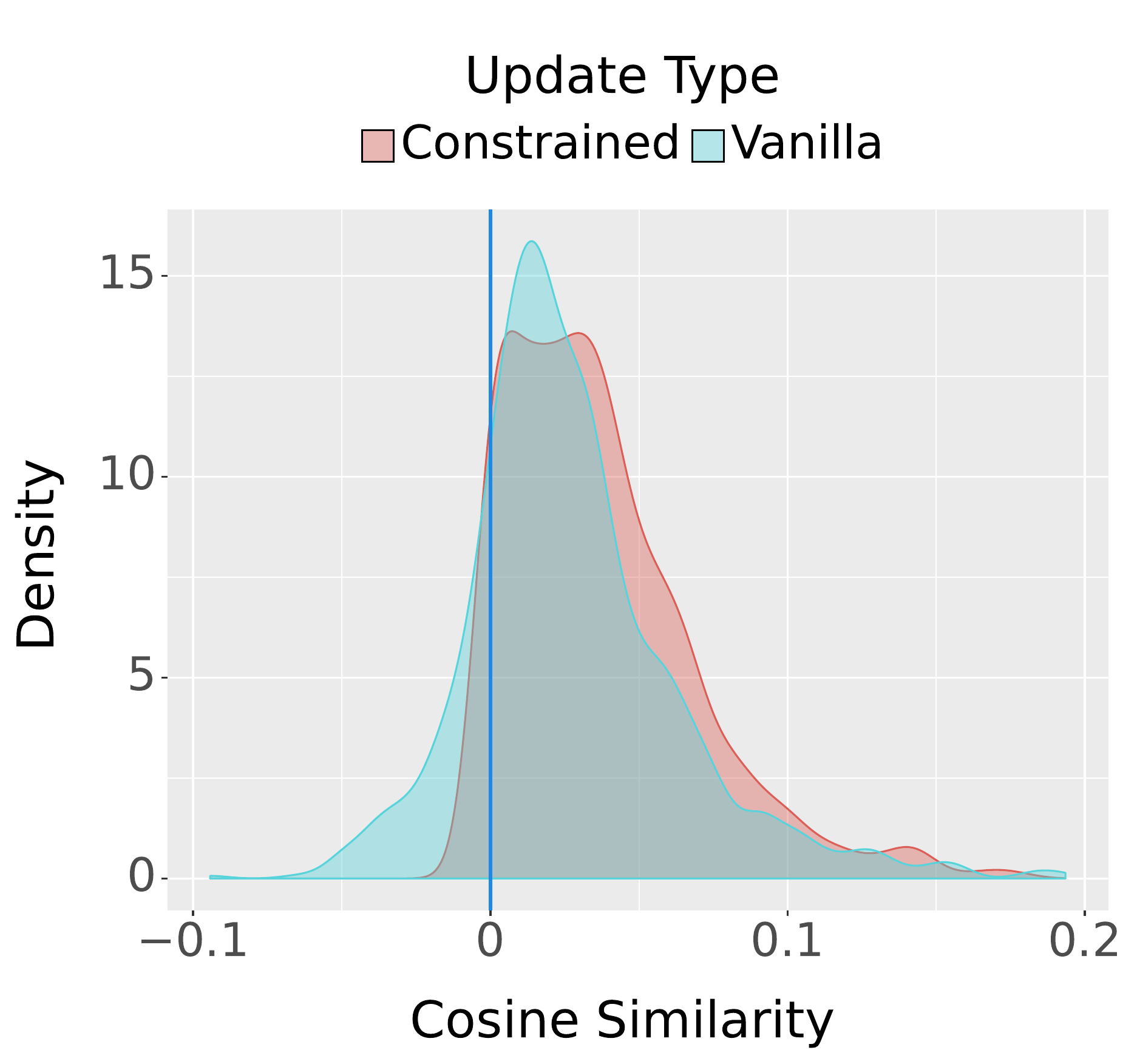}
\caption{}
\end{subfigure}
\hfill
\begin{subfigure}[t]{0.23\linewidth}
\centering
\includegraphics[width=0.99\columnwidth]{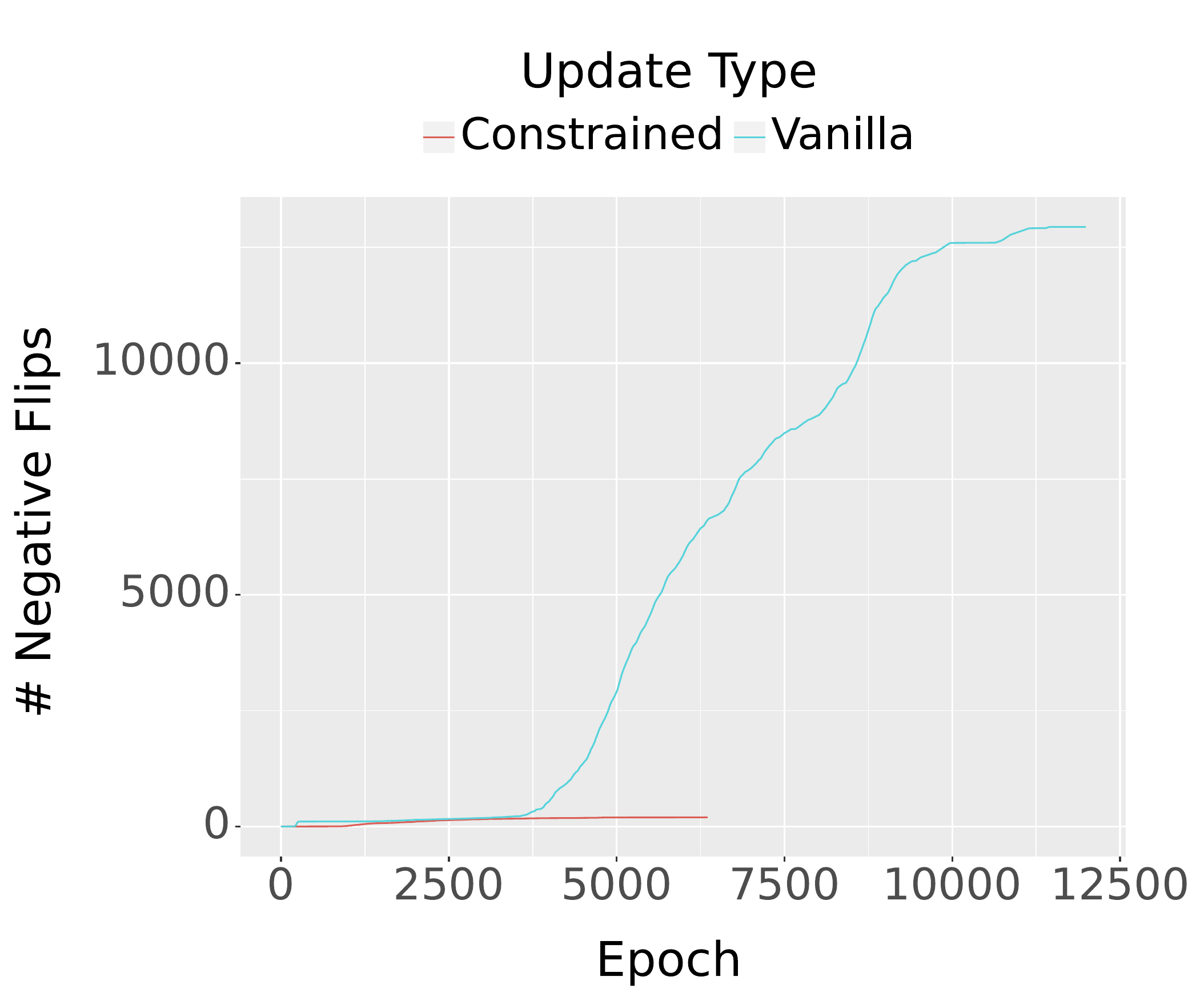}
\caption{}
\end{subfigure}
\hfill
\begin{subfigure}[t]{0.23\linewidth}
\centering
\includegraphics[width=0.99\columnwidth]{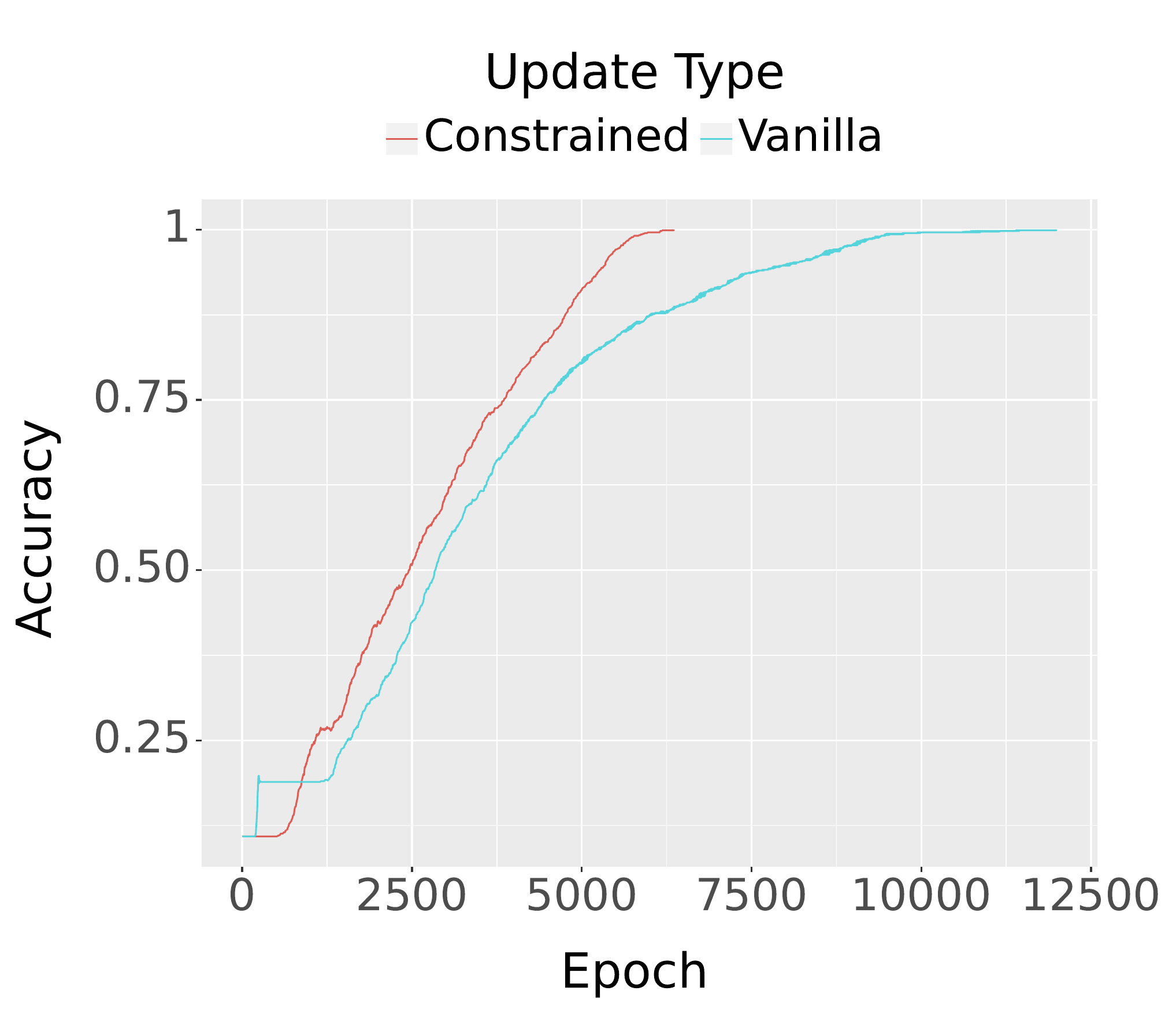}
\caption{}
\end{subfigure}
\hfill
\begin{subfigure}[t]{0.23\linewidth}
\centering
\includegraphics[width=0.99\columnwidth]{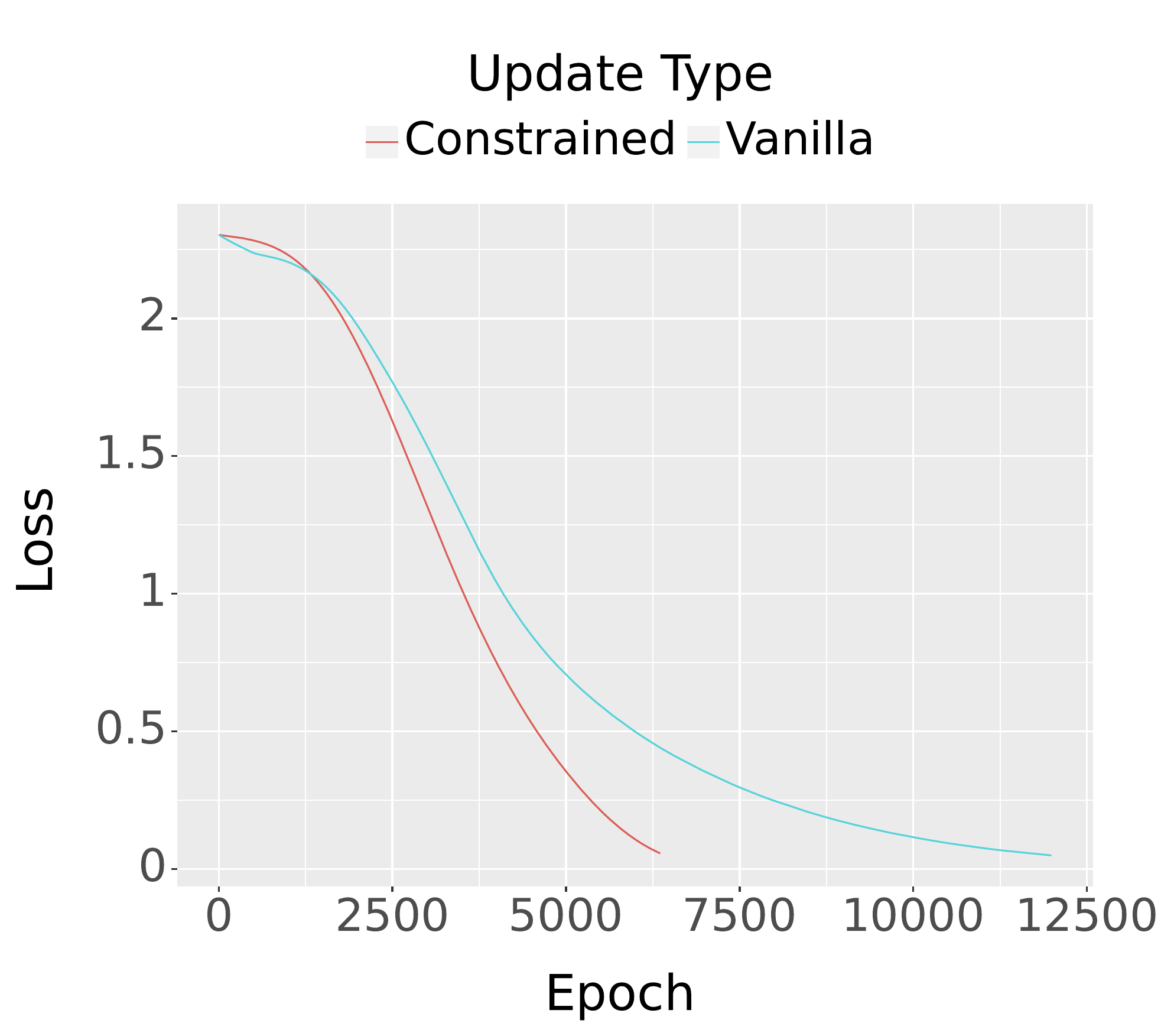}
\caption{}
\end{subfigure}
\caption{a) Gradient cosine similarity at a mid-stage epoch when training a LeNet model on a random subset of 1000 samples from SVHN using gradient descent with learning rate 0.001. The average batch gradient when using regular gradient descent is incompatible with many per-sample gradients (blue area). Using constrained optimization, we can find a direction that is compatible with all per-sample gradients (red area). b) The cumulative number of negative flips while training until 100\% accuracy. The constrained optimization approach is able to eliminate many negative flips, confirming that some negative flips are a result of incompatible gradients. c) and d) show that both approaches achieve 100\% accuracy and near-zero loss by the final epoch of training.}
\label{fig:sgd_high_lr_quadprog_flips}
\end{figure}

Figures \ref{fig:adam_quadprog_flips} and \ref{fig:sgd_high_lr_quadprog_flips} show the effectiveness of self-consistent training using the Adam optimizer with learning rate 0.0001, and gradient descent with learning rate 0.001 respectively. In both cases, the projected gradient method which reduces incompatible gradients results in significantly fewer negative prediction flips throughout training. 

\section{kNN AvgConf Approximation}
\label{sec:knn_avgconf}
Algorithms \ref{alg:augmented_train} and \ref{alg:estimated_score} detail the procedure. First, the \avgconf{} for samples in the validation set is computed throughout the training process. This introduces almost no computational overhead as inference on the validation set is already performed during training for performance monitoring purposes. Second, validation sample embeddings using the final learned model are extracted and stored. Lastly, to compute \avgconf{} for a new test sample $x$, the embedding of $x$ is extracted, the nearest validation set neighbors are found, and the mean of the \avgconf{} of those neighbors is used as the approximate \avgconf{} of $x$. Using the ball tree structure for k-nearest neighbors, computing \avgconf{} is $O(kd \log{}(n))$ which is inexpensive relative to the inference cost of an architecture such as ResNet18 which performs more than $10^9$ operations \cite{He2015-ov}. Here $k$ is the number of neighbors, $d$ is the embedding dimensionality, and $n$ is the number of validation set samples.

\begin{algorithm}[H]
   \caption{Augmented training procedure}
   \label{alg:augmented_train}
\begin{algorithmic}
   \STATE {\bfseries Input:} training set $\dtrain$, validation set $\dval$, number of epochs $T$, feature extractor $e$, classification head $h$
   \STATE $f := h \circ e$
   \FOR{$t=1$ to $T$}
   \STATE Train $f$ on $\dtrain$\ \COMMENT{one epoch of training}
   \STATE $P^{t} \gets [\phi(f(x_{\mathrm{val}}^{(i)}))]_{i=1}^{\nval}$ \COMMENT{epoch probabilities}
   \ENDFOR
   \STATE $\hat{y} \gets [\sigma(P^{T}_{i})]_{i=1}^{\nval}$\ \COMMENT{final predictions}
   \STATE $S \gets [\frac{1}{T} \sum_{t=1}^{T} P^{t}_{i, \hat{y}_{i}}]_{i=1}^{\nval}$ \COMMENT{AvgConf scores}
   \STATE $E \gets [e(x^{(i)}_{\mathrm{val}})]_{i=1}^{\nval}$ \COMMENT{embeddings}
   \STATE {\bfseries Output:} Trained feature extractor $e$, trained classification head $h$, validation embeddings E, validation AvgConf scores S
\end{algorithmic}
\end{algorithm}

\begin{algorithm}[H]
   \caption{Estimated AvgConf computation}
   \label{alg:estimated_score}
\begin{algorithmic}
   \STATE {\bfseries Input:} evaluation sample $x$, validation embeddings $E$, validation AvgConf scores $S$, number of neighbors k, feature extractor $e$
    \STATE $B(z, r) := $ $\ell^{2}$ ball of radius $r$ centered at $z$
   \STATE $r \gets \inf \{r: |B(e(x), r) \cap E| \ge k \}$ \COMMENT{ball radius}
   \STATE $\mathcal{I} \gets \{ i: E_{i} \in B(e(x), r) \cap E \}$ \COMMENT{neighbor indices}
   \STATE $s \gets \frac{1}{|\mathcal{I}|} \sum_{i \in \mathcal{I}} S_{i}$
   \STATE {\bfseries Output:} Estimated AvgConf score s
\end{algorithmic}
\end{algorithm}

We consider how effective our kNN estimate of AvgConf is by comparing the exact AvgConf score to the estimated score. Note that we choose k=10 as a fixed hyperparameter that we do not optimize over, as our goal is to keep \amc{} efficient, even if that sacrifices accuracy or churn reduction slightly. Table \ref{table:knn_correlation} shows the correlation between the scores for a single run on CIFAR10, CIFAR100, and FashionMNIST. Table \ref{table:knn_churn} shows the churn for both the exact and estimated AvgConf scores on their own, as well as combined with Conf. For CIFAR10 and CIFAR100, there is a significant decrease in churn reduction ability when using the estimated score instead of the exact score, either using AvgConf on its own or in combination with Conf. A similar observation can be made for accuracy as seen in table \ref{table:knn_accuracy}. This is to be expected as the performance of the models on these two datasets is much lower than on FashionMNIST, so the extracted embeddings for kNN regression purposes are not as effective. On FashionMNIST there is essentially no difference between using the estimated and exact AvgConf score. 

\begin{table}[H]
    \centering
    \caption{Correlation between AvgConf and kNN-estimated AvgConf scores.}
    \begin{tabular}{lcc}\toprule
    Dataset           & Pearson $r$ & Spearman $\rho$ \\\midrule
    CIFAR10           & 0.84 & 0.87   \\
    CIFAR100          & 0.75 & 0.67  \\
    FashionMNIST      & 0.86 & 0.95  \\
    \bottomrule
    \end{tabular}
    \label{table:knn_correlation}
\end{table}

\begin{table}[H]
    \centering
    \caption{Churn of exact and estimated AvgConf Scores}
    \begin{tabular}{lcccc}\toprule
    Dataset           & \multicolumn{1}{p{1.5cm}}{\centering AvgConf \\ Exact} & \multicolumn{1}{p{1.5cm}}{\centering AvgConf \\ Estimated} & \multicolumn{1}{p{1.5cm}}{\centering Combined \\ Exact} & \multicolumn{1}{p{1.5cm}}{\centering Combined \\ Estimated} \\\midrule
    CIFAR10           & 2.10 & 3.15 & 1.27 & 1.81  \\
    CIFAR100          & 3.22 & 5.78 & 2.18 & 2.80 \\
    FashionMNIST      & 1.19 & 1.22 & 0.72 & 0.7  \\
    \bottomrule
    \end{tabular}
    \label{table:knn_churn}
\end{table}

\begin{table}[H]
    \centering
    \caption{Accuracy of exact and estimated AvgConf Scores}
    \begin{tabular}{lcccc}\toprule
    Dataset           & \multicolumn{1}{p{1.5cm}}{\centering AvgConf \\ Exact} & \multicolumn{1}{p{1.5cm}}{\centering AvgConf \\ Estimated} & \multicolumn{1}{p{1.5cm}}{\centering Combined \\ Exact} & \multicolumn{1}{p{1.5cm}}{\centering Combined \\ Estimated} \\\midrule
    CIFAR10           & 85.04 & 84.33 & 84.34 & 84.26  \\
    CIFAR100          & 53.24 & 49.70 & 52.35 & 50.43 \\
    FashionMNIST      & 91.76 & 91.53 & 91.61 & 91.54  \\
    \bottomrule
    \end{tabular}
    \label{table:knn_accuracy}
\end{table}

\section{Flip Counts}
\label{sec:flip_counts}
The number of negative and positive flips made by each version of \amc{} for a LeNet model trained on FashionMNIST, and a ResNet18 model trained on CIFAR10. For Conf, AvgConf, and Combined, $\psi$ is not capable of making additional negative or positive flips compared to just using $\fnew$ since $\psi$ chooses between $\fbase$ and $\fnew$. However, \amc{} Learned is capable of making new predictions that differ from both $\fbase$ and $\fnew$, so while it is as effective as Conf at PFs, and also adds new ones (green and blue bar for Learned PFs higher than Conf), it introduces new errors on samples that both $\fbase$ and $\fnew$ correctly predict which results in a higher total number of NFs compared to Conf.

\begin{figure}[H]
\centering
\begin{subfigure}{0.4\textwidth}
\centering
\includegraphics[width=0.9\textwidth]{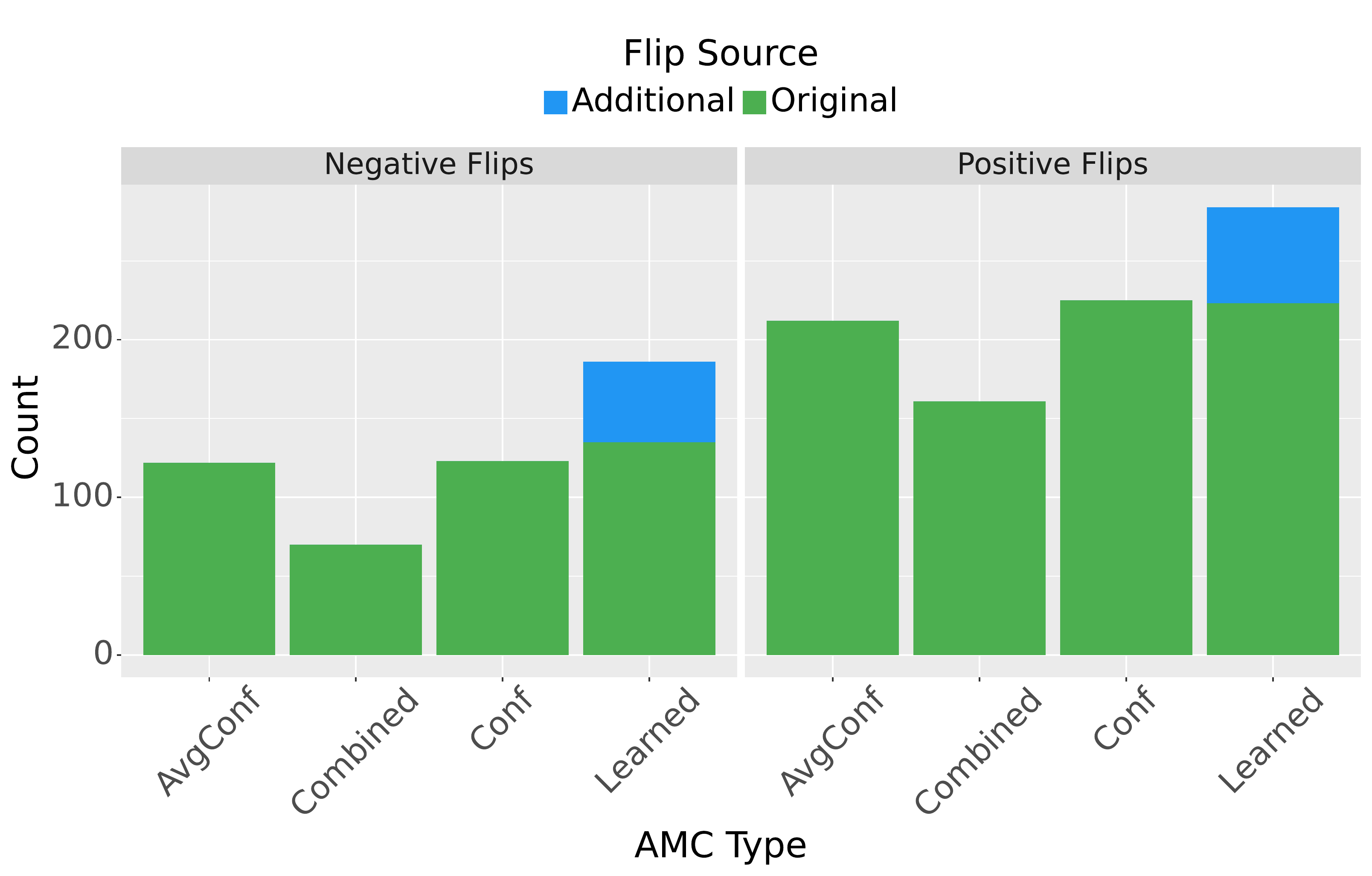}
\caption{FashionMNIST}
\end{subfigure}
\begin{subfigure}{0.4\textwidth}
\centering
\includegraphics[width=0.9\textwidth]{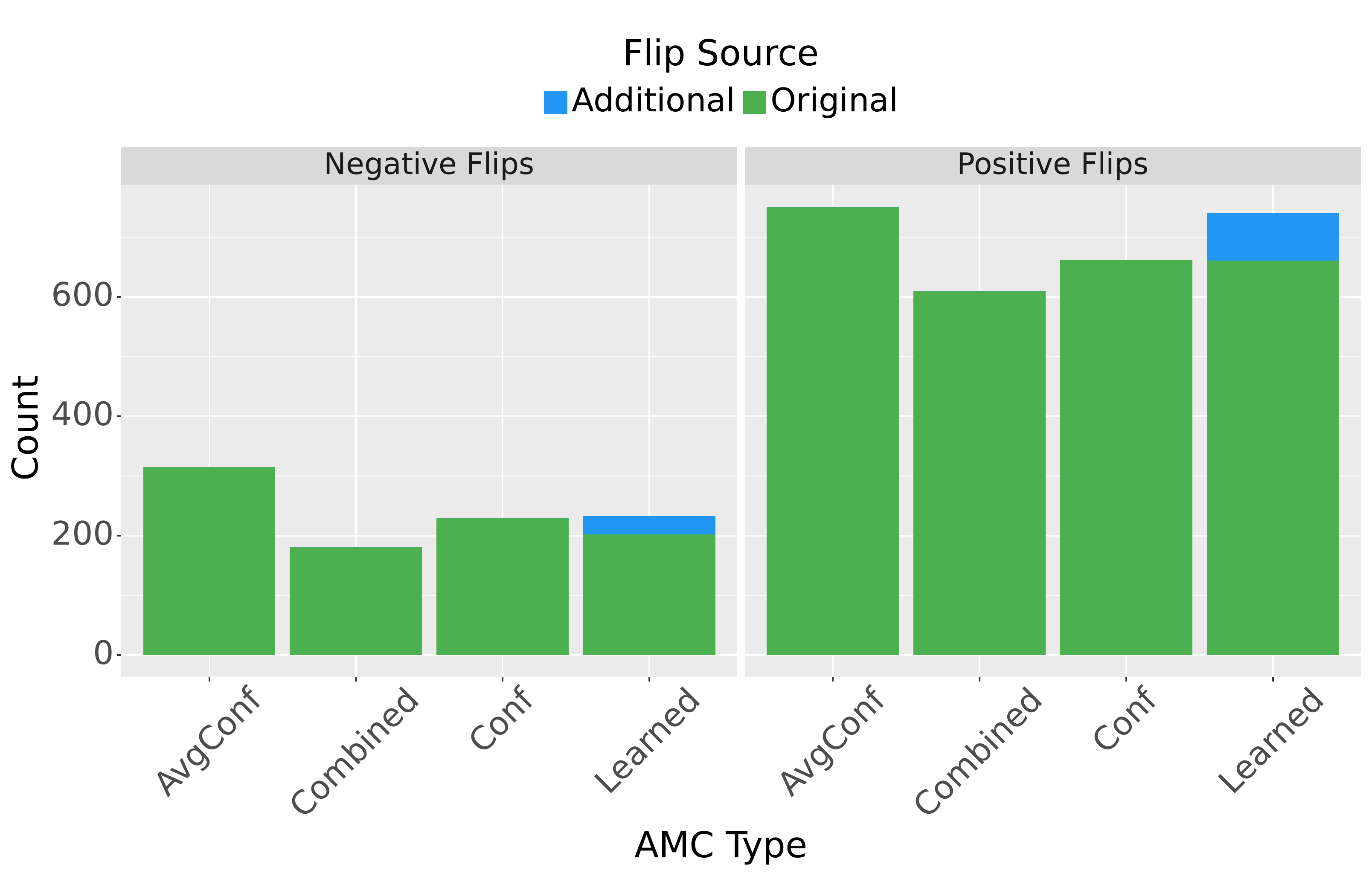}
\caption{CIFAR10}
\end{subfigure}
\caption{
  Flip counts for the 4 versions of \amc{} for a LeNet model trained on FashionMNIST. The additional NFs and PFs of Learned come from its ability to generate new predictions rather than choose between $\fbase$ and $\fnew$.
  }
\label{fig:flip_counts}
\end{figure}

\section{Proposition 4.1 Proof}
\label{sec:proposition_proof}
We restate proposition 4.1 for convenience.

\begin{proposition}
Assume $\fbase$ and $\fnew$ are perfectly calibrated such that $\mathbf{Pr}(\hat{Y} = Y | \hat{P} = p) = p, \forall p \in [0, 1]$ where $\hat{Y} = \argmax_{k} f(X)_{k}$ (hard prediction) and $\hat{P} = \max_{k} \phi(f(X))_{k}$ (predicted probability). If $s = \max_{k} \phi(f(x))_{k}$ so that $\psi$ is also perfectly calibrated, then $\acc(\psi) \geq \acc(\fnew)$.
\end{proposition}
\begin{proof}
Perfect calibration implies that model accuracy $\acc(f) = \mathbb{E}_{\hat{P}} [\mathbf{Pr}(\hat{Y} = Y | \hat{P} = p)]$ where the expectation is taken w.r.t. the distribution of predicted probabilities $\hat{P}$. The choice of $s$ results in $\psi$ choosing the highest confidence model, so we end up with 
\begin{align*}
    \acc(\psi) &= \mathbb{E}_{\hat{P}_{\psi}} [\mathbf{Pr}(\hat{Y}_{\psi} = Y | \hat{P}_{\psi} = p)] \\
    &= \mathbb{E}_{X} \left[ \max_{k} \phi \left( \psi(X) \right)_{k} \right] \; \; \; \; \; \; \mathrm{\# \; by \; calibration} \\    
    &\geq \mathbb{E}_{X} \left[ \max_{k} \phi \left( \fnew(X) \right)_{k} \right] \\
    &= \mathbb{E}_{\hat{P}_{\mathrm{new}}} [\mathbf{Pr}(\hat{Y}_{\mathrm{new}} = Y | \hat{P}_{\mathrm{new}} = p)] \\
    &= \acc(\fnew) \qedhere
\end{align*}
\end{proof}